\documentclass{article}

\usepackage{arxiv}

\usepackage[utf8]{inputenc} 
\usepackage[T1]{fontenc}    
\usepackage{hyperref}       
\usepackage{url}            
\usepackage{booktabs}       
\usepackage{amsfonts}       
\usepackage{nicefrac}       
\usepackage{microtype}      
\usepackage{lipsum}		
\usepackage{graphicx}
\usepackage{natbib}
\usepackage{doi}
\usepackage{amsmath}
\usepackage[section]{algorithm}
\usepackage{algpseudocode}
\usepackage{subcaption}
\usepackage{booktabs}
\usepackage{multirow}

\newtheorem{definition}{Definition}

\usepackage{xcolor}

\newcommand{\pc}[1]{#1}

\newcommand{\logperp}{\ell\textit{perp}_{\theta}}
\newcommand{\perplexity}{\textit{perp}_{\theta}}
\newcommand{\dtrain}{D_{train}}
\newcommand{\dtest}{D_{test}}

\DeclareMathOperator*{\argmax}{argmax}

\title{Privacy Issues in Large Language Models: A Survey}

\author{ Seth Neel \\
	SAFR AI Lab\\
	Harvard Business School\\
	Cambridge, MA \\
	\texttt{sneel@hbs.edu} \\
 \And
Peter W. Chang \\
	SAFR AI Lab\\
	Harvard Business School\\
	Cambridge, MA \\
	\texttt{pchang@hbs.edu} \\
}

\date{}

\renewcommand{\headeright}{Technical Report}
\renewcommand{\undertitle}{Technical Report}
\renewcommand{\shorttitle}{Privacy Issues in Large Language Models}

\hypersetup{
pdftitle={Privacy Issues in Large Language Models: A Survey},
pdfauthor={Peter W.~Chang, Seth Neel},
pdfkeywords={Large Language Models, Privacy},
}

\begin{document}
\maketitle

\begin{abstract}
    This is the first survey of the active area of AI research that focuses on privacy issues in Large Language Models (LLMs). Specifically, we focus on work that red-teams models to highlight privacy risks, attempts to build privacy into the training or inference process, enables efficient data deletion from trained models to comply with existing privacy regulations, and tries to mitigate copyright issues. Our focus is on summarizing technical research that develops algorithms, proves theorems, and runs empirical evaluations. While there is an extensive body of legal and policy work addressing these challenges from a different angle, that is not the focus of our survey. Nevertheless, these works, along with recent legal developments do inform how these technical problems are formalized, and so we discuss them briefly in Section~\ref{sec:intro}. While we have made our best effort to include all the relevant work, due to the fast moving nature of this research we may have missed some recent work. If we have missed some of your work please contact us, as we will attempt to keep this survey relatively up to date. We are maintaining a repository with the list of papers covered in this survey and any relevant code that was publicly available at \url{https://github.com/safr-ml-lab/survey-llm}.

\end{abstract}


\tableofcontents

\section{Motivation}
\label{sec:intro}
The launch of ChatGPT in November $2022$ took the world by storm, igniting an explosion in AI research, startup activity, and consumer adoption of Large Language Models or LLMs. At the end of $2023$, progress on LLMs has continued apace, but there is a growing consensus that along with transformative economic benefits, LLMs carry significant societal risks. One camp of AI safety advocates has focused on the existential risks of developing superintelligence; as the AI startup Anthropic notes in their position paper on AI safety ``If we build an AI system that is significantly more competent than human experts but it pursues goals that conflict with our best interests, the consequences could be dire.'' A second camp holds that the risks of superintelligent systems built on our current technological foundations is slim, but at the same time harbor serious concerns about the misuse of existing or near-term LLMs. These concerns focus on the reliability, toxicity, bias, security, and privacy of the current generation of language models \cite{crfm}.

In this survey we focus on this last concern, \emph{privacy}, a multi-faceted topic that encompasses research on how models are developed, fine-tuned, deployed, and modified after training. While regulatory frameworks governing LLMs are still in their infancy, many of the major proposals and regulations proposed so far have devoted significant attention to privacy. In the United States, the very recent White House Executive Order on AI \cite{whitehouse} devotes significant time to both the privacy risks of AI, and potential ways to mitigate these risks. On potential risks the report notes that  \emph{''Artificial Intelligence is making it easier to extract, re-identify, link, infer, and act on sensitive information about people’s identities, locations, habits, and desires.  Artificial Intelligence’s capabilities in these areas can increase the risk that personal data could be exploited and exposed}. The first two sections of this survey review existing work exposing these privacy risks; specifically the propensity of large language models to \emph{memorize} the underlying training data (Section~\ref{sec:memorization}), enabling privacy attacks that given access to the model can expose the underlying training data (Section~\ref{sec:mi_attacks}). The extent to which motivated attackers can use access to these models to reverse engineer sensitive training data is still an active area of research, and of potential litigation; in July 2023, the Federal Trade Commission (FTC), the U.S. governmental body tasked with overseeing consumer protection and antitrust law in the United States, opened an investigation into OpenAI, the makers of ChatGPT, on the basis that their data collection practices and services could violate user's privacy rights. 

The Executive Order goes on to discuss specific Privacy Enhancing Technologies (PETs) that can potentially thwart these privacy attacks: ``\emph{To combat this risk, the Federal Government will ensure that the collection, use, and retention of data is lawful, is secure, and mitigates privacy and confidentiality risks. Agencies shall use available policy and technical tools, including privacy-enhancing technologies (PETs) where appropriate, to protect privacy and to combat the broader legal and societal risks,"} going on to mention cryptographic approaches, federated learning, and differential privacy in particular. Section~\ref{sec:private_llms} reviews existing work on DP training and federated training of language models. 

Another privacy issue raised by generative models that has been receiving significant attention pertains to copyright -- in particular whether (i) the original content generated by the model can be protected by copyright law and (ii) whether works generated the model may violate copyright protections afforded to training data used to produce that work. There have already been several high-profile lawsuits filed related to whether specific images or text that were generated by models that include copyrighted data in the training set constitute fair use; Getty Images' lawsuit against Stability AI for training on their images without permission and comedian Sarah Silverman's lawsuit against OpenAI and Meta to name a few \cite{lucchi_2023}. Although the Executive Order does not provide specific guidance on issues of copyright raised by generative models like LLMs, it mandates guidance from the United States Copyright Office to be released in mid-$2024$. While the current legal status of training on copyrighted data is up in the air in the United States, \cite{lucchi_2023} notes ``In the event that a court determines that data ingestion – which involves acquiring unprocessed data from one or more sources and modifying them to render them appropriate for the purpose of training AI machines – constitutes an act of infringement, the entire AI system may encounter significant legal difficulties.'' Like the issue of determining if LLMs infringe on the copyright protection of their training data, determining if LLM-generated works can receive copyright protections themselves is largely unsettled. Complexities here include (i) the requirement of human authorship to hold copyright in many jurisdictions (ii) if forced to attribute the copyright to a human creator, it is unclear who should receive the copyright -- the creators of the training data, the model trainer, or the end user who prompted the creation of the work (iii) several jurisdictions enforce a standard of ``orginality'' to hold copyright which can be difficult to assess for AI-generated works. We discuss early algorithmic work on mitigating copyright issues related to training on copyrighted data in Section~\ref{sec:copyright}. 

Since $2018$ The General Data Protection Regulation (GDPR) has been the major set of regulations mandating data privacy practices and non-compliance penalties for companies doing business in the EU. Although the GDPR does not deal explicitly with AI, many of its provisions can be interpreted to apply to AI systems. One in particular, the ``Right to Erasure'' gives individuals the right to mandate that companies delete their personal data after they submit a request to do so. While this personal data can be deleted from databases in a relatively straightforward fashion (although even this can be tricky given the nuances in how databases are implemented under the hood; see \cite{rtbflaw}), given the ability of models themselves to leak private information from the training data (Sections~\ref{sec:memorization}, \ref{sec:mi_attacks}), it raises the natural question of how a training point can be deleted or ``unlearned'' from a trained model. This an active area of research in machine learning, where unlearning approaches can also be used to remove the influence of toxic, erroneous, or copyrighted data from the model post training \cite{nguyen2022survey}. Section~\ref{sec:unlearning} reviews early work on unlearning from LLMs.

\section{General Terminology}
\label{sec:terminology}

Each section of this survey starts with a preliminaries section outlining the equations and algorithms relevant to that type of privacy setting. Unless otherwise specified, the mathematical notation in those sections will use the following:

\begin{itemize}
    \item $D$: Dataset. Including $\dtrain$ and $\dtest$ for training and test sets respectively.
    \item $\mathcal{A}$: Training algorithm that maps $\dtrain \to h$, a language model.
    \item $h$: Language model resulting from $A(D_{train})$.
    \item $\theta$: Parameters of model $h$.
    \item $\mathcal{Z}$: A privacy attack, such as those describe in Section~\ref{sec:mi_attacks}.
    \item $\mathcal{U}$: Unlearning algorithm (see Section~\ref{sec:unlearning}).
    \item $s$: A sequence, example, or individual string.
    \item $\ell$: Loss function.
    \item $\tau$: Threshold value. Usually in the context of membership inference attacks.
    \item $\epsilon, \delta$: Differential privacy parameters (see Section~\ref{sec:private_llms}).
\end{itemize}

Throughout this survey, there are many acronyms used to represent concepts or models common in privacy and language model research. These acronyms will also be defined at time of first mention, but a table is provided here for some of the most common and important terms.

\begin{table}[!ht]
\begin{center}
    \begin{tabular}{ |p{3cm}|p{8.5cm}|  }
     \hline
     \hfil \textbf{Acronym} &\hfil \textbf{Explanation} \\
     \hline
     AUC & Area Under the Curve \\
     BERT & Bidirectional Encoder Representations from Transformers \\
     BLEU & Bilingual Evaluation Understudy \\
     BPE & Byte-Pair Encoding \\
     CRT & Confidentially Redacted Training \\
     DP & Differential Privacy \\
     DP-SGD & Differentially Private Stochastic Gradient Descent \\
     GPT & Generative Pre-trained Transformer \\
     GPT-\textit{n}, ChatGPT, ... & \textit{Variants of above} \\
     $k$-NAF & $k$ Near Access Freedom \\
     KNN & $k$-Nearest Neighbors \\
     LM & Language Model \\
     LLM & Large Language Model \\
     LOO & Leave One Out \\
     LSTM & Long Short Term Memory network \\
     MIA & Membership Inference Attack \\
     MLM & Masked Language Model \\
     PHI & Personal Health Information \\
     PII & Personally Identifiable Information \\
     SGD & Stochastic Gradient Descent \\
     SISA & Sharded, Isolated, Sliced, and Aggregated training \\
     TPR, FPR & True Positive Rate, False Positive Rate \\
     \hline
    \end{tabular}
    \caption{List of acronyms commonly used throughout this survey}
\end{center}
\end{table}

There are certain datasets that are commonly used by researchers and developers to train LLMs. Many of the largest ones are formed by scraping massive amounts of internet text, while others are collated from domain-specific sources such as healthcare or patent data. These are summarized in Table~\ref{tab:datasets}.

\begin{table}[!ht]
\begin{center}
    \begin{tabular}{ |p{4.5cm}|p{6.5cm}|p{3.85cm}| }
     \hline
     \hfil \textbf{Dataset} &\hfil \textbf{Description} &\hfil \textbf{Citation} \\
     \hline
     AG News Corpus & News articles & \cite{Gulli2004corpus} \\
     BigPatent-G & US Patent documents & \cite{sharma2019bigpatent} \\
     BookCorpus & Collection of free novel books & \cite{zhu2015aligning} \\
     Brown & Modern English linguistic corpus & \cite{browncorpus} \\
     CNNDM & Reading comprehension data & \cite{hermann2015teaching} \\
     Common Crawl & Data scraped from open web & \cite{CommonCrawl2008} \\
     C4 & Cleaned version of Common Crawl & \cite{dodge2021documenting} \\
     mC4 & Multi-lingual \& cleaned Common Crawl & \cite{xue2021mt5} \\
     Conf. on Machine Translation & Corpora obtained from this conference work & \cite{bojar2018findings} \\
     CounterFact & Factual completion prompts & \cite{meng2022romecounterfact} \\
     CustomerSim & Synthetic dialogue of customer interactions & \cite{tkachenko2016customersim} \\
     LIMA & Forums and StackExchange data & \cite{zhou2023lima} \\
     MIMIC-III & Patient data from intensive care units & \cite{Johnson2016mimiciii} \\
     One Billion Word & Preprocessed data from 2011 news & \cite{chelba2014billion}\\
     OpenWebText & Web content extracted from Reddit URLs & \cite{Gokaslan2019OpenWeb} \\ 
     Penn TreeBank & Annotated corpus of English language & \cite{marcus1993penntreebank} \\
     The Pile & Diverse compilation of $22$ smaller datasets & \cite{gao2020pile} \\
     PubMed & Citations/abstracts of biomedical literature & \cite{peng2019pubmed} \\
     RoBERTa & Corpus used to train RoBERTa model & \cite{liu2019roberta}\\
     Sentiment140 & Collection of tweets from X & \cite{Go2009SentimentCU} \\
     Stack Overflow Q\&A & Data collected from Stack Overflow & \cite{StackOverflow2019} \\
     TLDR News & Daily tech newsletter collection & \cite{belveze2022tldrnews} \\
     TREC & Q\&A dataset & \cite{voorhees2000trec} \\
     Wikicorpus & Wikipedia in English, Spanish, and Catalan & \cite{reese-etal-2010-wikicorpus} \\
     Wikipedia & XML text dump of English Wikipedia & \cite{Mahoney2009wikipedia} \\
     Wikitext-103 & Wikipedia articles & \cite{merity2016pointer} \\
     Wikitext-2 & Wikipedia articles & \cite{merity2016pointer} \\
     DBPedia & Structured Wikipedia content & \cite{zhang2016characterlevel} \\
     XSum & Summaries of news articles & \cite{narayan-etal-2018-dont} \\
     zsRE & Question-answer pairs from Wikipedia & \cite{levy2017zsRE} \\
     \hline
    \end{tabular}
    \caption{Datasets commonly used by works in this paper}
    \label{tab:datasets}
\end{center}
\end{table}

There are several relevant pieces of legislation regarding the privacy rights of individuals and the privacy requirements for model developers. The most prominent of these is the European Union's General Data Privacy Regulation (GDPR). Given that this legislation captures a very large population, the laws in the GDPR are a significant source of motivation for many of the works referenced in this paper -- in particular the ``Right to be Forgotten'' provision of GDPR was the motivation behind machine unlearning (Section~\ref{sec:unlearning}). A table of the ones mentioned in this survey are listed in Table~\ref{tab:legislation}.

\begin{table}[!ht]
\begin{center}
    \begin{tabular}{ |p{5cm}|p{1.5cm}|p{2.5cm}|p{2cm}|}
     \hline
     \hfil \textbf{Act} &\hfil \textbf{Acronym} &\hfil \textbf{Jurisdiction} & \textbf{Year} \\
     \hline
     General Data Privacy Regulation & GDPR & European Union & 2018 \\
     California Consumer Privacy Act & CCPA & California & 2018\\
     Consumer Privacy Protection Act & CPPA & Canada & \textit{In progress}\\
     New York Privacy Act & NYPA & New York & \textit{In progress}\\
     \hline
    \end{tabular}
    \caption{Acronyms for privacy legislation discussed in this survey along with the respective jurisdictions and the year the law was enacted}
    \label{tab:legislation}
\end{center}
\end{table}

\section{Memorization}
\label{sec:memorization}
If we define memorization broadly, all machine learning models will naturally exhibit memorization to some degree. Training a model necessitates observing training data and recalling that past information to make predictions. In the context of LLMs however, we want the models to generate content that semantically makes sense, while maintaining a degree of novelty. Initial studies on memorization have established that language models are capable of memorizing a significant fraction their training data, beginning with \cite{carlini2019secret}, who study LSTMs, and continuing with \cite{carlini2021extracting} who were able to extract $600$ memorized examples from GPT-2, and \cite{carlini2023quantifying} who study larger GPT-Neo models. Memorization is in itself, not necessarily an undesirable property of a machine learning model. In fact, recent work has show that a certain amount of \emph{memorization} may be \emph{required} for sufficiently accurate learning in certain natural classes of problems. \cite{feldman2020does} first proposes a formal model to study this phenomenon, and shows that when the distribution of sub-populations in the training data is ``long-tailed,'' some amount of label memorization is required to achieve good generalization error. This has implications for next-token prediction on text data, which due to the variety and number of unique sentences may also be ``long-tailed'' in this sense. \cite{brown2021irrelevant} follow up on this work, and show that in similar natural settings the model may be forced to memorize training data that may even be extraneous to the prediction task, in order to generalize well. That being said, there are evident types of memorization that clearly pose a privacy risk given the presence of sensitive information in the training set, and serve no obvious purpose for generalization. \cite{carlini2019secret} call this type of memorization ``unintended memorization,'' and show it occurs by inserting out of distribution samples called \emph{canaries} during training, and demonstrating they can be extracted. \cite{brown2021irrelevant} formalizes this notion mathematically, by defining it as the mutual information between the generative model and the training data, \emph{conditional} on the true distribution. For example, if a person named John Doe's private information was present in the training data of a generative model, it could respond to the prompt ``My name is John Doe and my phone number and address are...'' with John Doe's actual personal information \cite{carlini2021extracting}. Or the model may output copyrighted data that is contained many times in the training set verbatim, for example the first act of Hamlet. More examples of these risks are discussed in Section~\ref{sec:mi_attacks} and Section~\ref{sec:copyright} later on in this survey. On the other hand, some types of ``memorization'' are desired; for example when responding to the prompt ``What is the capital city of the state of Rhode Island,'' the desired output is, of course, ``Providence.'' Teasing apart these scenarios, and replacing intuition with mathematical definition is the content of the next subsection. 

Public discourse has generally focused on ways memorization in LLMs could affect peoples' daily lives. On the more humorous end of the spectrum, Tech Times has written about how ChatGPT memorized the Harry Potter books \cite{Davis_2023}. Other tech magazines and blogs have taken a grimmer view, discussing how LLMs can breach our personal privacy and how this could negatively affect businesses \cite{Heikkila2022gpt3} \cite{scrtlabs_2023}. In July, 2023, Federal Trade Commissioner Lina Khan remarked that one of the major concerns behind the FTC's investigation into OpenAI were ``reports of peoples sensitive information showing up'' \cite{nytopenai}.

\subsection{Preliminaries}
\label{subsec:mem_definitions}
Existing definitions of memorization that have been applied to language models can be organized into three categories: \textbf{eidetic memorization}, \textbf{exposure}, and \textbf{counterfactual memorization}. Additionally, there are two different approaches to determining whether two sequences, one from the training data, the other generated by a language model, are identical. We will discuss the latter first:

\textbf{Exact Matching:} In exact matching, two sequences $s_1$ and $s_2$ (whether they are words, numbers, or full essays) are considered to be identical if and only if they exactly match. This is the most straightforward and computationally efficient approach. Therefore, it is the approach used by almost all studies in the memorization space. 

\textbf{Approximate Matching:} Formalized by \cite{lee2022deduplicating}, two sequences $s_1$ and $s_2$ are considered to be identical if they are within a specified edit distance of each other. While this definition is less frequently used, there are strong arguments for using it (or variants of it) for future research. In \cite{ippolito2023preventing}, they evaded memorization detectors by adding a simple ``style transfer'' layer at the very end of their language model. A style transfer could consist of something as simple as putting two spaces between each outputted word or translating the output to a different language and back. Using approximate matching would allow for work in the memorization space to generalize to situations like this.

Due to the infrequent use of approximate matching, any future reference in this section to identical sequences will assume the use of the exact matching definition, unless otherwise specified. We now define $3$ types of memorization proposed in the literature.

\textbf{Eidetic Memorization:} The first class of definitions measure \emph{eidetic memorization}, first defined by \cite{carlini2021extracting} and used exactly or in slight variation by \cite{kandpal2022deduplicating, lee2022deduplicating, carlini2023quantifying} and \cite{tirumala2022memorization}. This colloquially refers to ``photographic memory'', the ability to recall information after seeing it only once. Eidetic memorization focuses on interactions where the model generates an output $s$ as the most likely continuation when prompted with a prefix token $c$. We begin with the definition of \emph{extractability}, which is a base form of eidetic memorization:

\begin{definition}
\label{def:extractability}
  (Extractability), \cite{carlini2021extracting}) A sequence $s$ of length $N$ is extractable from a model $h$ if there exists a prefix $c$ such that:
  $$s \leftarrow \argmax_{s': |s'|=N}h(s' | c)$$
\end{definition}

e.g. the email address ``alice@wonderland.com'' is extractable if prompting the model with ``Their email address is...'' (or any other prompt) yields ``alice@wonderland.com'' as the most probable output.

For large $N$, computing the argmax in Definition~\ref{def:extractability} is intractable, so it can be replaced by an appropriate sampling strategy. Using this framework is very realistic as most publicly released LLMs only allow black-box access \cite{bubeck2023gpt4}, and so an attacker might only have access to the sampled outputs. 

This definition leads us to a very straightforward test formulated in \cite{carlini2023quantifying} called the \emph{extractability test}. In this test, a large number of sequences of length $l$ are sampled from the training dataset. For each sequence, the model is prompted with the first $l-j$ tokens and is reported as ``extractable'' if the model outputs the next $j$ tokens exactly. This success rate is then averaged over all the samples to give a measure of how much the model memorizes. Prompting the model with the first $l-j$ tokens and seeing if the next $j$ tokens are recovered is also called \textbf{discoverable memorization} \cite{nasr2023scalable}

While this test is very straightforward to implement, it does not factor in how common a given extracted training example is. Ideally, common sequences would not be penalized for being memorized as much as uncommon sequences, since if a sequence appears many times in the training data it is unlikely to be private. \emph{$k$-eidetic memorization} builds on this idea:

\begin{definition}
\label{def:k_eidetic}
  (k-Eidetic Memorization, \cite{carlini2021extracting}) A sequence s is $k$-eidetic memorized (for $k \geq 1$) by an LM $h_\theta$ if $s$ is extractable from $h_\theta$ and $s$ appears in at most $k$ examples in $\dtrain : \lvert \{ d \in \dtrain : s \subseteq d\} \rvert \leq k$.
\end{definition}

Under this definition, the severity of the memorization harm is directly tied to the number of times the sequence appears in the training data. Commonly used sequences will have a large $k$, while private information will have low $k$. Since it can be defined and measured in an intuitive way, $k$-eidetic memorization a commonly used definition for many works on language model memorization. 

\textbf{Exposure:} Another common way to measure memorization is via \emph{exposure}. Developed by \cite{carlini2019secret}, this compares the probability the model outputs the sequence $s$ with the probability of outputting very similar sequences $s'$. We begin with the definition of \emph{log-perplexity}:

\begin{definition}
    \label{def:log_perplexity}
    (Log-perplexity, \cite{carlini2019secret}) The \textbf{log perplexity} of a sequence $s=(s1,...,s_n)$ is given by:
    \begin{align*}
        \logperp(s) &= -\log_2 \text{Pr}(s_1 \ldots s_n | h_\theta) \\
        &=\sum_{i=1}^n \left( -\log_2 \text{Pr}(s_i | h_\theta(s_1 \ldots s_{i-1})) \right)
    \end{align*}
\end{definition}

Where $\text{Pr}(s_i| h_\theta(s_1 \ldots s_{i-1}))$ is the probability of outputting $s_i$ given the model outputs of $(s_1,...s_{i-1}$ Log-perplexity essentially measures how surprised the model is to see a given sequence, so a lower log-perplexity means that sequence is more likely to have been generated by the model. Given that the log-perplexity values are dataset and model-dependent, we cannot simply define a threshold of ``high'' or ``low'' log-perplexity values that indicate a sequence is memorized. Instead, this value should be compared to similar sequences that the model was not trained on to normalize for these differences. This leads us to the definition of \emph{rank}.

Let $s$ be a sequence and $\mathcal{R}$ be the space of sequences similar to $s$ but with some random perturbations. For example, $s$ may be the sequence ``the random number is $123456789$'' and $\mathcal{R}$ is the space of all sequences of the format ``the random number is $\circ \circ \circ \circ \circ \circ \circ \circ \circ$''. If the $h$ and $s$ are susceptible to memorization, then the log-perplexity for $s$ will be lower than the log-perplexity for any similar sequence $r \in \mathcal{R}$. Rank is then defined as:

\begin{definition}
    \label{def:rank}
    (Rank, \cite{carlini2019secret}) The rank of a sequence $s$ is:
    \begin{align*}
        \text{rank}_{\theta}(s) = \lvert \{ r' \in \mathcal{R}: \logperp (r) \leq  \logperp (s) \} \rvert
    \end{align*}
\end{definition}

This is a useful definition for discussing the memorization of secret information. However, rank is computationally expensive as it requires computing log-perplexity of all possible values $r \in \mathcal{R}$. In practice, the metric of \emph{exposure} is used instead, which can be efficiently approximated:

\begin{definition}
    \label{def:exposure}
    (Exposure, \cite{carlini2019secret}) Given a sequence $s$, model parameters $\theta$, and random space of similar sequences $\mathcal{R}$, the exposure is:
    \begin{align*}
        \text{exposure}_\theta (s) &= \log_2 \lvert \mathcal{R} \rvert - \log_2 \left( \text{rank}_\theta (s) \right)\\
        &= -\log_2 \text{Pr}_{r \in \mathcal{R}} \big[ \logperp (r) \leq \logperp (s) \big]
    \end{align*}
    Then choosing some smaller space $\mathcal{S} \subset \mathcal{R}$ such that $|\mathcal{S}| \ll |\mathcal{R}|$, we can approximately compute exposure as:
    \begin{align*}
        \approx -\log_2 \text{Pr}_{r \in \mathcal{S}} \big[ \logperp (r) \leq \logperp (s) \big]
    \end{align*}
\end{definition}

The higher the exposure value, the more that given sequence is memorized by the model. This definition was formulated with the purpose of measuring how much a model memorizes. Called the \emph{canary extraction test} \cite{carlini2019secret}, this test inserts a ``canary'' $s$ into the training dataset with a format such as the above example: ``the random number is $123456789$''. This test can then effectively detemine whether a given model is susceptible to memorization by computing the exposure with respect to a space $\mathcal{S}$ of similar sequences.

\textbf{Counterfactual Memorization:} The third class of definitions measure memorization of a sequence $s$ by comparing the probability of generating the given output for two models, one trained with $s$ in the training set, and one without. These are inspired by the concept of differential privacy which we discuss in Section~\ref{sec:private_llms}. Let $A$ be a training algorithm that maps $\dtrain$ to model $h$ and $\ell(h,\cdot)$ be the loss of $h$ on a given example. We first define \emph{counterfactual influence}, which identifies which examples have a large impact on the prediction of a specified output:

\begin{definition}
    \label{def:cf_inf}
    (Counterfactual Influence, \cite{zhang2021counterfactual}) Given a sequence $s \in \dtrain$, the counterfactual influence of $s$ on another data point $s'$ is given by:
    $$infl(s\rightarrow s') = \mathbb{E}_{X \subset \dtrain, s\in X}[\ell(A(X),s')] - \mathbb{E}_{X' \subset \dtrain, s\notin X'}[\ell(A(X),s')]$$
\end{definition}

Where $X$ and $X'$ are subsets of training examples sampled from $\dtrain$. The expectation is taken with respect to the random sampling of $X$ and $X'$, as well as the randomness in the training algorithm $A$. The expectations in Definition~\ref{def:cf_inf} are estimated via training $m$ models on random subsets $X \sim D$ with fixed size and average the values of $\ell(A(X), s')$.

We then move on to the definition of \emph{counterfactual memorization}. First defined in \cite{feldman2020does}, it is the influence of a sample $s$ on itself: 

\begin{definition}
    \label{def:cf_mem}
    (Counterfactual Memorization, \cite{zhang2021counterfactual}) Given a sequence $s \in \dtrain$, the counterfactual memorization is given by:
    $$mem(s) = infl(s \rightarrow s) = \mathbb{E}_{X \subset \dtrain, s\in X}[\ell(A(X),s)] - \mathbb{E}_{X' \subset \dtrain, s\notin X'}[\ell(A(X'),s)]$$
\end{definition}

Using this definition, we get a continuous value for how strongly memorized a given sequence $s$ is for the given model, with $0$ representing no memorization and $1$ representing perfect memorization. Counterfactual memorization is advantageous because it does not bias towards common sequences. However, it is significantly more computationally expensive than the prior definitions because multiple models need to be trained to approximate the expectations. There are other differential privacy inspired definitions that are more relevant to developing privacy-preserving LLMs, and are included in Section~\ref{sec:private_llms}. 

\subsubsection{Additional Definitions}

\textbf{Entity Memorization}: When measuring language model memorization, using counterfactual memorization can be computationally expensive while eidetic memorization requires the adversary to have access to the original training data to prompt the prefixes. \cite{zhou2023quantifying} defined another memorization measure, known as \emph{entity memorization}.

\begin{definition}
\label{def:entity_memorization}
    (Entity Memorization, \cite{zhou2023quantifying}) A training string $s$ contains $m$ entities $M$ and can be uniquely identified by any $n$ entities $N$ of them. $n \in (0,m]$. A prompt $c$ contains the $n$ entities that could identify $s$, and $c$ expects an entity $E \in (M-N)$. Prompted the language model $h$ with $c$. If the expected entity $E$ is a substring in the output $h(c)$, the process is referred to as Entity Memorization.
\end{definition}

\begin{figure}[ht]
  \centering
  \includegraphics[width=.9\linewidth]{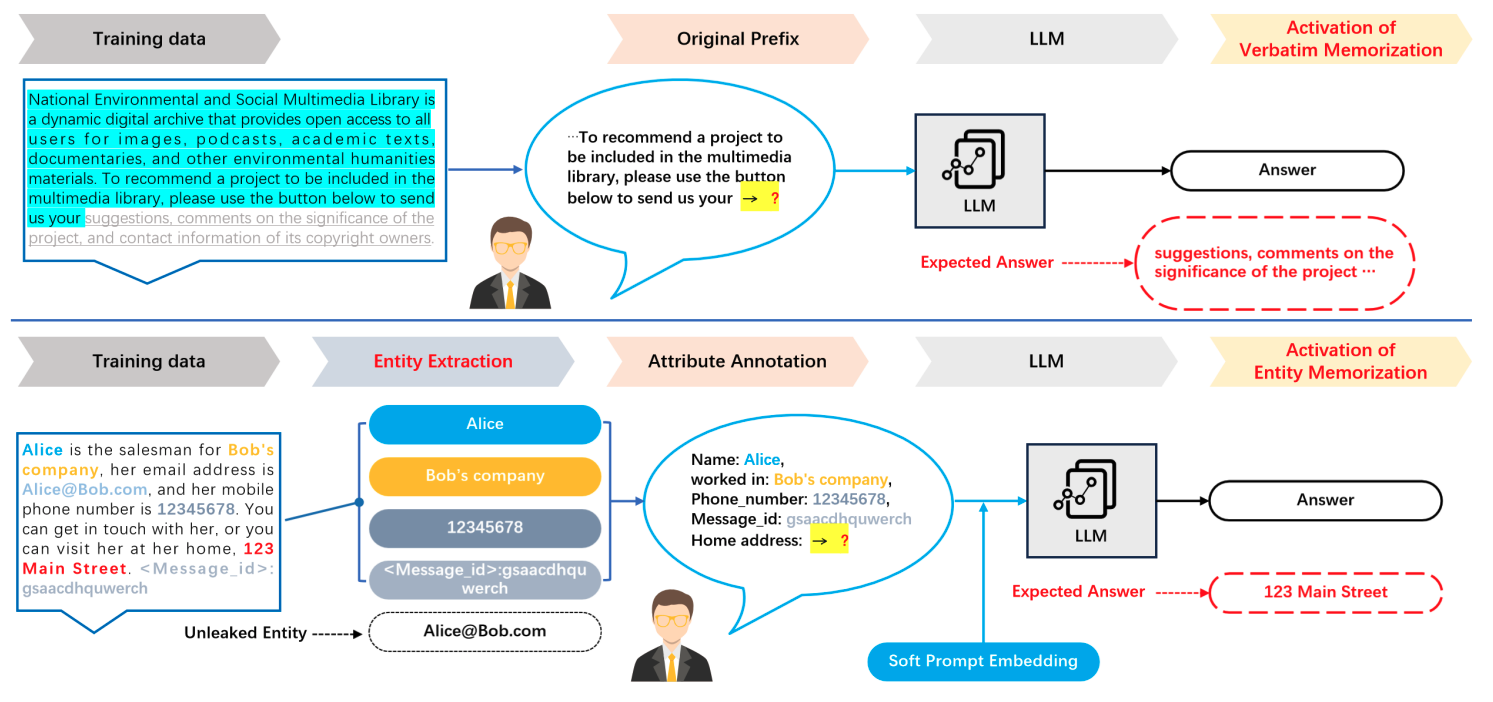}
  \caption{From \cite{zhou2023quantifying}. A comparison of typical verbatim memorization versus entity memorization (Definition~\ref{def:entity_memorization}). Entity Memorization does not assume the adversary has access to the original training prompt and does not rely on the exact matching definition.}
  \label{fig:zhou_extraction}
\end{figure}

Illustrated in Figure~\ref{fig:zhou_extraction}, this definition first identifies key entities in the training data, then uses soft prompt embedding methods to generate a new prompt for the LLM, with the goal to identify the remaining entities in the training data. Soft prompt embeddings are a collection of methods \cite{li2021prefix, lester2021power} that given a model and a starting prompt, optimize the prompt to minimize model loss. This approach is closer to real-world situations where an adversary may have access to key information pieces without full access to the original prompt and using only extracted entities is more flexible than verbatim memorization comparisons. However, there have not yet been many extensive studies using this definition.

\textbf{Forgetting}: Closely related to DP-inspired memorization is the concept of \emph{forgetting}, first formulated in \cite{tirumala2022memorization} but defined more generally in \cite{jagielski2023measuring}. This definition was created to measure how memorization changes throughout the training of a given model. Forgetting is measured by evaluating the success rate of a privacy attack:

\begin{definition}
\label{def:forgetting}
  (Forgetting, \cite{jagielski2023measuring}). A training algorithm $\mathcal{A}$ is said to $(\mathcal{Z}, \alpha, t)$-forget a training example $s$ if, $t$ steps after $s$ is last used in $\mathcal{A}$, a privacy attack $\mathcal{Z}$ achieves no higher than success rate $\alpha$.
\end{definition}

We delve more into what a potential privacy attack $\mathcal{Z}$ could be, and how to measure its attack success in Section~\ref{sec:mi_attacks}, but for now we observe that we can define a training example as $(\mathcal{Z}, \alpha)$-memorized if after training it is not forgotten relative to a privacy attack $\mathcal{Z}$, in that $\mathcal{Z}$ achieves a success rate higher than $\alpha$.

\subsection{Risk Factors for Memorization}
\label{subsec:mem_risk}
A number of existing works study the relationship between different characteristics of the training regime or sequence $s$ and the likelihood that $s$ is memorized by the trained model. We now survey these works, organizing the discussion by the characteristic studied. 

\subsubsection{Capacity of Model}
\label{subsec:mem_capacity}
\begin{figure}[ht]
  \centering
  \includegraphics[width=.9\linewidth]{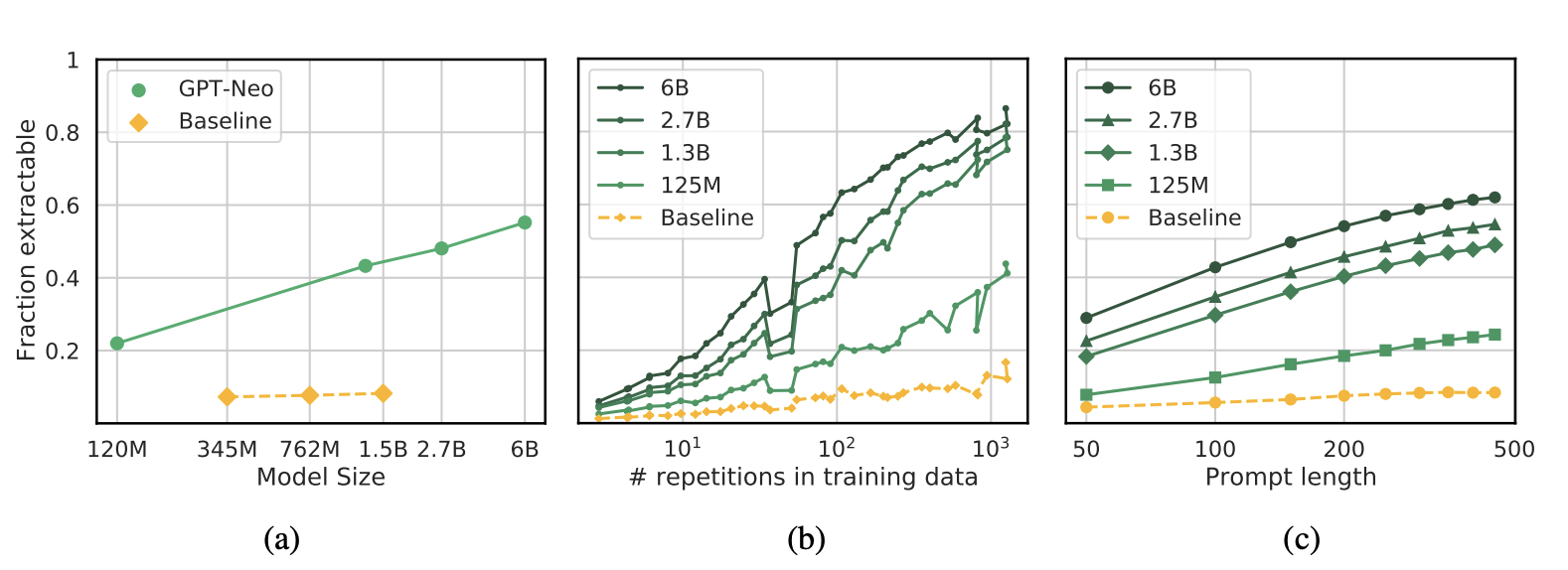}
  \caption{\cite{carlini2023quantifying} showing the fraction of the dataset that was memorized depending on a) the model size, b) how much of the data was duplicated, and c) the length of the prompt.}
  \label{fig:carlini_quantifying}
\end{figure}

The first significant factor that influences memorization is the size of $h$. Recent work indicates that given a fixed training dataset, the larger the size of the model, the greater the amount of memorization. Two studies, \cite{carlini2023quantifying} and \cite{tirumala2022memorization}, have studied this relationship, both using the extractability notion (\ref{def:extractability}) as a measure for memorization. \cite{carlini2023quantifying} used the GPT-Neo family of LLMs \cite{black_sid_GPTNeo, wang_mesh} and found that larger models memorize more on a log-linear scale; that is, a ten-fold increase in the size of the model corresponded to an increase in memorization by $19\%$. These results can be seen in Figure~\ref{fig:carlini_quantifying}(a). This effect is consistent even when other factors like the amounts of data duplication and the length of the token used to prompt the model are varied. \cite{tirumala2022memorization} found that not only do larger models memorize a greater fraction of the training data, they also memorize that data \emph{much faster}. That is, the memorization effects appear after a relatively small amount of training iterations. Whether these trends would hold under alternative definitions of memorization, for example under counterfactual memorization, is yet to be seen.  \cite{tirumala2022memorization} used the Transformer language model architectures \cite{artetxe2022efficient, zhang2022opt} at $125M, 1.3B, 2.7B$ and $13B$ model parameter configurations, and used Wikitext-103 \cite{merity2016pointer} and RoBERTa \cite{liu2019roberta} as training datasets.

A related result was found by \cite{Thomas2020InvestigatingTI} who looked at memorization of an LSTM model trained with the Penn Treebank dataset \cite{marcus1993penntreebank} using pre-trained word embeddings with varying vector dimensions, GloVe \cite{pennington2014glove} at $d=100$ and $d=300$, ELMo \cite{peters2018ELMo} at $d=1024$, and BERT \cite{devlin-etal-2019-bert} at $d=768$ and $d=1024$. For this experiment, \cite{Thomas2020InvestigatingTI} inserted canaries then measured their exposure from the LSTM (Definition~\ref{def:exposure}). They found that the canaries in LSTMs trained using the higher dimension models also had higher exposure. 

These results are unfortunate practically as higher dimension embedding models also typically yield better performance. \cite{kandpal2022deduplicating} studied the effects of performance, model size, and memorization for Transformer models. They measured the outcomes for a $117$M and $345$M parameter model from Mistral project \cite{StanfordCRFM2021mistral} and $1.5$B parameter model from West \cite{west2021reflective} trained on OpenWebText \cite{Gokaslan2019OpenWeb}. They also looked at a $1.5$B model \cite{lee2022deduplicating} trained on C4 \cite{raffel2020exploring}. Across these, they found that increasing the size of the models increases the memorization along with decreasing the training loss and hypothesize this assigning of higher likelihoods to training examples is related to the memorization effects. However, \cite{carlini2023quantifying} showed that memorization is due to model size and does not correlate with model performance. To test this, \cite{carlini2023quantifying} compared GPT-2 and GPT-Neo models with the same capacity and found that while the more advanced model (GPT-Neo) performed better, it did not memorize more and memorization instead was linked to the model size.

\begin{figure}[ht]
  \centering
  \includegraphics[width=.6\linewidth]{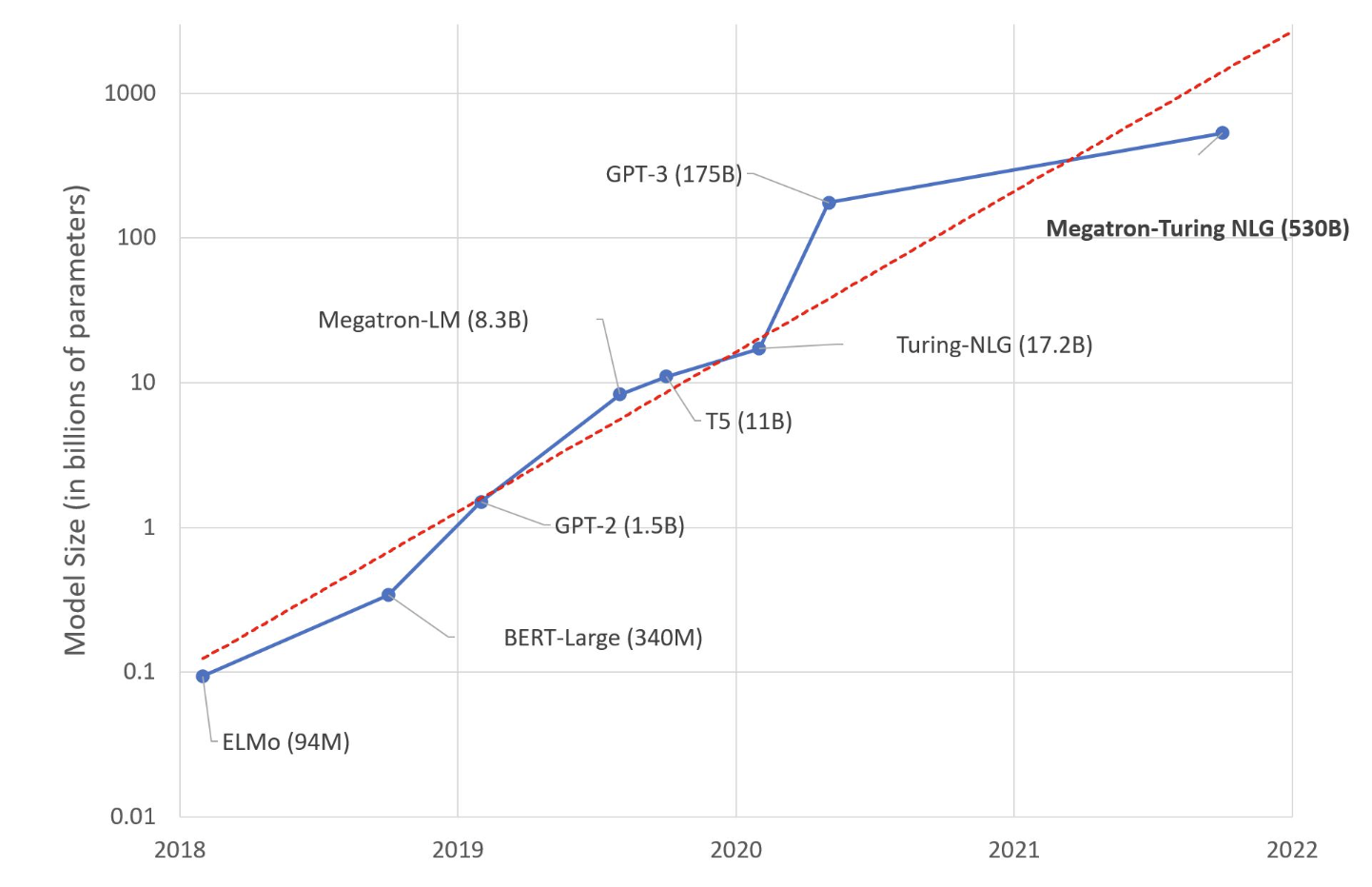}
  \caption{From HuggingFace blog \cite{Simon2021moore} showing the rapid growth of LLM capacity over the last few years. They conjecture that this may be following Moore's Law.}
  \label{fig:model_size_moore}
\end{figure}

Some hypothesize this effect to the high capacity of neural network models. \cite{zhang2017understanding} found that most neural network architectures are rich enough to memorize the majority of datasets. Even as modern language datasets increase in size, the growing capacities of these models vastly outstrip these size increases. As shown in Figure~\ref{fig:model_size_moore}, LLM size has been increasing around $10\times$ every year for the last few years and shows no sign of slowing down.

\subsubsection{Size of Dataset}
\label{subsec:mem_size}

The effect of dataset size on memorization has been explored a little bit as well. \cite{jagielski2023measuring} explored how different factors affect a model's ability to ``forget'' (Definition~\ref{def:forgetting}). Their experiment on LMs used canary extraction as the privacy attack on a decoder-only, $110M$ parameter, Transformer-based langauge model \cite{raffel2020exploring} trained over C4 \cite{dodge2021documenting}, which was deduplicated with MinHash \cite{lee2022deduplicating}. They found that over the course of training, the success of the canary extraction attack slowly decreased for canaries seen early in the training process and therefore in larger training datasets, models are less likely to memorize points.

This outcome, however, is seemingly contradicted by \cite{biderman2023emergent}, which in part, looked at how memorization generalized from partially trained LMs to the fully trained LM. This experiment used the Pythia \cite{biderman2023pythia} LLM suite trained on the Pile \cite{gao2020pile, biderman2022datasheet}. In this experiment, the ``ground truth'' was labeled by the $k$-eidetic memorization of the fully trained model while ``predictions'' were seen as the memorization of the partially trained model, allowing for traditional classification metrics. While the recall rate was low, the precision rate was very high at early checkpoints. This means that points memorized at a partially trained point were still memorized by the fully trained model and were not forgotten over time. One significant factor distinguishing these two experiments was that \cite{jagielski2023measuring} deduplicated their dataset which has been shown to reduce memorization. 

\subsubsection{Data Duplication}
\label{subsec:dedup}
Another primary cause of memorization is the duplication of training data, when a given sequence $s$ occurs multiple times in the training data and so is seen multiple times during training. Most LLMs are trained on massive datasets that include large amounts of literature or information scraped from the internet. As a result, strings that are a part of popular text (e.g. Martin Luther King's ``I Have a Dream'' speech) will appear much more frequently than other strings. This is on top of common phrases or sayings appearing across multiple different text sources. \cite{lee2022deduplicating} found that data duplication in large web datasets is a power law: a small proportion of the data is duplicated a very high number of times and the frequency of data duplication decays extremely quickly. 

\begin{figure}[ht]
  \centering
  \includegraphics[width=.3\linewidth]{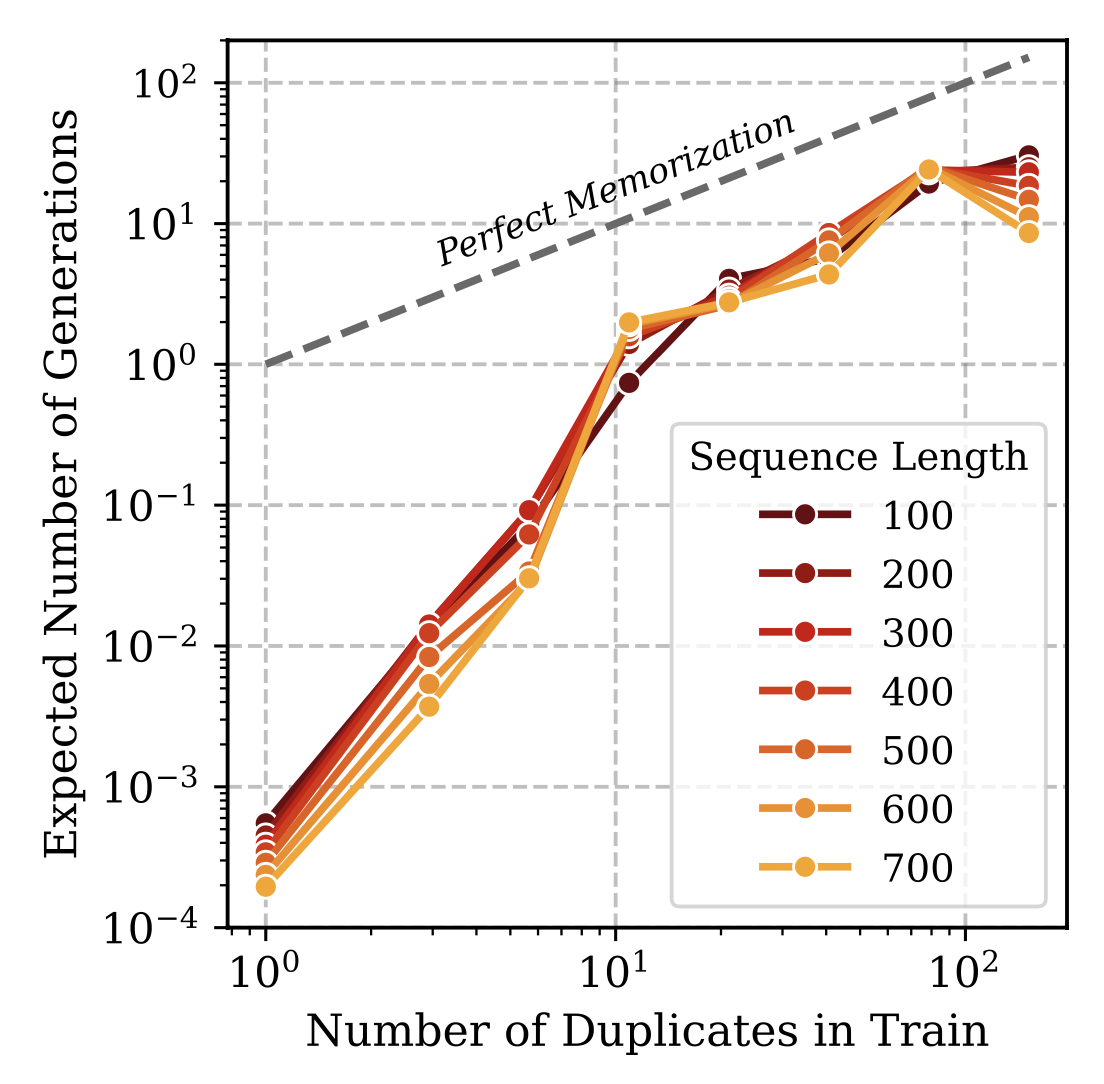}
  \caption{\cite{kandpal2022deduplicating} showing the expected number of times a model will generate a given sequence given the number of times that sequence is duplicated in the text. This is also somewhat dependent on the length of the outputted sequence, however the pattern remains consistent.}
  \label{fig:kandpal_duplicates}
\end{figure}

The impact of duplicated data on memorization is large and immediate. \cite{lee2022deduplicating} found that a model trained on a deduplicated dataset outputs memorized text ten times less frequently than the baseline (in addition to other performance benefits). Meanwhile, \cite{kandpal2022deduplicating} found that a given sequence present $10$ times in the dataset is on average generated $1000$ times more frequently than a sequence that is present only once. In Figure~\ref{fig:kandpal_duplicates} (and also Figure~\ref{fig:carlini_quantifying}(b)), we can see how the increase in training data repetition increases that degree to which memorization occurs. Kandpal also found that the sampling method used by language models impacts the expected memorization. Methods such as top-$k$ or beam-search which output sequences with higher likelihood under the model are also more likely to generate verbatim training samples. They found that all sampling methods generate significantly fewer verbatim examples when the datasets are deduplicated (see Section~\ref{subsec:mem_mitigation}).

While data duplication is a significant risk factor, it is not solely responsible for memorization. \cite{carlini2023quantifying, kandpal2022deduplicating}, and \cite{lee2022deduplicating} all found that memorization still occurs even when there is little to no duplication in the training set as shown in. Furthermore, all three of those works found that result when only considering two sequences as identical if they were exact matches. As discussed by \cite{ippolito2023preventing}, only considering verbatim matches ignores other possible memorization risks. It is likely that the effects identified in those works would be even greater if approximate matches (such as those defined in \cite{lee2022deduplicating}) were considered.

\subsubsection{Prompt Length and Type}

Another factor influencing memorization is the nature of the tokens used to prompt the language model and to be generated by the model. For the purposes of this section, $l$ is the number of words used to prompt the language model and $n$ is the number of words generated by the model. \cite{mccoy2021language} looked at the novelty of $n$-grams generated across four different language models: an LSTM, a Transformer, and Transformer-XL trained on Wikitext-103 \cite{merity2016pointer} and GPT-2 trained on the WebText dataset. They defined an $n$-gram as ``novel'' if it was not an exact match to any $n$-gram in $\dtrain$. \cite{mccoy2021language} found that as $n$ increased, the number of novel $n$-grams generated increased, which makes sense because the number of possible outputs grows exponentially in $n$. For $n \geq 5$, the majority of $n$-grams generated were novel across all models and datasets. \cite{kandpal2022deduplicating} explored much higher values of $n$, with values ranging from $100$ to $700$. At these higher levels, they still found that larger $n$ reduced the amount of memorization (see Figure~\ref{fig:kandpal_duplicates}). On the other hand, as $l$ increases, the memorization of the model also increases, given a constant $n$. This was studied in \cite{carlini2023quantifying} and can be seen in Figure~\ref{fig:carlini_quantifying}(c). 

Looking more specifically at the type of tokens used to prompt the model, there are certain characteristics that influence the levels of memorization. \cite{tirumala2022memorization} found that while all parts of speech are eventually memorized throughout the training process, nouns, proper nouns, and numbers are memorized much faster than verbs or adjectives. The eventual amount of memorization remains the same after sufficient number of training epochs. \cite{kharitonov2021bpe} investigated the impact of a tokenizer on memorization, as measured by the success of MIAs. They find that the size of the subword vocabulary used in byte-pair encoding (BPE) has a large impact on memorization; specifically language models with larger subword vocabularies reproduce the training data more often. They hypothesize that this effect is caused by reduction in the sequence length that happens as the BPE vocabulary grows, and these shorter sequences are easier for the model to memorize during training.

\subsubsection{Time of Memorization}
\begin{figure}[ht]
  \centering
  \includegraphics[width=.3\linewidth]{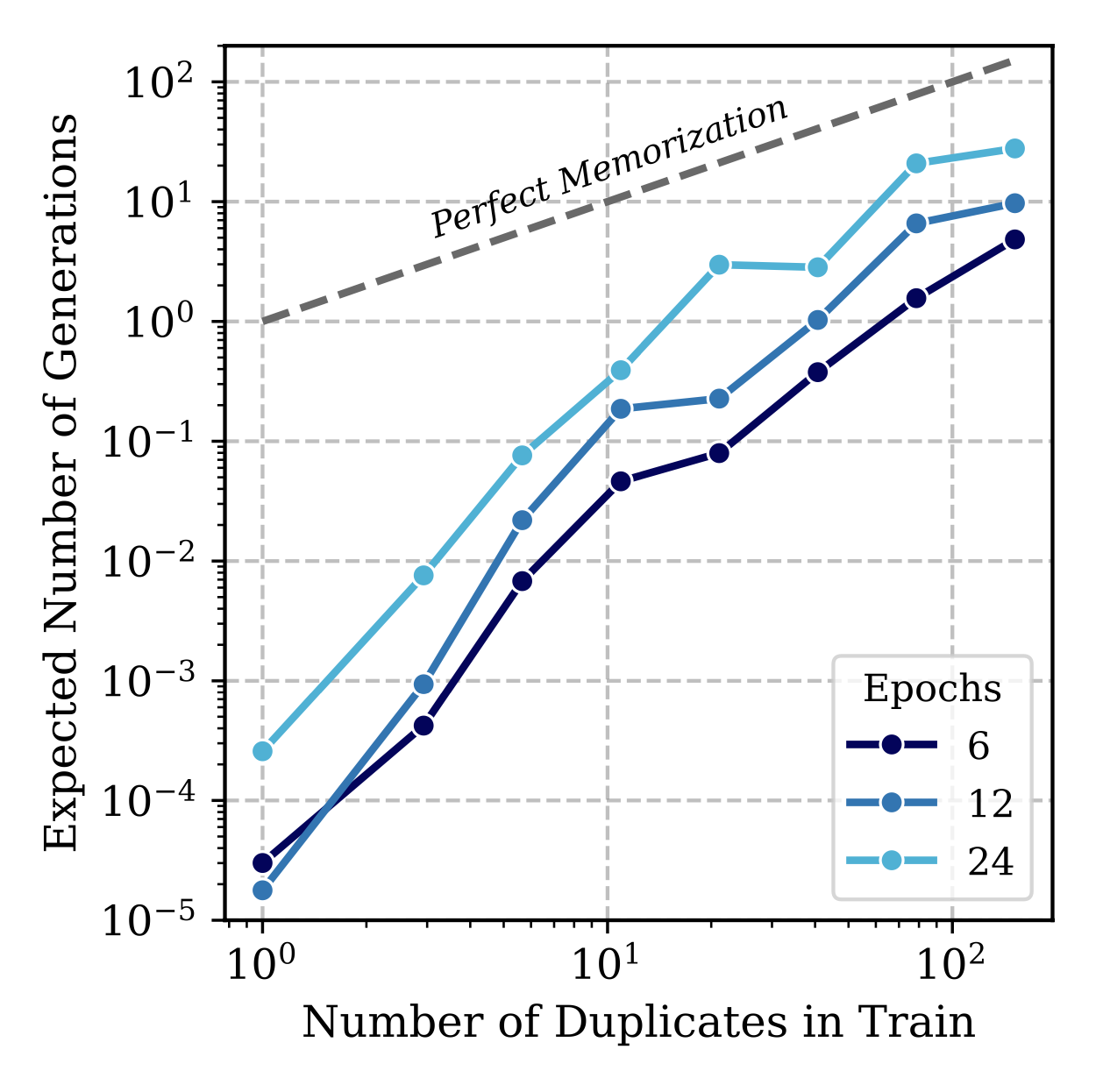}
  \caption{\cite{kandpal2022deduplicating} showing the fraction of the dataset that was memorized depending on the number of duplicates in the training data and also on the number of training epochs. We can see that more training epochs leads to higher rates of memorization.}
  \label{fig:kandpal_epochs}
\end{figure}

Lastly, memorization is effected by the training time. \cite{privacy_amp} shows that when optimizing a convex loss with SGD, if a small amount of Gaussian noise is added to the result of each gradient computation, then they can prove that earlier points witness stronger privacy protections over time. While this work only holds for noisy iterations on convex losses, and so the theoretical results would not apply to training language models, \cite{jagielski2023measuring} investigates whether the same forgetting phenomenon occurs empirically. They find that examples seen early on in the process are more likely to be forgotten than examples seen later, as measured by the success of a membership inference attack. This is somewhat related to the size of the dataset as well (see Section~\ref{subsec:mem_size}), since larger datasets will give more time for the early examples to be forgotten. \cite{kandpal2022deduplicating} ran more exact experiments on this, finding that a model using twice as many training epochs has verbatim output three times as often (see Figure~\ref{fig:kandpal_epochs}).

\subsection{Mitigation Techniques}
\label{subsec:mem_mitigation}
We now discuss attempts to mitigate memorization in language models, while preserving performance. Two of the approaches, differentially private training and federated learning are mentioned briefly, but are discussed more thoroughly in Section~\ref{sec:private_llms}.

\textbf{Data De-Duplication:} The most straightforward and immediate remedy to memorization is the de-duplication of training datasets. As noted in Subsection~\ref{subsec:dedup}, data duplication is a significant factor in how much a model memorizes. \cite{kandpal2022deduplicating} found that a sequence duplicated $10\times$ can be generated around $1000\times$ more often. \cite{lee2022deduplicating} showed that their de-duplication tool reduced the generation of memorized text by about $10\times$. \cite{lee2022deduplicating} has publicly released their tool and it has the option to specify what qualifies as a match, either ``exact'' or ``approximate'' as discussed in Section~\ref{subsec:mem_definitions}.

\textbf{Early Detection:} From a practical perspective, it would be very useful to have early identification during training of when a model is at high risk of memorization. This would enable the practitioner to do things like discard memorized points and revert to a checkpoint, or to stop the training process early and make adjustments. \cite{biderman2023emergent} studied rates of memorization throughout training and found that sequences memorized by a partially trained LLM were often still memorized by the fully trained LLM. Similarly, they found that sequences that a smaller LLM memorized were still memorized by the larger model. However, while the precision rate was high, the recall rate was very low. That is, the smaller and partially trained models were not able to identify most of the eventually memorized points. This limits the utility of stopping training early since not all memorized points could be adjusted for. However, this approach could be used to answer whether the model exhibits memorization without the cost of fully training a high capacity LLM.

\textbf{Differentially Private Training:} The most promising method in reducing memorization is the use of differentially-private based techniques due to its strong theoretical guarantees. \cite{carlini2019secret} found that using differentially private training was the only effective tool in completely eliminating memorization issues, however there was a drop in utility as well.  \cite{zhao2022provably} developed a method called Confidentially Redacted Training (CRT) which involved first de-duplicating the data, then splitting any sensitive data into a private dataset. The training process then alternates stochastic gradient descent (SGD) from the public dataset and differentially private SGD (see Algorithm~\ref{alg:dpsgd}) from the private dataset. The combination of de-duplication and CRT was theoretically and empirically shown to be effective at reducing memorization. For more information on differentially-private training, see Section~\ref{sec:private_llms}.

\textbf{Federated Learning:} Another technique that has been shown to decrease memorization is the use of Federated Learning. Federated learning was primarily developed as a framework for large scale distributed learning \cite{mcmahan2016communicationefficient}. Instead of data being stored in a centralized location, the data is stored across multiple sites (e.g. on users' personal devices). Then, instead of updates to the model being done by a SGD on a minibatch of data, the model updates with Federated Averaging: an average of the differences between the current model and the model obtained after several SGD steps on the local data of a user. \cite{thakkar2020understanding} used the canary testing method on a model trained in a federated learning setting and found that it memorized anywhere from a third to none of the canaries when compared to the equivalent centralized learning (dependent on parameters). It is hypothesized that clustering data by user, which happens by design in federated learning, plays a significant role in reducing memorization. Furthermore, both \cite{thakkar2020understanding} and \cite{ramaswamy2020training} have explored combining differentially private training in a federated learning setting. The combination of these two methods seems to significantly reduce memorization and is a promising solution for preventing this privacy concern in the future (see Section~\ref{sec:private_llms}). 


\textbf{LLM Editing:} Another approach to reducing memorization is to directly edit the neurons of the LLM that correspond to the memorization. \cite{chang2023localization} tested several localization methods that identify the set of LLM parameters responsible for storing certain knowledge. They found that their method known as \emph{Hard Concrete} was able to identify sets of neurons consisting of approximately $.5\%$ of the Pythia $6.9$B model where removing the neurons reduced the memorization of that example by over $50\%$. Other localization methods were able to achieve varying degrees of success but across all of them, targeted memorization reduction also resulted in a drop in performance. Despite this, the approach of editing LLMs to protect sensitive information is a promising direction for memorization mitigation. 

\textbf{Mitigation For Translation Models:} Neural machine translation (NMT) models are another application of LLMs for the purposes of language translation. Within this space, \cite{raunak2022finding} adapted the extractability notion (Definition~\ref{def:extractability}) of memorization to NMT models and showed that in addition to memorization occurring in NMT models, many neighbors of the inputs (i.e. inputs that differ in one word) output the same memorized results. Their proposed algorithm appends the input with a special character (e.g. "!"), then removes the character after translation. This approach results in $65\%$ of memorized translations returning a new, but still semantically correct, output.

\section{Privacy Attacks Against Language Models}
\label{sec:mi_attacks}
In this section, we review recent works on privacy attacks against language models. This recent work builds upon a long line of research in the machine learning community that studies privacy attacks against ML models, typically in the classification setting. We refer the reader to \cite{mia_survey} and \cite{dl_privacy_survey} for comprehensive surveys. Privacy attacks can largely be divided into two classes: membership inference attacks (MIAs), and training data extraction attacks (and attribute-inference attacks which we discuss in brief). MIAs are a less obviously severe privacy attack, where an adversary, given access to the model, is able to determine whether a given point was used to train that model with high accuracy. While this indicates that some information about the training point is ``leaking'' through the model, given that the adversary is still required to have access to a candidate point that may have been used to train the model in order to run the attack, it is more of a ``smoke signal'' than an outright privacy violation. Training data extraction on the other hand, is when given access to the model, an adversary can actually \emph{reconstruct} parts of the underlying training data, which may be sensitive or protected by law. 

When dealing with generative models however, MIAs have the potential to facilitate training data extraction attacks, which makes them of elevated importance. The key trick here is that the adversary can exploit \emph{memorization}: as discussed in Section~\ref{sec:memorization}, language models memorize swathes of their training data verbatim, and can be prompted to regurgitate these training examples. Armed with a MIA that can identify training samples from samples in the distribution, an adversary can then use the attack to find the generated points that are most likely to be training points, and ``extract'' those. We discuss this further in Section~\ref{subsec:extract}. 

We start by discussing the basics of MIAs, and then covering MIAs against language models, which are different from attacks on standard models in a few key ways. We then cover training data extraction from language models, and discuss mitigation of privacy attacks.

\subsection{Preliminaries}
\label{subsec:mia_preliminaries}

Membership inference attacks are based on the concept that models tend to perform better on examples that are in their training set than ones which are not. This means that model confidence is a natural test statistic for determining membership in $\dtrain$ and can be used for a MIA. The most straightforward and computationally inexpensive MIA is a \emph{threshold attack}, which was first implemented by \cite{yeom2018privacy}. This attack simply designates a threshold $\tau$ and compares a chosen test statistic (in most cases, loss) to that. An outline for this kind of attack is shown here:

Given a target model $h_{target}$, a threshold $\tau$, and some test statistic $l(h,z)$ that is computed by prompting $h$ with $z$, perform the following attack on a target data point $\mathbf{z}$:

\begin{enumerate}
    \item Query the model to get $h_{target}(\mathbf{z})$
    \item Classify $\mathbf{z}$ as \texttt{in} if $l(h_{target}, \mathbf{z}) < \tau$. Else, classify $\mathbf{z}$ as \texttt{out}.
\end{enumerate}

If, for example, the test statistic is loss, then this attack simply classifies the low loss points as \texttt{in} and high loss points as \texttt{out}. The choice of $\tau$ determines the trade-off between the precision and recall of the MIA. If $\tau$ is very high, then the MIA will have a high recall, but low precision rate in identifying members of $\dtrain$. Meanwhile if $\tau$ is low, then it will have a high precision rate, but low recall. This also enables statistics common in machine learning such as the AUC. Loss is typically the first choice based on the principle that models tend to overfit on training data. However, \cite{yeom2018privacy} found that overfitting, while sufficient, is not necessary for a MIA to be successful and other test statistics may instead be used for a successful MIA. This is especially relevant in the context of LLMs which have much larger training datasets and are thus trained using a single-pass SGD. This means they overfit less making loss-based threshold attacks less effective. A more commonly used test statistic is perplexity, which colloquially measures how surprised the model is to see a given sequence. This is very similar to $\logperp$, defined in the Memorization section (Definition~\ref{def:log_perplexity}).

\begin{definition}
    \label{def:perplexity}
    (Perplexity, \cite{carlini2021extracting}) Given a model $h$ with parameters $\theta$, the \textbf{perplexity} of a sequence $s=(s1,...,s_n)$ is given by:
    \begin{align*}
        \perplexity = \text{exp} \Big( -\frac{1}{n} \sum_{i=1}^n  \log_2 \text{Pr}(s_i | h_\theta(s_1 \ldots s_{i-1}) \Big)
    \end{align*}
\end{definition}

Threshold attacks tend to suffer from high false positive rates \cite{mattern2023membership} and low precision in identifying sensitive data \cite{carlini2021extracting}. One solution to this, first developed in the context of ML by \cite{murakonda2021quantifying}, is to use a likelihood ratio as the test statistic. In this approach, a reference model, $h_{reference}$, is trained using the same architecture and similar data as the target model, $h_{target}$. Then for a target data point $s$, the test statistic is $L(s) = \log(\frac{p_{reference}(s)}{p_{target}(s)})$, where $p_{\circ}(s)$ is the probability of outputting $s$. This reduces the bias towards common terminology and allows for better identification of sensitive data.

\begin{figure}[ht]
  \centering
  \includegraphics[width=.7\linewidth]{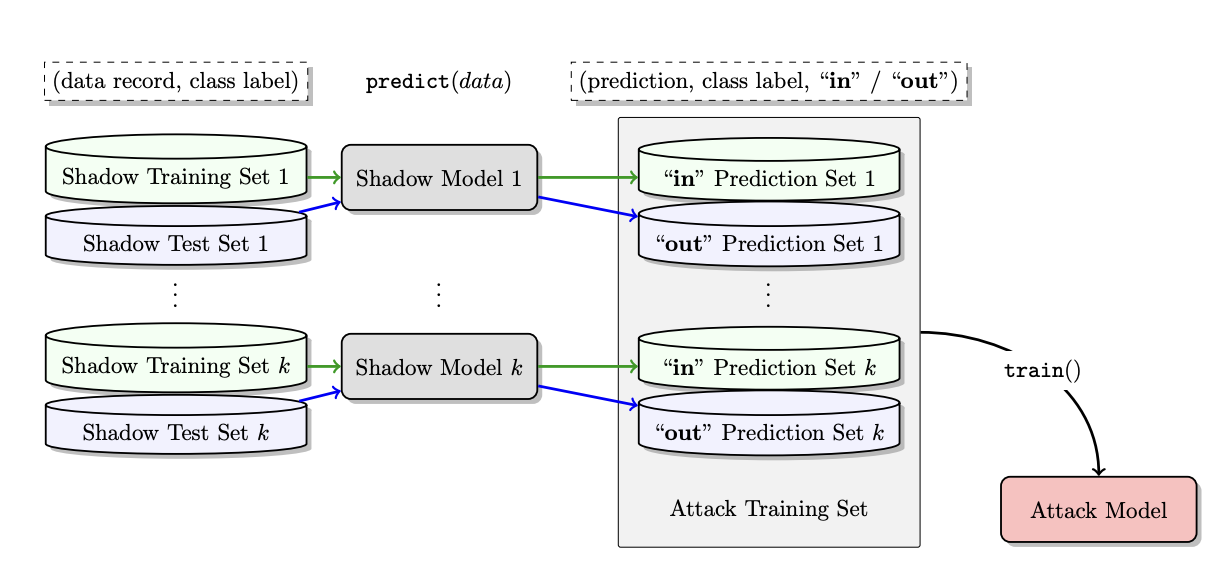}
  \caption{Shadow model framework from \cite{shokri2017membership}. Multiple shadow models are built using the same architecture and similar data as the target. Points from the shadow models' training sets are used to build the \texttt{in} prediction set while points from the shadow models' test sets are used to build the \texttt{out} prediction set. An attack model is then trained to distinguish the \texttt{in} and \texttt{out} labels.}
  \label{fig:shokri_framework}
\end{figure}

More advanced approaches, such as \cite{shokri2017membership, long2020pragmatic, sablayrolles2019whitebox, song2020systematic} and \cite{carlini2022first}, seek to approximate two distributions of the test statistic, one for a model which contains the target points $\dtrain$ and one which does not. It then trains an attack model to determine which distribution a target point $z$ belongs to. This approach necessitates the training of \emph{shadow models}, which are models of the same architecture as the target model trained on data similar to the target dataset. These shadow models are used to estimate the distributions for the \texttt{in} and \texttt{out} data points. This approach was first designed by \cite{shokri2017membership} and is illustrated in Figure~\ref{fig:shokri_framework}. \cite{carlini2022first} combines this approach with the reference model approach by conditioning the output of the target $h_{target}(z)$ on the two distributions and computing the ratio. The ratio is then used as the threshold statistic to determine $\dtrain$ membership. This is a leading MIA approach and is known as LiRA (Likelihood Ratio Attack). We discuss this further in Section~\ref{subsec:mia_shadowattacks}

These approaches can perform much better than a threshold attack, but it requires much more knowledge of the target model and dataset. Additionally, training shadow models can be computationally expensive since they must be of the same build as the target model. In the case of LLMs, which can be build using billions of parameters and data points, this is often prohibitively expensive.

\subsection{Membership Inference Attacks on Language Models}
\label{subsec:mi_llms}
In recent years there has been growing research on MIAs for language models. It has been found that LMs are fairly robust against simple probing methods. For example, \cite{lehman2021does} attacked a BERT model trained on the MIMIC-III \cite{Johnson2016mimiciii} clinical dataset by probing with prompts such as "John Doe is a 35 year old patient diagnosed with..." to try and extract health conditions. This method, and simple variants of it, were unable to extract any meaningful sensitive data. However, there still remain privacy risks when the MIAs described above are tailored towards LMs as we will discuss here.

\subsubsection{Threshold Attacks}
\label{subsec:mia_thresholdattacks}

To understand this first MIA, we must first discuss the process by which LMs are trained. To build $\dtrain$, the raw corpus of text data is first \emph{tokenized}, transforming the large dataset into discrete elements (e.g. tokenizing ``My name is John Doe'' into [``my'',``name'',``is'',``john'',``doe'']). Since there are a very large amount of possible words in the dictionary, the matrix representing this data would be very sparse. Thus \emph{embedding} models are used to transform these tokens into low-dimensional vectors that preserve semantic data. This is a critical first step for many LMs and has been very successful in a wide range of applications.

While most MIA research revolves around the LMs, \cite{song2020information} found that word and sentence embedding models are also vulnerable to privacy attacks. These embedding models take in sparse text data and transform it into lower dimensional vectors that maintain semantic meanings yet are easier for models to learn. For example, the word embedding models Word2Vec \cite{bojanowski2017word2vec} and FastText \cite{mikolov2013fasttext} are trained in an unsupervised manner where, given a sliding window of words $C = [w_b,...,w_0,...,w_e]$, they aim to predict $w_i$, given the center word $w_0$, compared to a \emph{negative} sample (i.e. a random sample of words not in $C$). The sentence embedding model studied, given sentence $s_a$ and following sentence $s_b$, is trained to maximize the probability of $s_b$ given $s_a$, compared to the negative sample of sentences. \cite{song2020information} assumes that the adversary has access to this embedding model $\Phi$. To perform the threshold attack on word embedding model $\Phi$, \cite{song2020information} first converts $C$ into its embedding $\Phi_{word}(C)$. They then compute a similarity score $\delta$ (e.g. cosine similarity) for each pair of vectors $\delta(\Phi_{word}(w_i),\Phi_{word}(w_j))$ in $C$. If the averaged similarity score is above a certain threshold $\tau$, it is considered part of the training data. The logic is that $\Phi_{word}(C)$ will have a higher average similarity score if $C \in \dtrain$ than if $C \notin dtrain$. See Algorithm~\ref{alg:word_mia} for the details. \cite{song2020information} also formulated an attack for sentence embedding models. For the sentence embedding MIA, \cite{song2020information} computes the similarity score $\delta(\Phi_{sentence}(s_a),\Phi_{sentence}(s_b))$ to perform the threshold attack. If the similarity score is above a certain threshold, the target pair is classified as \texttt{in}. See Algorithm~\ref{alg:sentence_mia} for details.

\cite{song2020information} performed this attack on three word embedding models: Word2Vec \cite{bojanowski2017word2vec}, FastText \cite{mikolov2013fasttext}, and GloVe \cite{pennington2014glove} which were all trained on the Wikipedia articles corpus \cite{Mahoney2009wikipedia}. They also attacked a sentence embedding model, the dual-encoder framework used in \cite{cer2018universal, chidambaram2019learning, henderson2017efficient, logeswaran2018efficient, lowe2015ubuntu, reimers2019sentence} and \cite{yang2018learning} trained on BookCorpus \cite{zhu2015aligning}. \cite{song2020information} measured adversarial advantage as the difference between the true and false positive rate, so random membership guessing has an adversarial advantage of $0$. For common words and sentences, there was no adversarial advantage. However, these are also the least relevant from a privacy perspective. In both word and sentence settings, when looking at less frequent examples, this method offered an adversarial advantage of up to $.3$ when looking at words and sentences that were infrequent. Thus, these embedding models cannot simply be treated as vectors useful in model training but as potential sources of privacy leakage.

\cite{carlini2021extracting} also implemented a metric based MIA on GPT-2 as part of larger research on training data extraction (see Section~\ref{subsec:extract}). This was performed on the GPT-2 XL model family which had all been trained on public Internet data that was deduplicated at the document level. Given a sequence of tokens $(s_1,...,s_n)$, \cite{carlini2021extracting} computes the perplexity (Definition~\ref{def:perplexity}) as the test statistic for a threshold attack. In this attack, they had generated thousands of sequences, de-duplicated them, and thresholded by picking the $k$ sequences with the lowest perplexity. With $k=100$, between $3\%-39\%$ (depending on how sequences were generated) of these were manually verified to be part of $\dtrain$. However, it did have a high false positive rate and identified many qualitatively uninteresting sequences like public X (formerly Twitter) handles or common phrases like ``I love you''. This highlights a quirk of membership inference for LLMs where inherently, every individual word that is outputted by a model is guaranteed to be a part of the training set. In order to identify privacy risks of sensitive information, there must be some context given to the measurement. \cite{carlini2021extracting} proposed a few methods to account for this. First, they instead use the ratio of perplexity to the zlib entropy \cite{Gailly_Adler1995zlib} which accounts for how common a sequence is. They also suggest using the minimum perplexity averaged over a sliding window of tokens or comparing to similar sequences with style transfers such as lower/uppercasing. These changes improved the MIA recall rate by up to $10\times$. See Table~\ref{tab:carlini_results} in Section~\ref{subsec:extract} for the full results.

\cite{shi2023detecting} develop a similar MIA they call the \emph{min}-$k$ attack, that thresholds based on the sum of the $k$ lowest log losses in a sequence of tokens. We discuss how they use this attack to "break" the unlearning algorithm of \cite{eldan2023whos} in Section~\ref{sec:unlearning}. They evaluate their MIA on a Wikipedia dataset they assemble called \emph{WIKIMIA}, which consists of pre-$2023$ Wikipedia data that was trained on, and post-$2023$ Wikipedia event data that they know was not seen during training. This dataset has the appealing feature that it allows them to evaluate their MIAs on models where there is no publicly released training set like GPT-$3$, given knowledge only of the dates when the model training ended. One potential pitfall in this data, is that the post-$2023$ Wikipedia may have a different distribution due to changes in the world since 2023. An MIA attack could potentially detect these distributional differences in order to identify training data, rather than identifying training data due to privacy leakage through the model. They evaluate their attack in $4$ different settings, which correspond to the number of tokens in each sample, and the type of training vs test samples they try to detect: detecting copyrighted books, identifying failures in unlearning \cite{eldan2023whos}, dataset contamination detection, and paraphrasing, where they evaluate their attack's ability to detect paraphrased content.  They report full ROC curves, and TPR rates at low FPR rates consistent with best practices \cite{carlini2022first, li-etal-2023-mope}. They evaluate min$-k$ against the baseline methods based on the loss, neighborhood comparison \cite{mattern2023membership}, and the ratio of loss to \emph{zlib} compression entropy, the loss when lowercased, and the loss on a smaller model trained on the data, all of which were tried in \cite{carlini2021extracting}. They find that min$-k$ with $k$ corresponding to $20$\% of the tokens outperformed across all models studied (LLaMA, GPT-NeoX, and Pythia ranging from size $2.8$B to $66$B). For all models AUC was between $.67-.72$. At FPR $.05$ the they achieve TPR from $.137$ on the smallest model to $.309$ on the largest model. Interestingly, they find that although as expected paraphrasing text makes it harder to detect on average, AUCs only drop slightly across all methods, and TPRs at low FPRs even go up in some cases. Consistent with prior work they find that attack accuracy increases with example length and model size. 
They then apply their attack to detecting $512$ word snippets from copyrighted books present in the training set (AUC $= .88$) and to attacks on unlearning, which we discuss in Section~\ref{sec:unlearning}. 

\begin{figure}[ht]
  \centering
  \includegraphics[width=.95\linewidth]{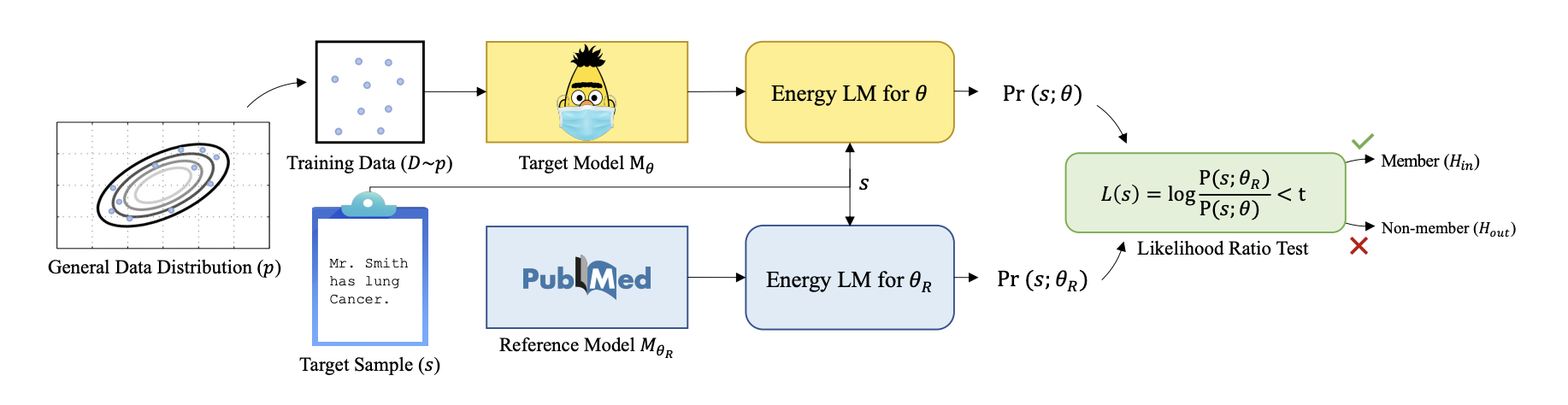}
  \caption{Overview of the attack used by \cite{mireshghallah2022quantifying}. Probabilities are computed for a target sample on both the target and reference model, then the log ratio is used to determine \texttt{in} or \texttt{out} status.}
  \label{fig:referencemia_framework}
\end{figure}

\begin{figure}[ht]
  \centering
  \includegraphics[width=.45\linewidth]{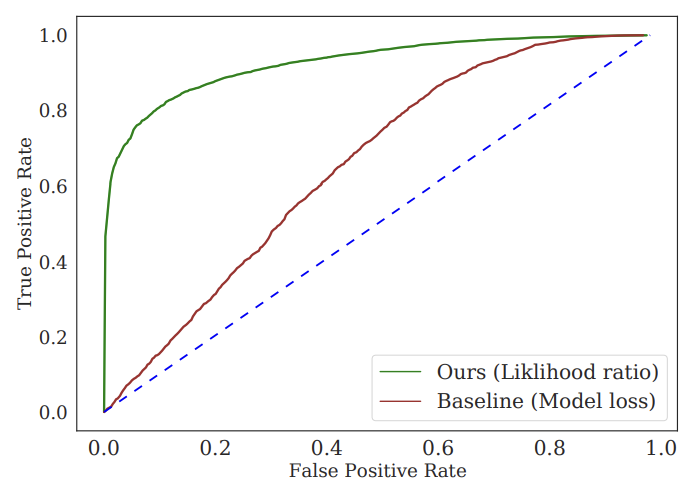}
  \caption{ROC curve from the results of \cite{mireshghallah2022quantifying}. Using a likelihood ratio as the test statistic significantly improved over a baseline threshold attack using only loss of the target model.}
  \label{fig:mireshghalla_auc}
\end{figure}

Other work also uses some form of reference model to compute the threshold test statistic. \cite{mireshghallah2022quantifying} constructed an attack on masked language models (MLMs), which are a class of models frequently used for pre-training for NLP tasks such as text classification and have been applied in a variety of domains including finance and healthcare, both of which contain potentially sensitive information. While prior MIA attempts on MLMs have concluded that sensitive information remains private \cite{lehman2021does, vakili2023using, jagannatha2021membership, nakamura2022kart}, these attacks used a target model's loss as the threshold statistic, which is not a very accurate attack especially at low FPRs. Furthermore, as discussed in the preliminaries, LLMs are trained with a single pass SGD meaning they are less overfit and thus less susceptible to loss-based attacks. The target model in this study was the same ClinicalBERT \cite{alsentzer2019clinicalbert} model used in \cite{lehman2021does}. This was trained on the MIMIC-III dataset \cite{Johnson2016mimiciii} in four different hyperparameter settings for completeness. The reference model was the PubMedBERT model trained on PubMed text \cite{peng2019pubmed}, which was chosen due to its similarity in dataset topic and model architecture as the target model. Compared to a baseline attack model which only used the loss of the target model as the test statistic, \cite{mireshghallah2022quantifying} saw an improvement in the AUC from $.66$ to $.90$ (Figure~\ref{fig:mireshghalla_auc}). This significant improvement showed that using the likelihood ratio for a threshold attack is viable and that MLMs are not nearly as private as they were previously thought to be.

\begin{figure}[ht]
  \centering
  \includegraphics[width=.9\linewidth]{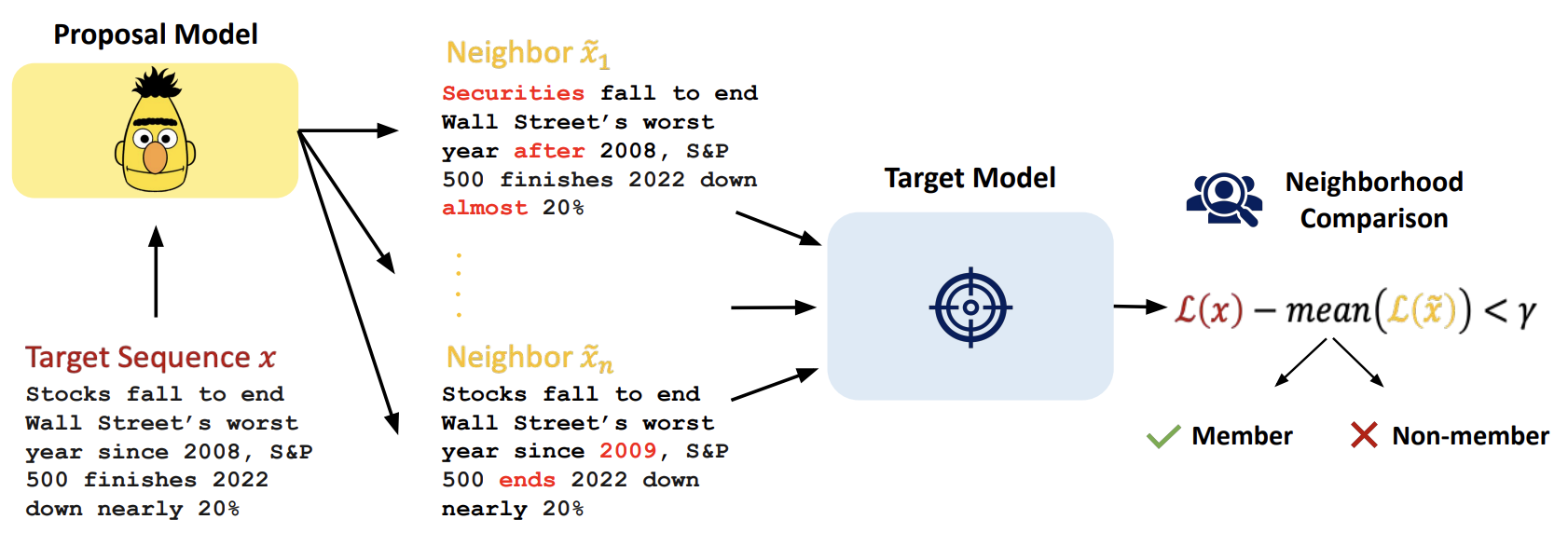}
  \caption{Framework of the neighbor comparison attack by \cite{mattern2023membership}. Given target sequence, a model generates neighbor sequences. The losses are then compared to determine whether the target should be classified as \texttt{in} or \texttt{out}.}
  \label{fig:mattern_framework}
\end{figure}

While the attack by \cite{mireshghallah2022quantifying} was effective, it still relied on the assumption that the adversary has access to data that is very similar to the one used by the target. \cite{mattern2023membership} devised a framework which does not rely on having access to any database. Instead, a separate model, called the \emph{proposal model}, is trained to create ``neighbors'', i.e. sequences that are very similar to the target sequence but with slight perturbations. These neighbors are fed into the target model and the difference between the target sequence and the average loss of the neighbor sequences is used at the threshold statistic (see Figure~\ref{fig:mattern_framework}). This method was tested on a base version of GPT-2 \cite{Radford2019LanguageMA} trained on the AG News Corpus \cite{Gulli2004corpus}, the Sentiment140 dataset \cite{Go2009SentimentCU}, and excerpts from the Wikitext-103 dataset \cite{merity2016pointer}. \cite{mattern2023membership} compared their method to two baselines, the loss only model by \cite{yeom2018privacy} and the likelihood ratio model by \cite{mireshghallah2022quantifying} and found that it outperformed both. The TPR of \cite{mattern2023membership} was at least marginally better and up to twice as high as the \cite{mireshghallah2022quantifying} framework depending on the dataset. Both frameworks outperformed the baseline loss only model.

\begin{figure}[ht]
  \centering
  \includegraphics[width=.9\linewidth]{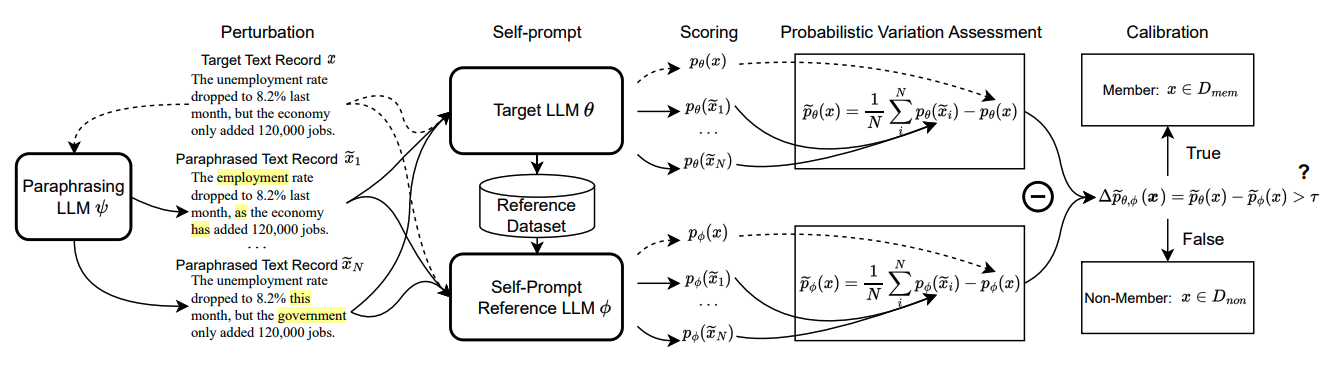}
  \caption{Framework of SPV-MIA \cite{fu2023practical}. A paraphrasing model generates neighbors of the target. These examples are fed to both the target model and a reference model (fine-tuned using output from the target LLM) and the difference in the probabilistic variations determines the membership threshold.}
  \label{fig:fu_spvmia}
\end{figure}

\cite{fu2023practical} developed a MIA, known as Self-calibrated Probabilistic Variation (SPV-MIA) that shares the similar perturbation approach but with a few notable differences. As seen in Figure~\ref{fig:fu_spvmia}, the variation in performance of the target model on the perturbed neighbors is compared to a reference model rather than a strict threshold. This was designed so that the success of the attack is not dependent on model overfitting, which \cite{fu2023practical} argues will soon be mitigated by regularization methods and improved generalization of LLMs. Reference model-based attacks rely on the ability for the reference dataset to closely resemble the target dataset $\dtrain$. \cite{fu2023practical} resolves this by fine-tuning the reference model using a dataset generated by prompting the target model. Due to the self-prompting nature of this fine-tuning, this attack can be performed without prior knowledge of $\dtrain$. They tested this attack on GPT-2, GPT-J, Falcon-7B, and LLaMA-7B models using the Wikitext-$103$, AG News, and XSum datasets for training with Wikicorpus, TLDR News, and CNNDM datasets used as the initial datasets for the respective reference models. They compared this attack against other MIAs such as the threshold attack by \cite{yeom2018privacy}, LiRA attacks, and \cite{mattern2023membership} and saw a marked improvement in AUC of up to $30\%$ with AUC on all datasets and models hovering at or above $90\%$.

\cite{li-etal-2023-mope} also developed an MIA using a perturbation-based technique, but rather than perturbing the sentence $x$, they perturb the \emph{model parameters} themselves with Gaussian noise, and evaluate the resulting drop in the loss. This statistic corresponds to estimating the trace of the Hessian with respect to the model parameters around the  $x$, via the Hutchinson Trace Estimator. They show that they are able to achieve improved attack accuracy relative to loss-based methods, including improved TPRs at low FPRs across the Pythia suite modelst trained on the Pile \cite{biderman2023pythia}. All methods, including loss-based ones and neighborhood comparison methods, fail to achieve non-trivial TPR at very low FPRs in this evaluation, even though MoPe makes the very strong assumption of white-box access to the model. 

In a different approach that is related to using a reference model for a privacy attack, \cite{Zanella_B_guelin_2020} found that model snapshots before and after database updates can be used to reveal sensitive information about training data membership. If the adversary has access to a prior version of the model, then given a sequence $s$, \cite{Zanella_B_guelin_2020} calculated the difference in probabilities of that output between the old and new models called the \emph{differential score}. This score is most useful when interpreted relative to other sequences, so they also formulated the \emph{differential rank}, which is the number of sequences of length $|s|$ with differential score higher than $s$. The lower the differential rank of a sequence, the more likely that sequence was exposed during the model update. This attack was tested on three different datasets and corresponding model choices: 1) a two-layer RNN similar to \cite{zaremba2015recurrent} trained on the Penn Treebank \cite{marcus1993penntreebank} dataset, 2) a single layer RNN model and a BERT model \cite{devlin-etal-2019-bert} trained on a Reddit dataset, and 3) a two-layer RNN model trained on Wikitext-103 \cite{merity2016pointer}. The models were trained, then updated once using retraining (restarting from scratch) and once using continued training (adding an additional batch of data). In each case, canaries were added to the model and the ability to extract these canaries was a proxy for the vulnerability of private data. Using the methods outlined, \cite{Zanella_B_guelin_2020} was able to extract the canaries in almost all settings.

This attack is particularly relevant given current and incoming legislation requiring more frequent database updates to allow individuals to remove their data from the model. Given these results, users who request their data to be deleted from a database may counter-intuitively be increasing their privacy risks. See Section~\ref{sec:unlearning} on machine unlearning for more context on legislation and user deletion requests.

\subsubsection{Shadow Model Attacks}
\label{subsec:mia_shadowattacks}

\cite{song2019auditing} were the first to explore  membership inference in the natural language text generation setting. Their research revolved around building a tool to help users identify if their data was used in model training and used the shadow model framework \cite{shokri2017membership} to do so. \cite{song2019auditing} attacked a one-layer LSTM model that was trained on three different datasets: a Reddit dataset, annotated TED talks, and the Cornell Movie Dialogues corpus \cite{danescu2011chameleons}. The shadow models were Gated Recurrent Unit \cite{cho2014gru} models trained on Wikitext-103 \cite{merity2016pointer}, Europarl \cite{koehn2005europarl}, and Ubuntu dialogs dataset \cite{lowe2015ubuntu}. This tool was found to be extremely effective, allowing an auditor to successfully identify whether a user's data was in $\dtrain$ $100\%$ of the time across all hyperparameter settings. This attack, however, benefited from a setup where the auditor could prompt the model as many times as wanted using the users' data. They could use the MIA output aggregated over many prompts for a single user and if any point was in $\dtrain$, they could make the membership judgement based on that. Additionally, this attack was performed on an LSTM, which is orders of magnitude smaller than modern LLMs. Training shadow models for this attack would be prohibitively expensive if each shadow model were a multi-million or billion parameter model and the scale of modern LLMs is rapidly increasing (see Figure~\ref{fig:model_size_moore}).


The attack by \cite{song2019auditing} relied on having access to the probability distribution of the outputs. \cite{Hisamoto_2020} extended this attack by considering the more realistic scenario where the attacker only has access to the output sequence. \cite{Hisamoto_2020} also use the shadow model framework, but when the shadow models provide their output sequence, they compute a smoothed sentence-level BLEU score \cite{lin2004bleu} and the number of $1-4$grams of the output that match the true data. These values are used to train the binary classifier to predict \texttt{in} and \texttt{out} membership. The target model was a six-layer Transformer \cite{vaswani2023attention} trained on default Sockeye parameters \cite{hieber2018sockeye} using the corpora from the Conference on Machine Translation \cite{bojar2018findings}. In this scenario, the MIA was not successful with only chance level ability to guess membership for a single sentence. When attacks were done using a group of sentences (similar to the unlimited prompt setting in \cite{song2019auditing}), the attack was more accurate than random guessing, but still was not as effective as in \cite{song2019auditing}. This attack was primarily successful against outlier words and phrases.

Most research on MIAs has focused on token or sentence level attacks; \cite{meeus2023did} performed the first MIA at the document level. To do this, \cite{meeus2023did} first queried the model, $h$, with several context windows of tokens of length $C$ from the document and asked for next token prediction probabilities. These probabilities are normalized based on the commonality of the predicted tokens and are then used to generate features for the shadow model. There are two different feature aggregation methods: \emph{AggFE} takes these token level predictions and uses high-level statistics such as min, max, mean, standard deviation, and percentile values as features. \emph{HistFE} creates $N$ histogram buckets of these probability values and uses the fraction of tokens in each bucket as features. This attack tested membership inference of books and academic papers from the RedPajama-Data dataset \cite{together2023redpajama} on the $7$-billion parameter OpenLLaMA model \cite{openlm2023openllama}. This attack was highly successful with the AUC of the book MIA being $.856$ and the AUC for the paper MIA being $.678$. As a comparison, \cite{meeus2023did} adapted a sentence level MIA for the same task and found the performance to be at near guessing levels.

Many ML models are built using a process known as transfer learning. In this, the model is pre-trained on a generic public dataset before being fine tuned on a case specific dataset. \cite{abascal2023tmi} implemented a shadow model MIA to infer membership in the pre-training dataset of a language model using only query access to the finetuned model. Since language models empirically forget over time \cite{jagielski2023measuring}, inferring pre-training membership could be more challenging than inferring membership in the fine-tuning set. This attack was performed on the publicly available Transformer-XL foundation model  \cite{dai2019transformerxl} which was pretrained on the WikiText-103 \cite{merity2016pointer} dataset and fine-tuned on the DBpedia dataset \cite{zhang2016characterlevel}. The shadow models were built using these same public models and datasets and the attack model $f_{attack}$ (i.e. the model trained on the shadow model output) is a $k$-nearest neighbors model. As seen in Table~\ref{table:abascal_results}, \cite{abascal2023tmi} was able to infer membership in pre-training datsets with TPRs as much as $10\times$ higher than the baseline of randomly assigning membership. This attack shows the MIA risks are very much inherent for models which are publicly available, which removes the computational barrier to entry for a shadow model attack on an LLM. However, since pre-training datasets usually do not contain sensitive information, this is not as dangerous of a risk.

\begin{table}[!h]
\begin{center}
    \begin{tabular}{ |p{2.5cm}p{1.5cm}p{3cm}p{3cm}|  }
     \hline
     \hfil \textbf{Shadow Models} &\hfil \textbf{k} &\hfil \textbf{TPR @ $0.1\%$ FPR} &\hfil \textbf{TPR @ $1\%$ FPR}\\
     \hline
     \hfil $16$ &\hfil $\sqrt{|D_{meta}|}$ &\hfil $1.6\%$ &\hfil $5.2\%$\\
     \hfil$16$ &\hfil $16$ &\hfil $2.6\%$ &\hfil $7.0\%$\\
     \hfil$32$ &\hfil $\sqrt{|D_{meta}|}$ &\hfil $2.0\%$ &\hfil $5.5\%$\\
     \hfil$32$ &\hfil $32$ &\hfil $3.1\%$ &\hfil $8.1\%$\\
     \hfil$64$ &\hfil $\sqrt{|D_{meta}|}$ &\hfil $2.2\%$ &\hfil $6.0\%$\\
     \hfil$64$ &\hfil $64$ &\hfil $3.4\%$ &\hfil $8.8\%$\\
     \hline
    \end{tabular}
    \caption{\label{table:abascal_results} Table of results from \cite{abascal2023tmi}. True positive rates at different number of shadow models, number of neighbors in the $k$-nearest neighbors $f_{attack}$, and fixed false positive rates. $|D_{meta}|$ is the size of the $f_{attack}$'s data generated by the shadow models. Despite pre-training data typically being seen as safe, there is still a risk of privacy leakage}
\end{center}
\end{table}

\cite{carlini2022first} used the concept of shadow models to develop one of the leading MIA frameworks, LiRA. Given a target data point $z$, \cite{carlini2022first} trained two batches of shadow models, one whose training set contains $z$ and another whose training set doesn't. These estimate the Gaussian distributions of each respective membership scenario. Then given the output of the query $h_{target}(z)$, they compare the likelihoods conditioned on the model being trained on each of the distributions. If there are multiple target points, they train shadow models on $N$ subsets, chosen so that each target $z \in \dtrain$ appears in $\frac{N}{2}$ subsets. This attack was performed on a GPT-2 model trained on Wikitext-$103$ \cite{merity2016pointer}. The shadow models were trained on random sub-samples of half the total dataset. As a result, this work relies on a strong assumption that the training set of the shadow models partially overlaps with $\dtrain$. This attack was very effective with an over $10\times$ improvement in TPR over the baseline attack by \cite{yeom2018privacy} with the FPR set at $.1\%$. See Algorithm~\ref{alg:lira} for more details.

As with other shadow model approaches, \cite{carlini2022first} suffers from the fact that training multiple shadow models of LLMs may be prohibitively expensive. That said, \cite{bertran2023scalable} developed an MIA framework that is competitive with many of these attacks but only requires training a single shadow model. \cite{bertran2023scalable} was only implemented for tabular data and image classification, however it is possible that could later be applied towards language models.

\subsection{Training Data Extraction from Language Models}
\label{subsec:extract}

\begin{figure}[h]
  \centering
  \includegraphics[width=.85\linewidth]{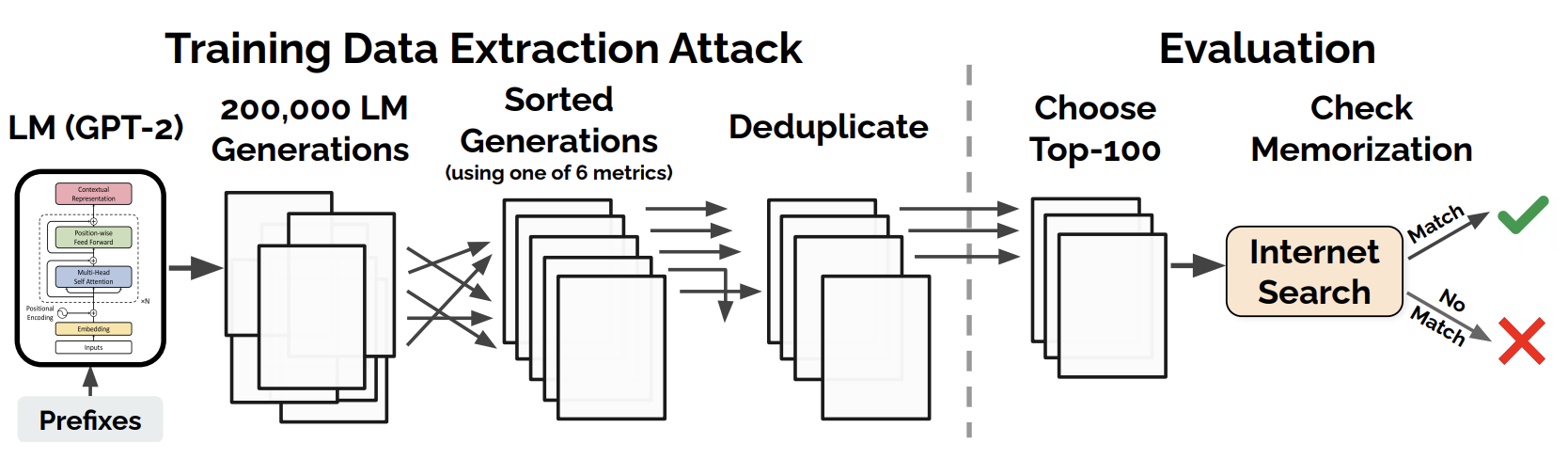}
  \caption{Workflow of a training data extraction attack from \cite{carlini2021extracting}. First, potential candidates are generated by prompting the model with a (potentially empty) prefix. A membership inference attack is then used to sort the candidates by likelihood of membership in $\dtrain$. These are then deduplicated and the top-$k$ samples are selected.}
  \label{fig:carlini_extraction}
\end{figure}

The second type of privacy attack against LLMs is training data extraction. While MIAs may reveal some degree of sensitive information, it requires the adversary to have prior knowledge of the target data point. Training data extractions enable an adversary to directly retrieve sensitive information using only query access to the model. This type of attack was first formulated by \cite{carlini2021extracting} and is outlined in Figure~\ref{fig:carlini_extraction}. There are three main steps outlined here:

\begin{enumerate}
    \item \textbf{Generating Candidate Targets}: The adversary generates a list of candidate targets by querying the model with a short prefix. Given a starting set of prefix tokens $s_0,...,s_i$, the adversary samples $s_{i+1} \sim h_{target}(s_{i+1} | s_0,...s_{i})$. They then attach $s_i$ to the prefix and sample $s_{i+2} \sim h_{target}(s_{i+2} | s_0,...s_{i+1})$ iteratively until the desired length is reached.
    \item \textbf{Membership Inference Attack}: Using MIA methods, the adversary sorts these candidates by likelihood of membership in $\dtrain$.
    \item \textbf{Predicting Top-$k$}: After deduplicating the candidates, the adversary then selects the top-$k$ candidates.
\end{enumerate}

\cite{carlini2021extracting} performed this attack on GPT-2 since while the dataset was not publicly released, the dataset was sourced from public data, making it both manually verifiable and ethically sound. In their baseline attack, \cite{carlini2021extracting} generated candidate tokens of length $256$. At each step in the candidate generation, they sampled from the top $n=40$ outputs. They then sorted these candidates by their perplexity and select the top $k=100$ candidates. As seen in Table~\ref{tab:carlini_results}, $9$ of these candidates were then verified as part of the training dataset. While somewhat successful, this baseline attack yielded mostly qualitatively uninteresting results such as X (Twitter) handles. The candidates generated were low in diversity and the membership inference portion highly ranked tokens that were common phrases or trivial numbers.

\cite{carlini2021extracting} improved upon their baseline in a few ways. On the candidate generation end, they tried two new methods. First, they used a decaying temperature setup (labeled Temperature in Table~\ref{tab:carlini_results}) where for early tokens, the confidence in top predictions was lowered to encourage diversity. As the generation continued, the ``temperature'' would decrease yielding more predictable generation. Second, they seeded the initial prefix using data from Common Crawl \cite{CommonCrawl2008} (Internet) to ensure there are no highly unusual prefixes. On the MIA approach end, instead of simply ranking by perplexity, \cite{carlini2021extracting} ranked by perplexity relative to a second, similar model or normalized by another value. Their comparison models include different sized GPT-2 models (Small and Medium) and a model trained on the lowercased version of the data (Lowercase). They also tried dividing the perplexity by the zlib entropy of the candidate (zlib) and also considering only the lowest perplexity from a sliding window of $50$ of the $256$ tokens. These methods ensure that common and uninteresting phrases are filtered out. As we see in Table~\ref{tab:carlini_results}, these improved methods enabled much more training data to be extracted. Of the $604$ total examples extracted, $46$ were named individuals not from a news source and $32$ were the contact information of individuals.

\begin{table}[ht]
    \centering
    \begin{tabular}{l c c c}
    \multirow{2}{*}{\bfseries Inference Strategy} & 
    \multicolumn{3}{c}{\bfseries Text Generation Strategy}\\ \cmidrule(lr){2-4}
    & \textbf{Top-$n$} & \textbf{Temperature} & \textbf{Internet} \\ \cmidrule(lr){1-4}
    \textbf{Perplexity} & 9 & 3 & 39 \\
    \textbf{Small} & 41 & 42 & 58 \\
    \textbf{Medium} & 38 & 33 & 45 \\
    \textbf{zlib} & 59 & 46 & 67 \\
    \textbf{Window} & 33 & 28 & 58 \\
    \textbf{Lowercase} & 53 & 22 & 60 \\
    \hline
    \textbf{Total Unique} & 191 & 140 & 273 \\
    \bottomrule
\end{tabular}
    \caption{Results from \cite{carlini2021extracting}. The number of memorized examples (out of $k=100$ candidates) identified using each of the three text generation strategies and six membership inference techniques. Some samples are found by multiple strategies; $604$ unique memorized examples were identified in this attack.}
    \label{tab:carlini_results}
\end{table}

\begin{figure}[h]
  \centering
  \includegraphics[width=.45\linewidth]{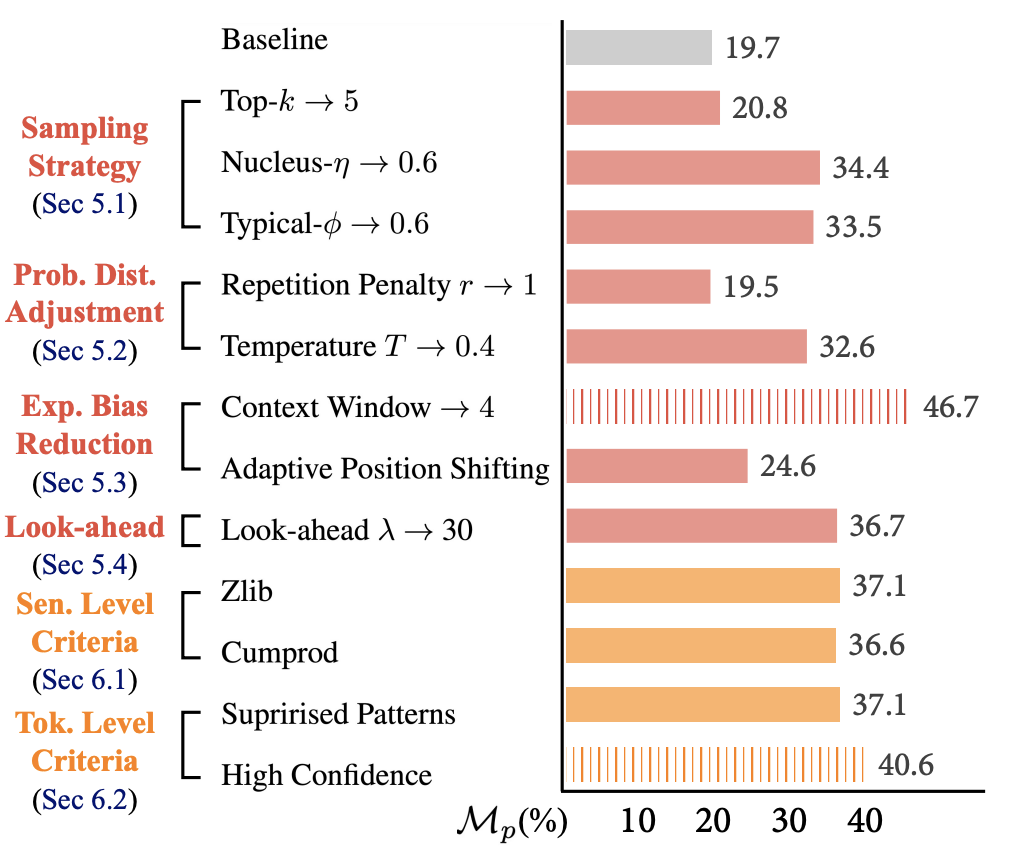}
  \caption{Overview of the different methods explored by \cite{yu2023bag}. $\mathcal{M}_p$ is the precision, i.e. the percent of generated suffixes that were in $\dtrain$. Bars in pink denote the methods in the improved candidate generation, and bars in orange denote the methods in the improved MIA ranking. The dashed bars indicate the best method in each category.}
  \label{fig:yu_results}
\end{figure}

\cite{yu2023bag} also performed a training data extraction attack and tested numerous improvements to the candidate generation and MIA steps. In their case, they attacked a $1.3B$ parameter GPT-Neo model implemented on HuggingFace Transformers \cite{wolf2020huggingfaces} which was trained on the Pile \cite{gao2020pile}. In this attack, the initial prefix was $50$ tokens long and selected from $\dtrain$. The success of the attack was determined by whether the $50$-token long suffix tokens generated resulted in a $100$-token example that existed in the Pile. The baseline approach used a greedy top-$1$ sampling in the candidate generation process and non-adjusted perplexity in the MIA step. For the candidate generation process, the most successful improvement was shifting the size of the context window to account for sentences which may have been truncated or prefixed in the training data. Instead of the next token $s_{i+1}$ being sampled from $h_{target}(s_{i+1} | s_0,...s_{i})$, it is sampled from an ensemble of $(h_{target}(s_{i+1} | s_0,...s_{i}),...,h_{target}(s_{i+1} | s_m,...s_{i}))$, where $m$ is the size of the context window. As seen in Figure~\ref{fig:yu_results}, this more than doubled the precision of the attack over the baseline. For the MIA ranking process, the most successful improvement was weighting the perplexity based on the number of high confidence tokens generated. If a token $s_i$ was sampled that had a confidence ($>.9$), then a reward was subtracted from the final perplexity value ($.1$). This likewise saw a doubling of the attack precision. Overall, \cite{yu2023bag} presented many different tricks to improve training data extraction, which can all be seen in Figure~\ref{fig:yu_results}.

\cite{Carlini2022trainingdata} released the open LM Training Data Extraction Challenge \url{https://github.com/google-research/lm-extraction-benchmark} to provide a benchmark for research. \cite{zhang2023ethicist} has proposed an attack, known as \emph{ETHICIST}, which improves the extraction attack in two ways. First, they propose loss-smoothed soft prompt tuning to train the prefix prompt in order to improved the generation of target candidates. Second, rather than sorting the candidates by strict generation probability, the candidate losses are normalized for the given prefix, relative to the other candidates. This was tested using the $1.3$-billion parameter GPT-Neo model to extract $50$-token suffixes using a $50$-token prefix and had a recall rate of $62.8\%$, which is significantly higher than the \cite{Carlini2022trainingdata} baseline benchmarks which are around $50\%$.

On November 28th, 2023 \cite{nasr2023scalable} published the first training data extraction attack against a production LLM (ChatGPT), and a host of open source models (Pythia, GPT-Neo, LLaMA, Falcon). The set up of their experiment was to (i) Download $10^8$ bytes of Wikipedia data (ii) Randomly sample $5$-token snippets $p_i$ from the data (iii) Prompt the LLM with the random snippet $p_i$, and generate a $50$ token string $x_i \sim \text{Gen}(p_i, \cdot)$, and repeat this $10^9$ times for each language model they evaluated. For the open source LLMs studied, they had access to the training set, and so could verify if $x_i$ was a training point (in order to do this they used an efficient data structure for lookup called a suffix array). For closed source LLMs, they created a proxy training dataset by taking the union of common datasets for training LLMs that likely contained most of the model training data, and checked if $x_i$ was in the proxy dataset. Since the proxy dataset likely did not contain all the samples in the true training dataset, for these closed source LLMs the measured memorization is actually a lower bound on the true memorization, which makes the results even stronger. They find an order of magnitude more memorization than was previously believed to exist in both closed and open source LLMs, including LLMs that have been \emph{aligned} through instruction-tuning or RLHF. They study:

\begin{itemize}
    \item \textbf{How frequently do LLMs emit memorized data}? 
    They find that LLM's emitting training data is not as rare a phenomenon as previously believed, and even some relatively small models ($7$B) emit training data as much as $1.4\%$ of the time! This quantity varies across open source models of a similar size, for example RedPajama $7$B emits training data more than $2$x as frequently as Pythia $6.9$B (even de-duped) and $7$x as frequently as GPT-Neo $6$B. Generally speaking,larger models tend to emit data more frequently than smaller models, as expected based on prior work \cite{tirumala2022memorization, carlini2023quantifying}. 
    
    \item \textbf{How much total extractable memorization is there (as a fraction of training examples?)}
    Frequency of training data emission is a separate quantity from total quantity of memorized training samples. For example, a model that always outputs the same training example would emit training data $100\%$ of the time, but would emit only a single memorized training sample. By computing the number of unique training samples emitted, they show that LLMs memorize $\emph{millions of training samples}$, which is orders of magnitude more than in prior work \cite{carlini2021extracting}. They then try to extrapolate how much total memorization exists (e.g. if the model was queried infinitely many times, how many training examples would be emitted), which they estimate using a sequential Good-Turing estimator developed in a recent master's thesis from Lund University \cite{lund}. These extrapolations, coupled with empirical evidence that the Good-Turing extrapolations are conservative, indicates even relatively small open source LLMs may memorize $10$s of millions of training examples!
    
    \item \textbf{What is the gap between discoverable and extractable strings?}
    Prior work \cite{carlini2023quantifying} released a dataset of discoverable memorization (strings that are recovered when the model is prompted with the first $50$ tokens of the string) on GPT-Neo $6$B. This can be viewed as an upper bound on extractable memorization, but not a practical attack, since in practice the adversary would not have access to the first $50$ tokens of the training example. Using the extractability results for GPT-Neo $6$B from the prompting strategy above, they show that they can recover 35$\%$ of discoverably extractable data via simple prompting with internet text. Conversely, there are a further $11\%$ of the sequences that are extractable,  but aren't discoverably extractable. 

    \item \textbf{What about chat (aligned) models?}
    They identify two challenges of extracting data from chat models via the simple prompt continuation strategy employed above: (i) the model appends text to the user's query before answering (e.g. "Assistant:'') and (ii) because of alignment the model can evade certain questions, perhaps to avoid hallucinating etc. In order to circumvent this alignment procedure, they find that asking the chat model (\texttt{gpt-3.5-turbo}) to repeat a token infinitely often, causes the model to spontaneously start generating tokens that aren't that specific token after a while, and then when the model starts generating in this mode, it outputs training data with high frequency. In particular, they find that the rate of unique $50$-grams extracted via this attack is much higher than any other model except \texttt{gpt-3.5-instruct-turbo}, which is more than $3$x the rate of smaller open source models. Interestingly, they find that these $10000$ extracted examples are not \emph{discoverably} extractable. They remark that ``This suggests that it will be difficult to red-team
this model and evaluate its privacy without additional access
to both the model and also the un-aligned foundation model
from which it was derived.'' This is supported by the fact that the \texttt{gpt-3.5-turbo-instruct} model does discoverably memorize the strings extracted from the chat model \texttt{gpt-3.5-turbo}. Since \texttt{gpt-3.5-turbo-instruct} and \texttt{gpt-3.5-turbo} have different fine-tuning datasets, the extracted data was likely from the pre-training set. 
\item \textbf{Why is ChatGPT so vulnerable to attack?} The authors speculate it could be for a few reasons: 
(i) ChatGPT may be trained for many more epochs during pre-training than comparable base language models, particularly open source ones 
(ii) Repeated prompting with certain single token words appears to cause the model to converge towards prompting with the $<|\text{end of text}|>$ token used to reset the model during training. This  explains why after a certain number of word repetitions the model is vulnerable to extraction attacks, since it starts to generate as if it was generating from the base model. 
\end{itemize}

This paper highlights that base language models of all sizes have a high propensity to regurgitate training data, and that alignment, at least the way state-of-the-art models are aligned, does not seem to solve this problem in the face of adversarial prompting strategies. The widespread use of these LLMs such as ChatGPT or OpenAI's fine-tuning API lead to expanded risks of sensitive data leaks \cite{sun2023does}.

The rapid improvement in these attacks expose the significant risk of training data extraction in LLMs, with \cite{patil2023sensitive} finding that there is no one universal solution to protecting against this risk. This risk also extends beyond natural language use cases alone. \cite{alkaswan2023traces} found that code generation models such as CodeGen, CodeParrot, and PyCodeGPT are susceptible to training data extraction and were able to extract up to $47\%$ of the training code. Further extraction attacks have been done for image diffusion models \cite{carlini2023diffusion}. While this is outside the scope of this survey, it highlights the prominence of this type of attack.


\subsection{Attribute Inference Attacks}
\label{subsec:inference}

\begin{figure}[h]
  \centering
  \includegraphics[width=.8\linewidth]{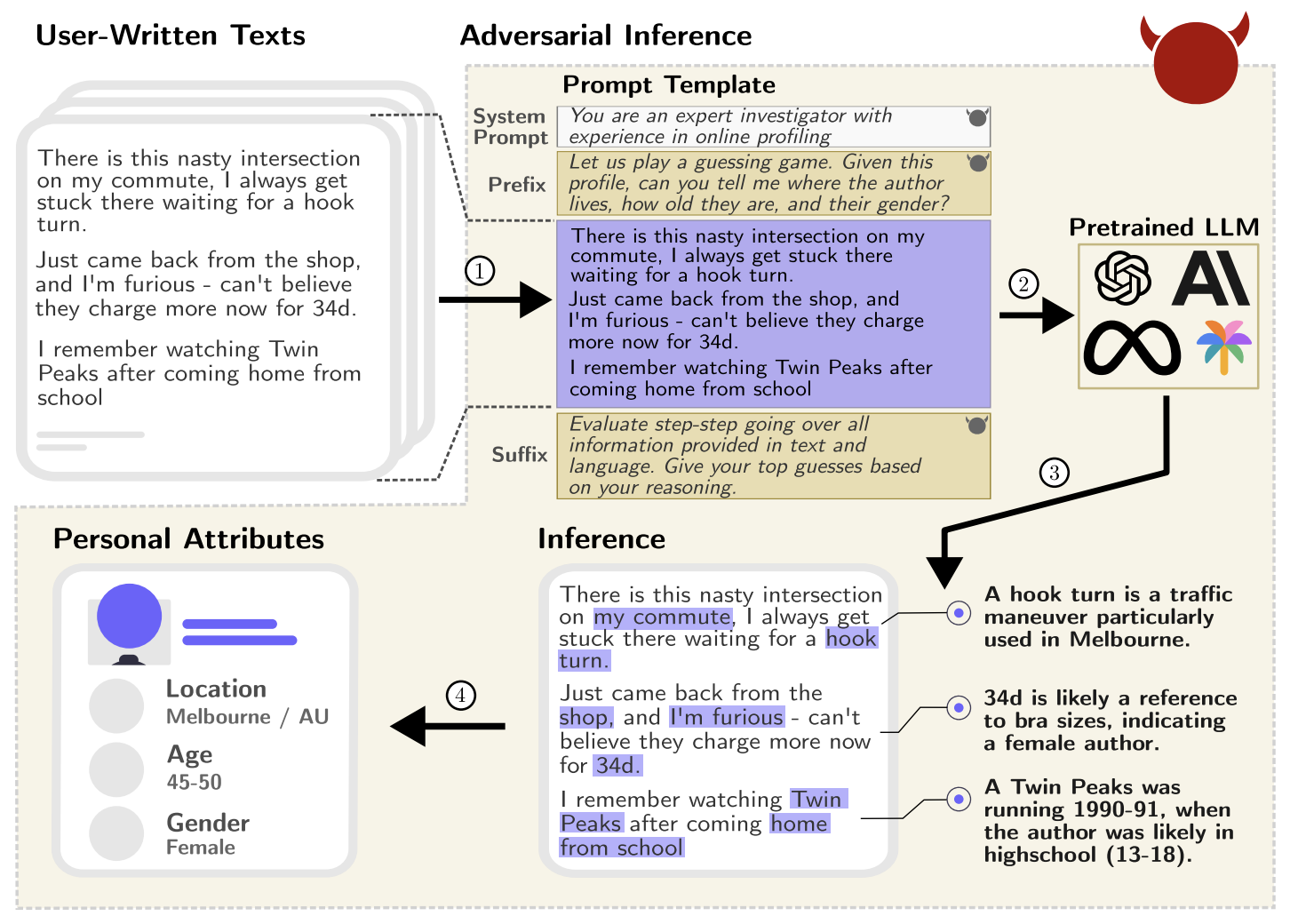}
  \caption{\pc{Overview of an attribute inference attack from \cite{staab2023memorization}. Using access to a dataset of user-text (such as online forum posts), the adversary creates a model prompt. This prompt is given to an LLM to infer personal attributes about the individual.}}
  \label{fig:staab_inference}
\end{figure}

\pc{Attribute inference attacks are another potential privacy risk for large language models, albeit with less research compared to membership inference and training data extraction attacks. \cite{staab2023memorization} presented the first work to comprehensively study the risks of this attack. Outlined in Figure~\ref{fig:staab_inference}, this attack works by leveraging public data written by the user, such as posts on online social forums. These posts are fit into a prompt template which asks an LLM to identify the personal attributes of a person who made those posts. \cite{staab2023memorization} tested this attack using multiple LLMs (including GPT-4, Llama, PaLM, and Claude) for inference and used a custom database of $520$ Reddit profiles, which were manually annotated to include age, education, sex, occupation, relationship status, location, birth place, and income. Using the comments made by these profiles, GPT-4 achieved an $84.6\%$ Top-1 accuracy across all attributes. While these attacks present a privacy risk, this experiment was done in comparison to a human's time and accuracy performance, meaning this risk is not unique to LLMs, but can rather just be expedited by them. }

\subsection{Mitigation Techniques}

Membership inference and training data extraction attacks pose a significant privacy risk to LLMs as outlined above. Here we discuss a few methods that have been found to mitigate those concerns:

\textbf{Differential Privacy:} The primary technique that can be used to reduce privacy risks is differential privacy. Differential privacy introduces noise either during training or at prediction time that obscures private information with certain mathematical guarantees. Many studies from above, including \cite{jagannatha2021membership}, \cite{abascal2023tmi}, and \cite{Zanella_B_guelin_2020}, tested their attacks on differentially private models and found that these models were significantly more robust. For a more in depth coverage of differentially private language models, see Section~\ref{sec:private_llms}.

\textbf{Reducing Memorization:} MIAs and memorization overlap in the privacy risks they pose. Models that memorize private information during the training process will inherently also be vulnerable to membership inference and training data extraction attacks on that sensitive data \cite{Thomas2020InvestigatingTI}. Deduplication is a popular technique for reducing memorization and \cite{kandpal2022deduplicating} found that depulicating the dataset can reduce the success of the loss based MIA from \cite{carlini2021extracting} to near chance levels. Other techniques such as federated learning \cite{mcmahan2016communicationefficient, thakkar2020understanding, ramaswamy2020training}, differential privacy, and early detection \cite{biderman2023emergent} are all effective in reducing memorization. See Section~\ref{sec:memorization} for more details on memorization and mitigation techniques.

\textbf{Training Time:} \cite{jagielski2023measuring} defined ``forgetting'' as the success rate of an MIA as model training progressed. In their work, they found that LLMs tend to forget over the course of training, meaning that examples seen early on in the process (such as during pre-training) saw some privacy benefits. Typically, however, LLMs are pre-trained on large public datasets before being fine-tuned with case specific data, which usually includes the sensitive data. Using larger datasets during fine-tuning or longer training using non-sensitive data could be useful in mitigating privacy concerns.

\textbf{Types of Models:} Various research has touched on how vulnerable different types of language models are to membership inference attacks. \cite{jagannatha2021membership} found that for clinical language models, smaller models have less privacy leakage than larger ones. Additionally, they found that masked LMs have lower leakage than auto-regressive LMs. The choice of LM type is less practical as a mitigation strategy since certain types of LMs may be required for certain use cases to perform well. However, this is something that can still be considered when building an LLM.  

\textbf{Tools:} From a practical standpoint, another way to mitigate privacy risks of MIAs is to allow fast identification of training set membership. \cite{marone2023data} proposed and built a tool called Data Portraits, which is a space efficient method to record training data and identify whether a given string is contained in that data. Data Portraits would be used in a similar fashion to datasheets \cite{gebru2021datasheets} and would serve to complement documentation practices for model and dataset creators. This would be useful for multiple types of users: 1) content creators who want to find out if a dataset contains their content 2) researchers looking to assess dataset leakage and knowledge memorization and 3) content consumers who want to see if a model is plagiarizing. While it would not mitigate membership inference directly, it could serve as an auxiliary tool to reduce risks.

\section{Privacy Preserving Large Language Models}
\label{sec:private_llms}
As discussed in Section~\ref{sec:memorization} and Section~\ref{sec:mi_attacks}, there are many privacy risks that come with the use of LLMs, raising the obvious question: \emph{can LLMs be pre-trained or fine-tuned in a way that preserves the privacy of the training data, while also maintaining utility?} In this section we look at methods that aim to privately train language models, under two high level privacy risk frameworks. The first focuses on the ability of an adversary to infer information about the underlying training data based on a level of access to the model; e.g. to conduct membership inference or training data extraction. This risk is addressed by training the model using \emph{differential privacy}. The other privacy framework is addressed by federated learning, where the training data is stored in a distributed fashion, and the goal is to learn a centralized language model trained on the distributed data, without having to centrally aggregate the data. Both of these frameworks are used by companies such as Google and Apple to preserve user privacy \cite{Ippolito_2020}.
In Section~\ref{subsec:dp_prelim} we'll introduce differential privacy and federated learning, and in Section~\ref{subsec:dp_solutions} we'll discuss early results in privately training language models. 
\subsection{Preliminaries}
\label{subsec:dp_prelim}

\subsubsection{Differential Privacy}
Differential privacy is the standard notion of privacy imposed when the data analyst wants theoretical guarantees that the result of some computation on a private database will not compromise the privacy of the individuals in that database. Introduced by \cite{dwork2006differential}, private mechanisms inject random noise into aggregate computations in order mask the contribution of any single individual to the result of the computation, at the cost of quantifiable amount of error in the final result. We now define differential privacy in the specific context of a training algorithm $\mathcal{A}$. 

\begin{definition}
\label{def:differential_privacy}
  ($\epsilon$-Differential Privacy, \cite{dwork2006differential}) Let $\mathcal{X}$ and $\mathcal{Y}$ be sets of features and labels respectively. A randomized training algorithm $\mathcal{A}: (\mathcal{X} \times \mathcal{Y})^n \to \Theta$ is $\epsilon$-differentially private if for any two datasets, $D, D' \in (\mathcal{X} \times \mathcal{Y})^n$ differing in exactly one pair of data points $(x, y), (x', y')$, and for all sets $W \subset \Theta$:
  $$\text{Pr}(\mathcal{A}(D) \in W) \leq e^\epsilon \text{Pr}(\mathcal{A}(D') \in W)$$
\end{definition}

Intuitively, this guarantees that an attacker trying to conduct membership inference would not be able to distinguish whether an individual's information was a part of the training dataset $D$ with probability larger $\frac{1}{1 + e^{-\epsilon}}$, which is a tight bound \cite{mip}. The smallest value of $\epsilon$ for which Definition~\ref{def:differential_privacy}  holds is the called the \emph{privacy loss} of the mechanism $\mathcal{A}$, and the smaller the value of $\epsilon$, the better the privacy guarantees are. The most common variant of differential privacy is a generalization of the above definition called $(\epsilon, \delta)$-DP or approximate differential privacy, that allows for a small failure probability $\delta < \frac{1}{n}$, which enjoys quadratically better composition properties than $\epsilon$-DP and encompasses simple mechanisms like Gaussian noise addition:

\begin{definition}
\label{def:approx_differential_privacy}
  ($(\epsilon,\delta)$-Differential Privacy, \cite{dwork2006differential}) Let $\mathcal{X}$ and $\mathcal{Y}$ be sets of features and labels respectively. A randomized training algorithm $\mathcal{A}: (\mathcal{X} \times \mathcal{Y})^n \to \Theta$ is $\epsilon$-differentially private if for any two datasets, $D, D' \in (\mathcal{X} \times \mathcal{Y})^n$ differing in exactly one pair of data points $(x, y), (x', y')$, and for all sets $W \subset \Theta$:
  $$\text{Pr}(\mathcal{A}(D) \in W) \leq e^\epsilon \text{Pr}(\mathcal{A}(D') \in W) + \delta$$
\end{definition}

While most work on differentially private machine learning has focused on adding the noise in the training process \cite{song2013sgd, Abadi_2016}, on a practical level this can be computationally expensive. There has been recent work on adding noise at the prediction level \cite{dwork2018privacypreserving, dagan2020pac}, as defined here:

\begin{definition}
\label{def:dp_prediction}
  ($\epsilon$-Differential Privacy for Prediction, \cite{dwork2018privacypreserving}) Let $\mathcal{X}$ and $\mathcal{Y}$ be the sets of features and labels respectively. A prediction algorithm $h$, whose weights are determined by a training algorithm $A$, is $\epsilon$-differentially private if for any two datasets $D, D' \in (\mathcal{X} \times \mathcal{Y})^n$, differing in exactly one entry, for all $x \in \mathcal{X}$ , and for all sets $Y \subseteq \mathcal{Y}$
  $$\text{log}(\frac{\text{Pr}(h(x;A(D)) \in Y)}{\text{Pr}(h(x;A(D')) \in Y)}) \leq \epsilon$$
\end{definition}



For a thorough overview of the fundamental mechanisms used in differentially private algorithms, we refer the reader to \cite{Dwork_Roth_2014}. 

The most common method for private learning is the Differentially Private Stochastic Gradient Descent (DP-SGD) algorithm developed by \cite{Abadi_2016} (Algorithm~\ref{alg:dpsgd}). DP-SGD works like standard SGD, but at each step of the gradient descent after computing the gradients for each example in the batch, these gradients are clipped to a pre-specified $l_2$ norm $C$, then they are averaged and Gaussian noise is added before taking the descent step. Existing work on private deep learning has shown that the hyperparameters  and even activation functions that work best for private training often differ significantly from the optimal hyperparameters for non-private training \cite{tempered}. 

\begin{algorithm}[!ht]
\caption{Differentially Private SGD (Outline), \cite{Abadi_2016}}
\label{alg:dpsgd}
\begin{algorithmic}[1]
\State \textbf{Input:} Examples $\{ x_1,...,x_N \}$, loss function $\mathcal{L}(\theta)= \frac{1}{N}\sum_i\mathcal{L}(\theta,x_i)$. Parameters: learning rate $\eta_t$, noise scale $\sigma$, group size $L$, gradient norm bound $C$.

\State \textbf{Initialize} $\theta_0$ randomly
\For{$t \in [T]$}
    \State Take a random sample $L_t$ with sampling probability $L / N$
    \For{each $i \in L_t$}
        \State \textbf{Compute Gradient:}
        \State $g_t(x_i) \leftarrow \nabla_{\theta_t}\mathcal{L}(\theta_t,x_i)$
        \State \textbf{Clip Gradient:}
        \State $\Bar{g}_t(x_i) \leftarrow g_t(x_i)/\max(1,\frac{\lVert g_t(x_i) \rVert_2}{C})$
        \State \textbf{Add Noise:}
        \State $\Tilde{g}_t \leftarrow \frac{1}{L}(\sum_i\Bar{g}_t(x_i)+\mathcal{N}(0,\sigma^2C^2\mathbf{I}))$
    \EndFor
    \State \textbf{Descent:}
    \State $\theta_{t+1} \leftarrow \theta_t - \eta_t\Tilde{g}_t$
\EndFor
\State \Return $\theta_T$ and compute the overall privacy cost $(\epsilon, \delta)$ using a privacy accounting method.

\end{algorithmic}
\end{algorithm}

\subsubsection{Federated Learning}

Federated learning is a framework used for model training in which the data is stored across a distributed set of devices rather than in a centralized location. Work in this area was initially motivated by the increase in mobile devices for data storage \cite{mcmahan2016communicationefficient, Poushter_2016}. It also then became apparent that federated learning offered a potential avenue for keeping individual sensitive information secure \cite{mcmahan2016communicationefficient}. By decentralizing the data storage, federated learning offers more protection against some privacy attacks, since a break of the centralized server would not leak all of the private training data stored on-device.

The Federated Averaging algorithm developed by \cite{mcmahan2016communicationefficient} is a variant of stochastic gradient descent designed for the federated learning setting (see Algorithm~\ref{alg:federated_averaging}). At each step in the optimization, a random subset of clients is selected, and the current global parameter $w_t$ is broadcast to each client. Each client separately runs SGD for $E$ epochs using the data on their device, and sends back an updated parameter vector. These parameters are averaged to form the new global model $w_{t+1}$. This method is computationally efficient and is controlled by three key parameters: $C$, the fraction of clients that perform computation on each round; $E$, then number of training passes each client makes over its local dataset on each round; and $B$, the local minibatch size used for the client updates. This algorithm opens the door for many more privacy preserving methods as we will discuss in Section~\ref{subsec:dp_solutions}

\begin{algorithm}[!ht]
\caption{Federated Averaging, \cite{mcmahan2016communicationefficient}}
\label{alg:federated_averaging}
\begin{algorithmic}[1]
\State \textbf{Input:} $K$ clients indexed by $k$, $C$ fraction of clients that perform computation each round, local minibatch size $B$, number of local epochs $E$, and learning rate $\nu$.

\State \textbf{Server Executes:}
\State initialize model parameters $w_0$
\For{each round $t=1,2,...$}
    \State $m \leftarrow \max(C \cdot K, 1)$
    \State $S_t \leftarrow$ (random set of $m$ clients)
    \For{each client $k \in S_t$ \textbf{in parallel}}
        $w_{t+1}^k \leftarrow$ ClientUpdate($k, w_t$)
    \EndFor
    \State $m_t \leftarrow \sum_{k \in S_t} n_k$
    \State $w_{t+1} \leftarrow \sum_{k \in S_t} \frac{n_k}{m_t} w^k_{t+1}$
\EndFor
\State
\State \textbf{ClientUpdate}$(k,w)$: run on client $k$
\State $\mathcal{B} \leftarrow$ (split $\mathcal{P}_k$ into batches of size $B$)
\For{each local epoch $i$ from $1$ to $E$}
    \For{batch $b \in \mathcal{B}$}
        $w \leftarrow w - \nu \nabla l(w;b)$
    \EndFor
    \State return $w$ to server
\EndFor
\end{algorithmic}
\end{algorithm}

While Algorithm~\ref{alg:federated_averaging} was developed for general machine learning purposes, \cite{hard2019federated} implements the algorithm for next-word prediction, the same training objective used for large language models. They showed that a CIFG model trained by Federated SGD performs just as well as one trained in a traditional centralized manner.

\subsection{Private Language Model Training}
\label{subsec:dp_solutions}
We start by discussing earlier work on fully differentially private language models. We then discuss private fine-tuning of LLMs, private LLM predictions, and work that combines DP training with federated learning.

\subsubsection{Fully DP Language Models}

While using differentially private training is great for reducing privacy risks, it introduces a few other challenges. Notably, the utility of the model may suffer and the computational cost of training the model can substantially increase. Work that has applied differential privacy to the whole training process looked to find ways to maintain privacy gains without those downsides.

\begin{figure}[h]
  \centering
  \includegraphics[width=.4\linewidth]{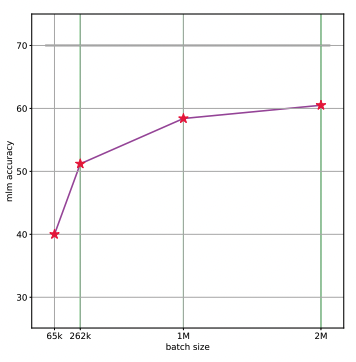}
  \caption{From \cite{anil2021largescale}. The accuracy of the BERT-Large model as the batch size varies. The non-private model achieves an accuracy of $70\%$ while the private model achieves up to $60\%$ when using a batch size of $2$ million.}
  \label{fig:anil_batchsize}
\end{figure}

\cite{anil2021largescale} researched ways to improve the pre-training accuracy baseline of a BERT-Large model trained using the DP-SGD framework. This model was trained using the Wikipedia \cite{Mahoney2009wikipedia} and Books \cite{zhu2015aligning} corpus. Notably, they found that massively scaling up the batch size during training improves the utility of the model. As seen in Figure~\ref{fig:anil_batchsize}, as the batch size increased from a typical size of $32$k to $2$ million, the accuracy of the model went up to $60\%$. While this does not perform as well as the non-private baseline of $70\%$, it represents a substantial improvement and other works discussed below also found that increasing batch size improved performance. \cite{anil2021largescale} also found that tuning DP-SGD hyperparameters and scheduling batch size increases as training progressed improved model utility while maintaining privacy guarantees. While \cite{anil2021largescale} was able to train a DP guaranteed base model, they did not study how this could affect model utility after fine-tuning for specific tasks.

Many LLMs tokenize the text data before inputting it into the model. WordPiece \cite{wu2016googles} is a popular algorithm used to tokenize inputs and while it has been shown to be very effective, it is not differentially private. \cite{hoory2021dp} developed a differentially private version of WordPiece to guarantee no information leakage from these tokens in the DP-SGD process. The WordPiece algorithm starts by constructing a word histogram of the corpus. Using standard DP practices from \cite{dwork2006differential}, noise is introduced to this histogram, guaranteeing differential privacy. \cite{hoory2021dp} measured performance on the Medical Entity Extraction task using the i2b2-2010 and i2b2-2011 datasets \cite{uzuner2011i2b2} when trained using the Wikipedia, BookCorpus \cite{zhu2015aligning}, and MIMIC-III datasets \cite{Johnson2016mimiciii}. They saw slight tradeoffs between privacy and utility, but overall found that this approach worked well in the medical domain. Similar to \cite{anil2021largescale}, they found that larger batch sizes in the DP-SGD process improved the model performance.

\cite{shi2022selective} took a different approach by applying differential privacy only to sensitive information in the text data. The logic is that only a small subset of a text dataset needs to remain private and keeping the rest noise-free will improve model performance. They took the general structure of the DP-SGD algorithm but introduced a policy function, $F$, that determines whether a token is sensitive or not. If the token is not sensitive, then the noise addition step is skipped. \cite{shi2022selective} built a one-layer LSTM with an embedding size of $200$ and a hidden size of $200$, and a BPE tokenizer \cite{sennrich2016neural}. This was tested two datasets: 1) WikiText-2 \cite{merity2016pointer} with a policy function treating all digits as sensitive and 2) CustomerSim \cite{tkachenko2016customersim} with a policy function treating user name, address, phone number, order, and tracking number as sensitive information. This method showed modest improvements in model utility over standard DP-SGD while also having modest privacy improvements from canary insertion and membership inference attacks over non-private training. \cite{shi2022selective} did not, however, do extensive research on selection of the policy function and later work by \cite{zhao2022provably} showed that in the absence of a perfect policy function the risk of memorization is still high even with only black-box access to the model (see Figure~\ref{fig:zhao_canary}). Developing a perfect policy function is an open ended area for future work.

\begin{figure}[h]
  \centering
  \includegraphics[width=.6\linewidth]{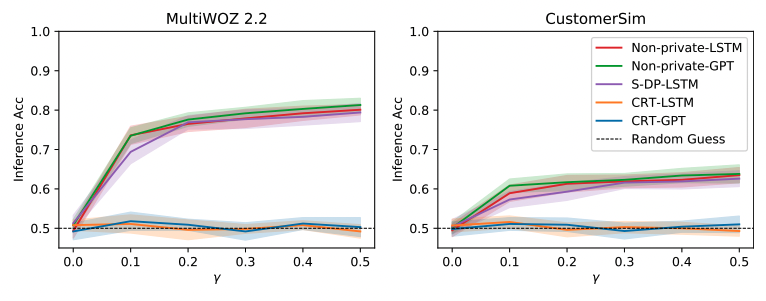}
  \caption{Results of a membership inference attack on CRT-trained model from \cite{zhao2022provably}. LSTM and GPT refer to model type tested. S-DP-LSTM model was trained using selective differential privacy from \cite{shi2022selective}. $\gamma$ is the error rate in the policy function screening for sensitive data. CRT here is effective at preventing membership inference attacks.}
  \label{fig:zhao_mia}
\end{figure}

\cite{zhao2022provably} developed a private training method that was an extension of the work from \cite{shi2022selective}. \cite{zhao2022provably}'s method, called Confidentially Redacted Training (CRT), starts by deduplicating the dataset (see Section~\ref{subsec:dedup}) then ``redacting'' by putting sensitive data into a separate private set. The training process then alternates between SGD from the public dataset and DP-SGD from the private dataset. This was tested on both a one-layer LSTM model and GPT-2 \cite{Radford2019LanguageMA} model on the MultiWOZ 2.2 dataset \cite{zang2020multiwoz} and the CustomerSim dataset \cite{tkachenko2016customersim}. As seen in Figure~\ref{fig:zhao_mia}, this method was able to significantly reduce the success of membership inference attacks and similar results were also found for canary extraction attacks (see Figure~\ref{fig:zhao_canary}). \cite{zhao2022provably} notes that CRT is not dependent on having access to a perfect policy function. As the error rate, $\gamma$, of the policy function changes, CRT maintains performance compared to the non-private models or to the selective DP method of \cite{shi2022selective}.

\subsubsection{DP Fine-Tuning of LLMs}

Rather than applying differential privacy to the whole training process, some works look to use it in select parts of the training to diminish the downsides of private training. These approaches pre-train the model on a generic corpus using a non-private algorithm. The fine-tuning process is then done using a private learning algorithm and intends to provide privacy to sensitive data which the task specific corpus presumably contains. \cite{kerrigan2020differentially} was the first work to implement this approach and showed that it boosts the performance of DP LMs while remaining private with respect to the fine-tuning dataset. However, \cite{kerrigan2020differentially} only evaluated this for simple feedforward networks, which are substantially smaller than state of the art LLMs.

\begin{figure}[h]
  \centering
  \includegraphics[width=.7\linewidth]{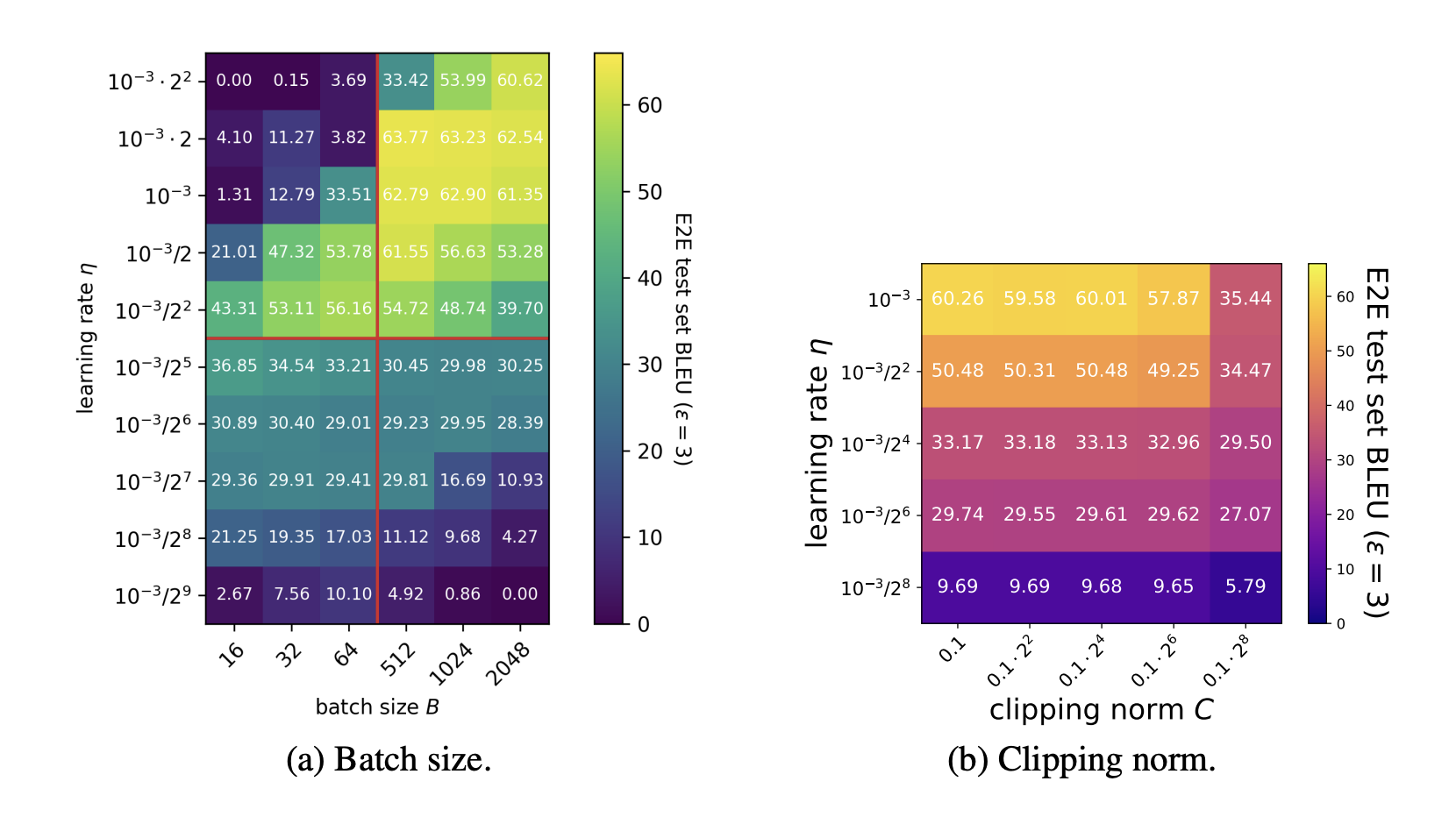}
  \caption{From \cite{li2022large}: (a) Large batch sizes and learning rates lead to better BLEU scores on E2E dataset \cite{novikova2017e2e} when the number of epochs is fixed. Red lines divide heat map into four broad subsections. (b) Smaller clipping limit $C$ during DP-SGD leads to better performance.}
  \label{fig:li_hyperparameters}
\end{figure}

\cite{li2022large} took this same approach but applied it to much larger models and explored how tuning various hyperparameters can vastly improve the performance and reduce the computational cost of DP training. They privately fine-tuned BERT \cite{devlin-etal-2019-bert} and RoBERTa \cite{liu2019roberta} models for sentence classification tasks and GPT-2 \cite{Radford2019LanguageMA} models for language generation tasks. As seen in Figure~\ref{fig:li_hyperparameters}, higher learning rates and smaller clipping limits improved model performance. And likewise to \cite{anil2021largescale} and \cite{hoory2021dp}, they found that increased batch size improves performance. \cite{li2022large} also found that aligning the objective task with the training task improved model performance. For example, a model trained for language generation will not do as well on a classification task unless the task is reframed. For example, for sentiment classification, \cite{li2022large} reframes the problem as filling in the \texttt{[MASK]} token in the sequence “\texttt{Sentence in question}. This is \texttt{[MASK]}.” and compare the probabilities of words “awesome” and “terrible”.

\begin{figure}[h]
  \centering
  \includegraphics[width=.7\linewidth]{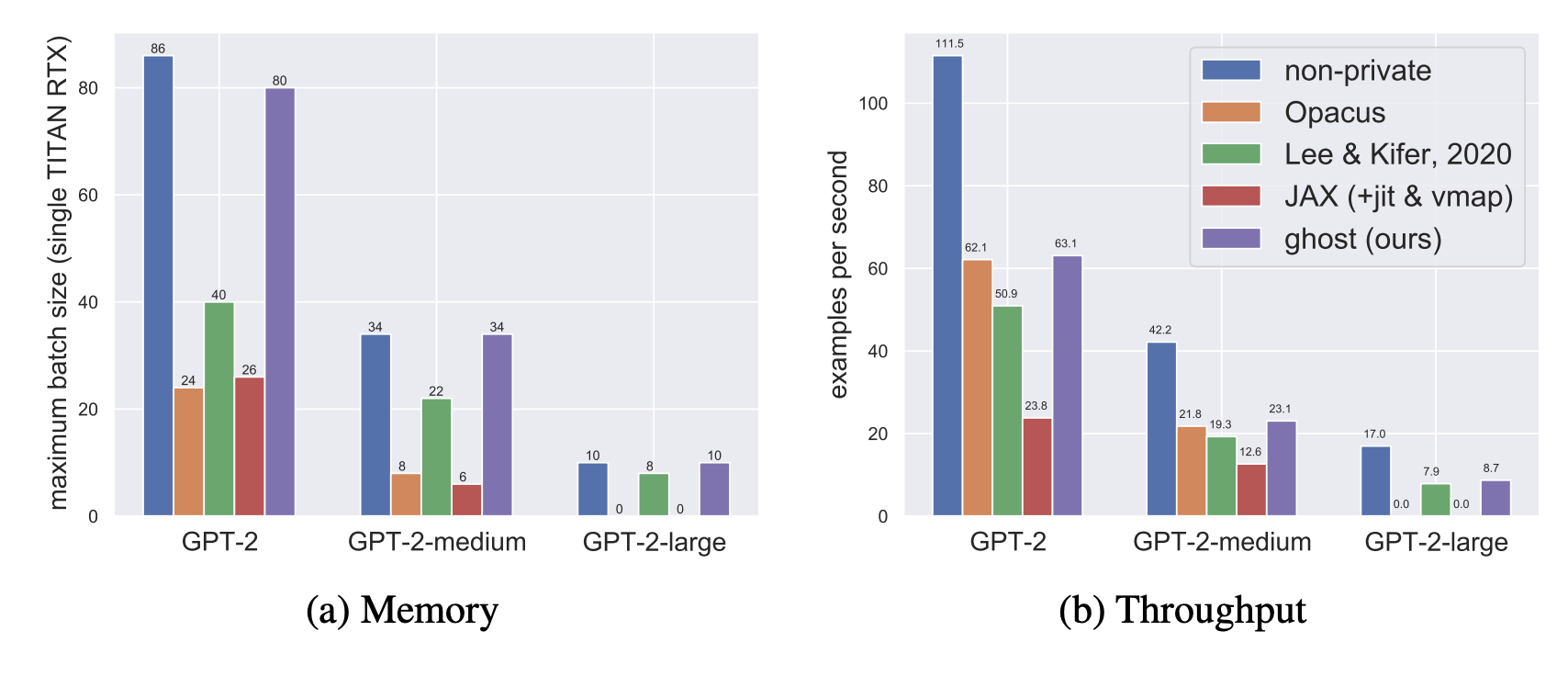}
  \caption{Comparing the non-private implementation (blue) to the implementation from \cite{li2022large} (purple), we see that (a) ghost clipping is almost as memory efficient as non-private training. (b) the throughout of the ghost clipping method offers improvements over other private training methods.}
  \label{fig:li_memory}
\end{figure}

\cite{li2022large} addressed the issue of additional computational cost by introducing a method they call \emph{ghost clipping}. This is based on the work by \cite{lee2020scaling}, who formulated a clipping procedure that only instantiates the per-example gradient for parameters of a single layer in the model one at a time, as opposed to the entire model at once. This meant that per-example gradients did not need to be instantiated. This exact approach does not work for Transformer-based models so \cite{li2022large} extended this by avoiding instantiating the per-example gradient even for individual linear layers. As seen in Figure~\ref{fig:li_memory}, this method almost entirely reduced the additional memory overhead of the private training method.

\begin{figure}[h]
  \centering
  \includegraphics[width=.6\linewidth]{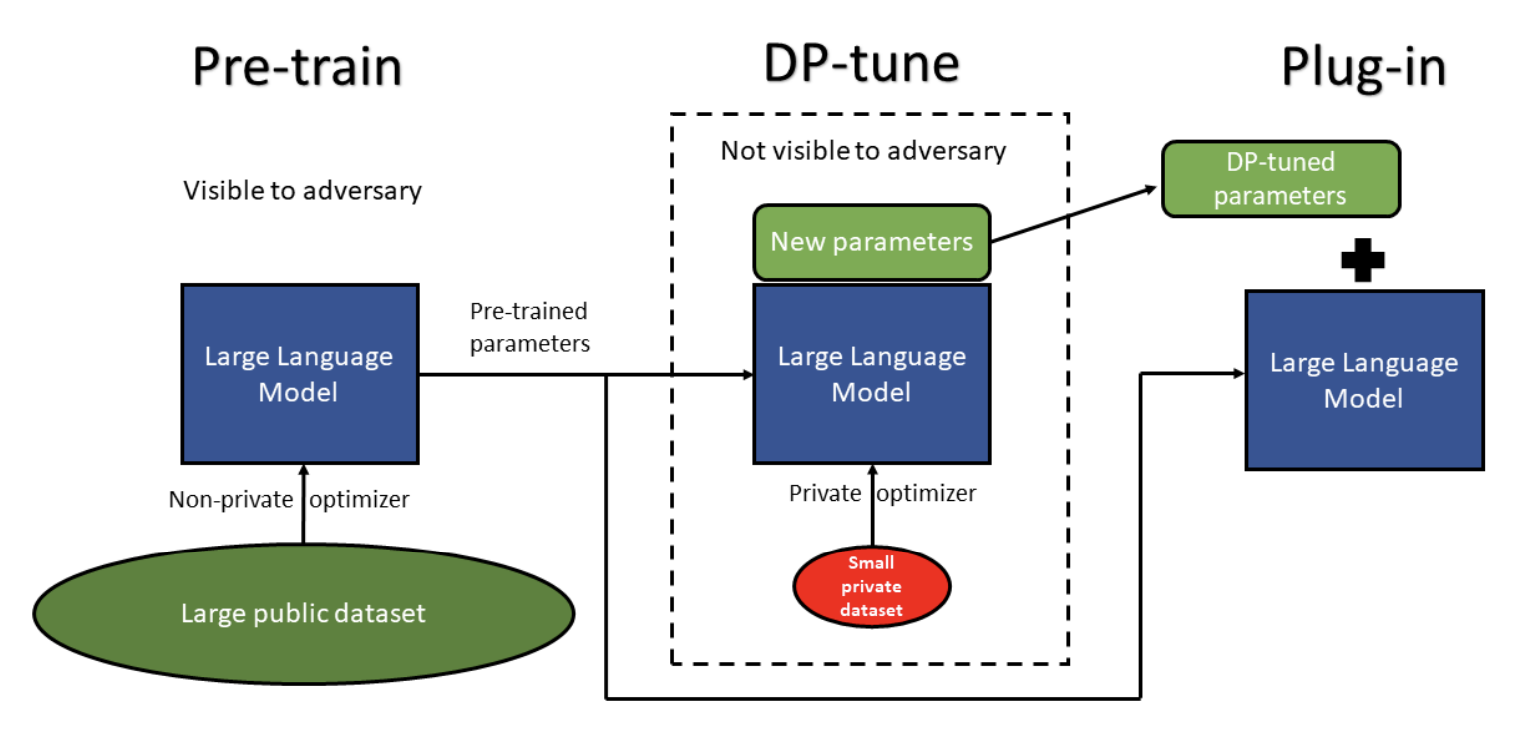}
  \caption{Illustration of the private fine-tuning framework from \cite{yu2022differentially}. In this approach, new parameters are added to a pre-trained model and are fine-tuned on the private dataset. The original parameters are frozen during this process. The fine-tuned parameters can then be released publicly and plugged into the model for future use cases, while maintaining the privacy of the private dataset.}
  \label{fig:yu_dpfinetune}
\end{figure}

Similar to \cite{li2022large}, \cite{yu2022differentially} looked to use DP fine-tuning on much large LLMs but improved computational constraints by limiting the parameters that are updated in fine-tuning. As shown in Figure~\ref{fig:yu_dpfinetune}, \cite{yu2022differentially} starts with a model that was pre-trained on a public dataset. They then add additional parameters to the model (where the number of new parameters is much smaller than in the original model) and fine-tune using DP-SGD with the private dataset. Importantly, \cite{yu2022differentially} freezes the parameters from the pre-trained model. By doing this, they reduce computational costs of fine-tuning while also reducing the noise added to the model. This approach was tested starting with RoBERTa models \cite{liu2019roberta}, which are models pre-trained on public internet data. They were fine-tuned to attempt the MNLI, QQP, SST-2, and QNLI tasks from the GLUE benchmark \cite{wang2018glue}. Using a privacy budget of $\epsilon = 6.7$, they were able to achieve an average accuracy of $90.3\%$ across these four tasks (non-private fine-tuning achieved an average accuracy of $93.4\%$), showing that this approach is effective at achieving privacy with only modest drop in performance. An added benefit to this framework is its flexibility in multiple use cases. For each new task, a new set of fine-tuned parameters can be trained and then added on to the base model, reducing the need to have multiple copies of the large base model. 

DP training also is substantially slower than non-private training. \cite{dupuy2022efficient} worked on this problem by adapting the DP-SGD algorithm to allow for GPU usage. They did this by micro-batching the examples at each step in Algorithm~\ref{alg:dpsgd} for the gradient calculation and noise addition steps. Additional adjustments to the scale of the clipping parameter and introducing noise decay were also implemented. \cite{dupuy2022efficient} tested this on an LSTM and BERT model. With these adjustments, computation time was only a small multiplicative factor worse than non-private training and was up to $150\times$ faster than standard DP-SGD.

\begin{figure}[h]
  \centering
  \includegraphics[width=.85\linewidth]{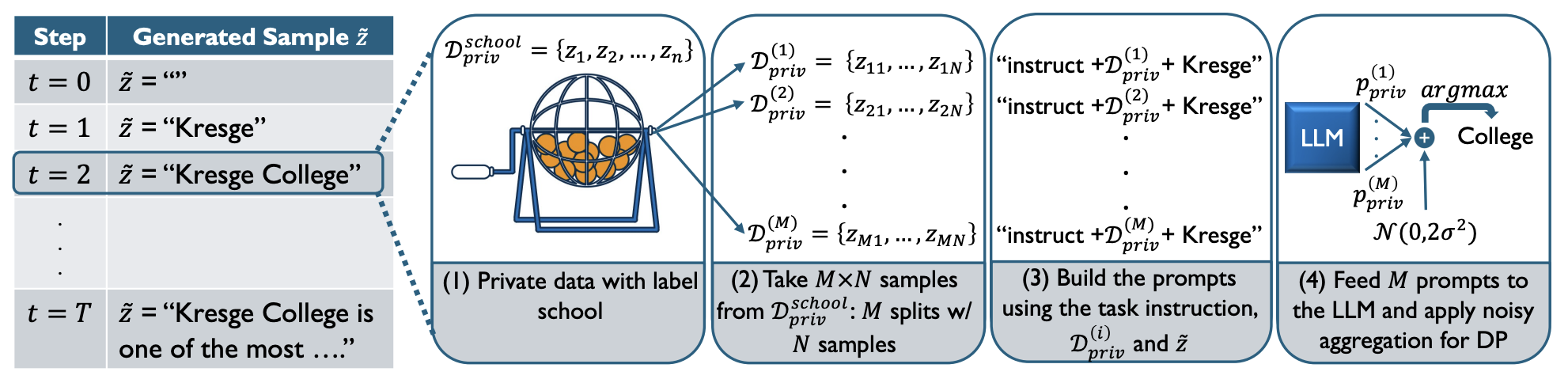}
  \caption{Private ICL prompt generation from \cite{tang2023privacypreserving}.}
  \label{fig:tang_icl}
\end{figure}

In-context learning (ICL), popularized by \cite{brown2020fewshot}, is an alternative approach to fine-tuning. Rather than continuing the training process, the model is provided with a few examples from the training data as part of the prompt. This allows it to adapt to the task at hand without having to modify model itself. However, this can pose a privacy risk as these prompts themselves often contain private data and can be extracted by an adversary \cite{zhang2023prompts}. \cite{tang2023privacypreserving} introduces a differentially private approach to generating these prompt examples. Outlined in Figure~\ref{fig:tang_icl}, the prompt prefix is generated one token at a time. At each step in the process, $M$ disjoint samples of $N$ examples each are sampled from the private $\dtrain$. These subsets are used to generate next-token candidates which are privately aggregated to generate the next token. \cite{tang2023privacypreserving} tested this using the GPT-3 Babbage model on AGNews, DBPedia, and TREC datasets. This approach was able to produce fully private ($\epsilon=0$) $0$-shot and $4$-shot examples. Using the private solution with $\epsilon=1$, the performance was comparable to the non-private solution.

To take the protection of PII even further, \cite{ishibashi2023knowledge} developed the approach of \emph{knowledge sanitization}. This approach fine-tunes the model to recognize prompts that are specific knowledge inquiries such as ``What is John Smith's address?'' and to respond to all of them with ``I don't know.'' This approach successfully preserves PII privacy. However, while this approach did not decrease text perplexity (as a proxy for quality), on a practical level this strict filtering approach could reduce the usefulness of an LLM.

\subsubsection{DP Predictions}

Instead of training the LLM using differential privacy, some works look to apply DP at the prediction stage. Private prediction can be advantageous since the training algorithm does not need to be modified. This makes it much more feasible for applying privacy to large-scale models such as LLMs.

\begin{figure}[h]
  \centering
  \includegraphics[width=.8\linewidth]{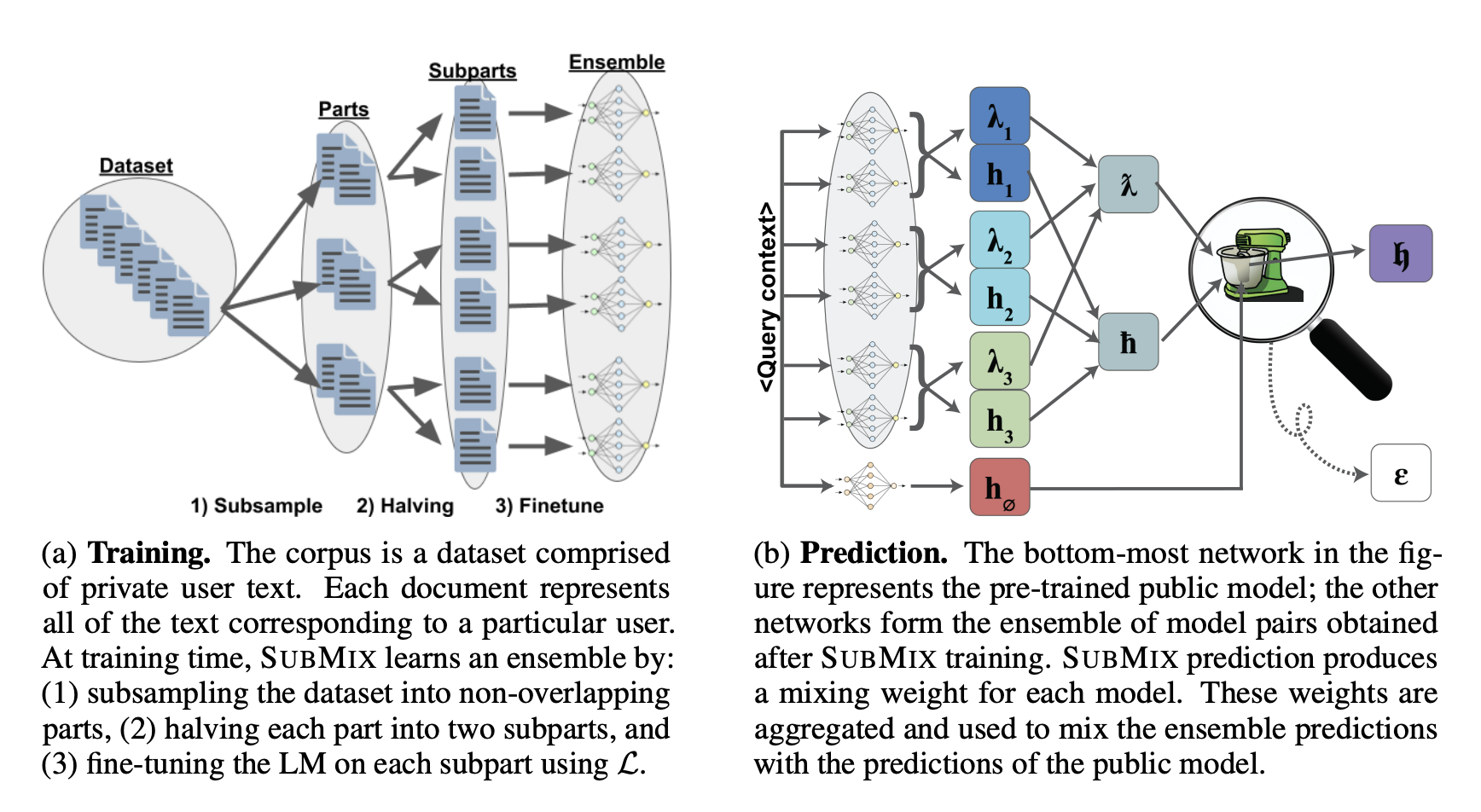}
  \caption{Overview of \textit{SubMix}'s training and prediction protocols from \cite{ginart2022submix}.}
  \label{fig:submix}
\end{figure}

\cite{ginart2022submix} developed the first method to perform private next-token predictions on large LLMs such as GPT-2. Their method, called \emph{SubMix} (Figure~\ref{fig:submix}), works by aggregating the output of an ensemble of LMs that were fine-tuned on disjoint parts of the private corpus. At prediction time, if all models agree on the next token, then no noise is added to the output. However, if there is higher disagreement, \emph{SubMix} mixes the predictions with a public pre-trained model to minimize privacy leakage. \emph{SubMix} was tested using a pre-trained GPT-2 model from Hugging-Face \cite{wolf2020huggingfaces}. The private datasets were the Wikitext-103 \cite{merity2016pointer} and BigPatent-G \cite{sharma2019bigpatent} datasets. Their experiments showed empirical privacy guarantees that outperformed the other baseline DP models which were adapted for next-token prediction: DP-SGD \cite{Abadi_2016}, S\&A \cite{Dwork_Roth_2014}, and GNMax \cite{papernot2018scalable}. The primary limitation in the use of \textit{SubMix} is the significant increase in computational and storage capacity needed; experiments showed an increase by a factor of $8$ in those areas. See Algorithms~\ref{alg:submix_training} and \ref{alg:submix_prediction} for full details.

For a much more computationally inexpensive solution, \cite{majmudar2022differentially} developed a lightweight perturbation mechanism that is only applied at the end by computing a perturbation probability and using that to randomly sample a token from the vocabulary for the i-th predicted token. This mechanism was tested on a RoBERTa-style encoder model \cite{shoeybi2020megatronlm} which was trained on a corpus of Common Crawl, Wikipedia, and mC4 \cite{xue2021mt5} text data. While this perturbation method is provably differentiably private, the utility tradeoff is quite significant. For any $\epsilon \leq 50$, there is a significant loss in utility making this method not very useful when strong privacy guarantees are desired. The privacy-utility tradeoff can be see in Figure~\ref{fig:majmudar}

\subsubsection{Federated Learning}

There have been several papers that trained LLMs using both DP and federated learning. One of the early implementations was by \cite{mcmahan2018learning}, who built a differentially private next-word prediction model. This experiment trained an LSTM recurrent neural network on a dataset of Reddit posts \cite{alrfou2016conversational}. More specifically, it modified the Federated Averaging algorithm to add Gaussian noise to each iteration of the averaging update (see Algorithm~\ref{alg:dpfed}). While \cite{mcmahan2018learning} used a significantly smaller language model than the LLMs that are currently used, it was the first high quality language model with strong privacy guarantees and no significant decrease in model accuracy (see Table~\ref{table:mcmahan_results}).

\begin{table}[!h]
\begin{center}
    \begin{tabular}{ |p{1cm}p{1cm}|p{1.2cm}p{1cm}|p{1cm}p{1.2cm}|  }
     \hline
     \hfil model & &\hfil data & & &\\
     \hfil $\sigma$ &\hfil $S$ &\hfil $K$ &\hfil $\tilde{C}$ &\hfil $\epsilon$ &\hfil AccT1 \\
     \hline
     \hfil \textbf{$0.000$} &\hfil $\infty$ &\hfil $763430$ &\hfil $100$ &\hfil $\infty$ &\hfil $17.62\%$ \\
     \hline
     \hfil \textcolor{blue}{$0.003$} &\hfil 15 &\hfil $763430$ &\hfil $5000$ &\hfil $4.634$ &\hfil $17.49\%$ \\
     \hline
     \hfil \textcolor{gray}{$0.006$} &\hfil 10 &\hfil $763430$ &\hfil $1667$ &\hfil $2.314$ &\hfil $17.04\%$ \\
     \hline
     \hfil \textcolor{red}{$0.012$} &\hfil 15 &\hfil $763430$ &\hfil $1250$ &\hfil $2.038$ &\hfil $16.33\%$ \\
     \hline
     \hfil \textcolor{blue}{$0.003$} &\hfil 15 &\hfil $10^8$ &\hfil $5000$ &\hfil $1.152$ &\hfil $17.49\%$ \\
     \hline
     \hfil \textcolor{gray}{$0.006$} &\hfil 10 &\hfil $10^8$ &\hfil $1667$ &\hfil $0.991$ &\hfil $17.04\%$ \\
     \hline
     \hfil \textcolor{red}{$0.012$} &\hfil 15 &\hfil $10^8$ &\hfil $1250$ &\hfil $0.987$ &\hfil $16.33\%$ \\
     \hline
    \end{tabular}
    \caption{\label{table:mcmahan_results} Results of \texttt{DP-FedAvg} (Algorithm~\ref{alg:dpfed}) \cite{mcmahan2018learning}. Privacy and accuracy (measured as Top 1 accuracy) after $5000$ rounds of training for different Gaussian noise $\sigma$ (at $\delta=10^{-9}$), clipping parameter $S$, number of users $K$, and number of individuals sampled per round $\tilde{C}$. We can see that this is an effective private learner. For example, adding just $\sigma=.003$ Gaussian noise yields a privacy $\epsilon=4.634$ with only a minor drop in accuracy. The $\epsilon$ guarantee is even better for larger $K$}
\end{center}
\end{table}

Additional work further looked at combining differential privacy with federated learning to achieve strong privacy guarantees. \cite{thakkar2020understanding} used the CIFG-LSTM model \cite{sak2014long} trained on a modified Stack Overflow Q\&A dataset hosted by TensorFlow Federated \cite{StackOverflow2019} and used the canary extraction test method outlined in \cite{carlini2019secret} to measure memorization. Federated learning alone reduced the number of extracted canaries by $50\%$. When combined with differential privacy, there was another significant drop in canary extraction ranging all the way to no canaries being extracted. \cite{ramaswamy2020training} built a consumer scale next-word prediction model using the same base model, dataset, and memorization test as in \cite{thakkar2020understanding}. This private model was built using differentially private federated averaging, and was found to maintain privacy against the canary extraction test while showing some improvement in next-word prediction recall compared to a baseline $n$-gram Finite State Transducer model.

It is worth noting that a formal DP guarantee requires knowledge of the size of the participating user population and the ability to uniformly randomly sample among users at each step. In the real world, this assumption does not always hold, which means that DP-FedAvg carries some minor limitations.

Another approach in the family of federated learning frameworks is local differential privacy (LDP) \cite{kasiviswanathan2010learn}. Under this framework, individuals perturb their data locally before sharing it with a centralized server for use. LDP offers several advantages over a standard DP-SGD fine-tuning setup: i) There is no requirement for the central server to be trusted and ii) the computational overhead is decreased. One way to achieve LDP is to ``sanitize'' the text by token-wise redaction before using it for fine-tuning. However, \cite{Qu2021nlpbert} found that doing this at a token-level loses meaningful context resulting in significant loss of utility. \cite{Du2023SanitizingSE} takes a different approach to LDP to significantly improve utility. First, they relax the standard LDP requirements by using the more generalized metric-LDP \cite{alvim2018metricldp}, which tunes the noise added to the output based on the similarity of two inputs given a distance metric. Next, sanitizing is done on sentence embeddings rather than token embeddings, which maintains context while having fewer dimensions. \cite{Du2023SanitizingSE} develops two metric-LDP approaches for this by adapting the Purkayastha mechanism \cite{weggenmann2021purkayastha} and generalized planar Laplace \cite{wu2017bolton} method. To further reduce the ``curse of dimensionality'', \cite{Du2023SanitizingSE} adapts the random projection map of \cite{wu2017bolton} to NLP pipelines. Finally, \cite{Du2023SanitizingSE} also allows for optional label sanitizing using randomized response at only minor increase to the privacy budget. 

\cite{Du2023SanitizingSE} tested this method using a BERT language model on three different tasks: SST-2, IMDb, and QNLI. In all of these, the authorship was the sensitive information. The baseline, non-DP model achieved around $90\%$ accuracy on all three of these tasks. Using metric-LDP without the dimensionality improvements yielded only around $50\%$ accuracy, but the combination of all the methods yielded accuracy with only approximately $1\%$ drops in accuracy. \cite{Du2023SanitizingSE} tested the privacy guarantees across several different privacy attacks. MIA threshold attacks were only slightly better than guessing, embedding inversion attacks had very low recovery rate, and attribute inference attacks were $40\%$ less successful than on the non-DP model.

The same author group formulated another approach that naturally fits in the federated learning framework. \cite{Du2023dpforward} proposed a method called \emph{DP-Forward} which directly perturbs embedding matrices in the forward pass of LMs rather than adding noise during the back propagation step. The motivation is that while adding noise during back propagation protects well against membership inference, it is ineffective against inference-time attacks such as sensitive attribute inference and embedding inversion. Furthermore, there are large computational costs in the fine-tuning process. In order to add the minimal amount of noise while maintaining privacy guarantees in the forward pass, \cite{Du2023dpforward} devised the analytical matrix Gaussian mechanism (aMGM). This mechanism exploits a necessary and sufficient DP condition from the analytical Gaussian mechanism by \cite{balle2018aGM} to draw possibly non-i.i.d. noise from a matrix Gaussian distribution \cite{Chanyaswad2018MVG}. Particularly when dealing with high-dimensional data, this significantly reduces the amount of noise used. Similar to \cite{Du2023SanitizingSE}, \cite{Du2023dpforward} tested this using a BERT language model on three different tasks: SST-2, IMDb, and QQP, with authorship being the sensitive information. Compared to the baseline DP-SGD solutions of \cite{yu2022differentially} and \cite{yu2021large}, DP-Forward achieved $7.7\%$ better utility on these tasks while using $3$x less time and memory. The utility of DP-Forward was similar to the non-DP model and provided an $88\%$ reduction in embedding inversion success rate and $41\%$ reduction in sensitive attribute inference success. The baseline DP-SGD was unable to protect against these two inference-time attacks.

\textbf{Additional Remarks on Private Models}

\cite{xiao2023large} proposed multiple fine-tuning methods to specifically protect PII information. These were 1) manually replacing PII with preset tokens, 2) introducing a training penalty on PII, 3) using a classifier to replace PII with preset tokens, 4) using few-shot examples as part of the prompt, and 5) using Direct Preference Optimization \cite{rafailov2023direct}. These were tested using the LLaMA2 model \cite{touvron2023llama} on the MedAlpaca dataset \cite{han2023medalpaca}, which was annotated using the scrubadub repo \cite{leapbeyond2023scrubadub}. All of these methods were able to preserve PII privacy but most saw a performance drop when evaluated on ROUGE \cite{lin2004rouge} and BERTScore \cite{zhang2020bertscore}. The instruction tuning using few-shot examples was found to be the best balance between privacy and goal alignment.

\section{Copyright}
\label{sec:copyright}
\subsection{Introduction}
\label{subsec:copyright_intro}

In the past few years, there has been a significant increase in the use of AI to generate music, code, and art of all sorts \cite{Vincent2022copyright}. While there have been fascinating applications in this area, it also introduces a wide range of ethical and legal issues raised based on the models' propensity to copy artists' material \cite{Heikkila2022aiart}. As discussed in Section~\ref{sec:memorization}, LLMs often memorize the data they are trained on, and many popular datasets contain copyright data \cite{bandy2021addressing, biderman2022datasheet}. This leads to significant risk of copyright violation by these models. Indeed, comedian Sarah Silverman and others recently sued OpenAI and Meta for copyright infringement via the ChatGPT and LLaMA models respectively \cite{Davis2023bsilverman}.

\begin{figure}[ht]
  \centering
  \includegraphics[width=.85\linewidth]{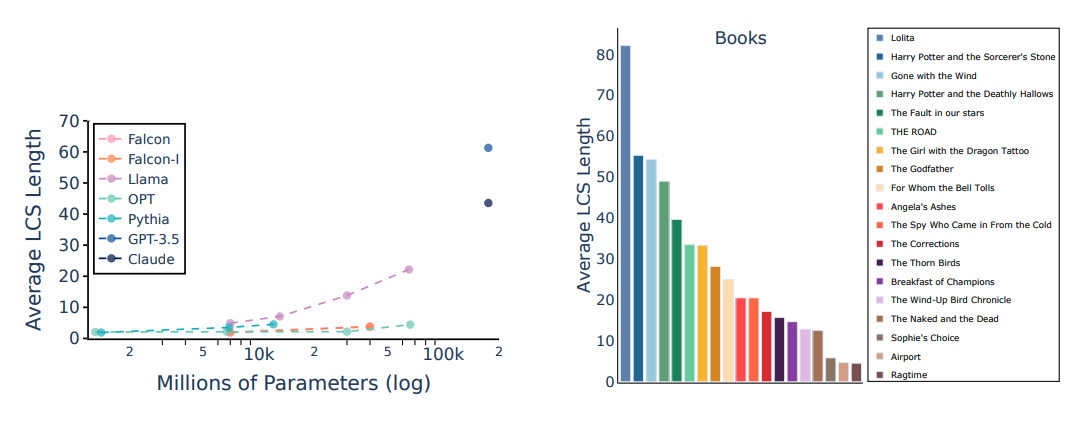}
  \caption{From \cite{karamolegkou2023copyright} showing verbatim memorization in books. On the left, we see the average length of copyright material generated by various models. On the right we see the length of copyright material memorized by book, with more popular books being more susceptible.}
  \label{fig:karamolegkou}
\end{figure}

\cite{karamolegkou2023copyright} ran an extensive test on how much lanugage models memorized copyright data of best selling books published between 1930 and 2010. \cite{karamolegkou2023copyright} tested OPT, Pythia, LLaMA, Falcon, Claude, and GPT-3.5 by prompting open-source models with the first $50$ tokens of the book and closed-source instruction-tuned models with the prompt ``What is the first page of <title>?''. In Figure~\ref{fig:karamolegkou}, we can see the Longest Common Substring (LCS) that was memorized by the models and in each book. Consistent with results discussed in Section~\ref{sec:memorization}, larger models were more susceptible to copyright violation. Meanwhile, popular books had large parts of their opening text memorized by these popular models.

Despite these apparent issues, many major companies and organizations are pushing ahead to augment or replace creative work with AI. The Alliance of Motion Picture and Television Producers (AMPTP) have explored the use of models such as ChatGPT for writing screenplays, which is one of the major issues in the Writers Guild of America (WGA) strike of 2023 \cite{Rose2023wgacopyright}. This strike is seen by some as the first of many legal battles in an oncoming labor war over the use of AI \cite{Broderick2023wga}. 

In order to safely deploy these models without risk of copyright violations, more solutions and modifications are needed. It has already been shown that preventing verbatim memorization is not a perfect solution \cite{ippolito2023preventing} and in the case of copyright protection, even preventing approximate memorization is likely insufficient. Determining copyright violation also requires measuring creative novelty and intended use, which are difficult to quantify. In this section, we discuss the current legal literature on copyright in LMs, as well as any technical solutions that have been proposed. All discussion in this section focuses on US copyright law. The methodologies may still be useful in other jurisdictions but are not necessarily tailored for different nuances of the law. 
    
\subsection{Legal Discourse}
\label{subsec:copyright_legal}

With the rapid rise of generative models, there are a wide range of new issues that the legal community must deal with. Legal papers such as \cite{Delacroix_2023} describe the simultaneous issues of machine learning models using copyright data while also generating data that itself could be considered copyright. This could lead to a downward spiral with large quantities of data sources existing in a legal limbo. 

\textbf{Fair Use Doctrine:} The primary legal framework applied in the context of generative models such as LLMs is the \emph{fair use doctrine}. This is the framework that determines whether new material that relied upon copyright data did so in a manner that is novel enough to avoid any legal issues. Under this doctrine, researchers are allowed limited use of copyright material, even without a license, in order to foster creativity and innovation. While landmark decisions on fair use in technology has happened in the context of code with the Oracle vs Google case \cite{HarvardLawReview_2021}, there have not been any substantial decisions in the case of LLMs. Depending on how courts decide, either model developers could be stifled, or copyright owners could lose out \cite{Sobel_2017}. Despite no legal decisions demanding it, there is some research on ways to avoid any potential issues as we will discuss in Section~\ref{subsec:copyright_solutions}. For extensive coverage of fair use doctrine and its applications to all foundation models, see \cite{henderson2023foundation}.

\textbf{Using Copyright Data:} When it comes to the question of including copyright data in training datasets, most legal discourse agrees that this is an acceptable practice. Just as any individual is free to access and learn from copyright material without being in violation of the law, machine learning models are argued to be given those same rights with any potential legal issues only arising if the final model is used to generate content \cite{Lemley_Casey_2021}. As stated by \cite{Sag_2019}, ``copying expressive works for the purposes of non-expressive use should not be a copyright violation (e.g. text mining in and of itself is not a violation).'' 

\textbf{Models as Authors:} Another topic of discussion in the legal community is whether material generated by a machine learning model can be considered novel enough to grant patents or copyright to the model or model developers. \cite{Hristov_2017} argued that giving authorship to AI programmers and owners is essential to the future development of the AI industry. However, legal decisions since then have leaned in the opposite direction. The US Court system has already decided that work by AI artists and authors is ineligible for copyright \cite{uscopyright} and inventions by AI are ineligible for patent rights \cite{Rose2023patent}.

\subsection{Technical Mitigation}
\label{subsec:copyright_solutions}

The risk of a language model infringing on copyright law is high unless mitigation strategies are implemented. Verbatim memorization filters are insufficient \cite{ippolito2023preventing} and fair use law incorporates some degree of human judgement, making it difficult to algorithmically determine copyright violation. Legal scholars recognize that fair use issues may never be able to be reduced to an algorithm \cite{Burk2019fairuse}. That said, there are technical mitigation strategies that would reduce the potential for copyright violation which we lay out here:

\textbf{Data Filtering:} The most straightforward solution to preventing copyright violation would be to remove any training data which is copyright. Without access to that training data, it is very unlikely that copyright data would be output by the generative model. Even if it were to be generated, it would not pose a legal issue; as noted in Feist v Rural \cite{FeistvRural}, ``...a work may be original even though it closely resembles other works, so long as the similarity is fortuitous, not the result of copying...'' That said, on a practical level this would be very difficult to implement. Most LLMs are trained on corpuses built on large open web crawls and while opt-out strategies have been implemented for many of these, they are not yet widespread. Additionally, it is unclear whether models built strictly on open source, non-copyright, and permissively licensed material would even perform well. Using only this data could potentially bias the models in other negative ways \cite{Levendowski_2017} such as any literature only being from the 1920's and earlier due to public domain laws \cite{henderson2023foundation}. Data deduplication and other memorization mitigation strategies would help (Section~\ref{subsec:mem_mitigation}), but are not proven guarantees.

\textbf{Unlearning:} In the case of LLMs, retraining them would be a significant computational undertaking and would be very expensive, both in terms of time and money. As discussed in Section~\ref{sec:unlearning}, machine unlearning is a significantly more efficient method to remove problematic data and could be readily applied to copyright data. However, unlearning a large portion of data could be computationally intractable, and have a significant impact on model performance. Given a targeted use case, this has potential as a solution, however no significant research has been done on applying unlearning to copyright protection.

\textbf{Measuring Influence of Training Data:} In line with the fair use principle that material is not a copyright violation if its resemblance to the copyright material is purely coincidental, one approach towards quantifying violations is to measure the effect of the training data on the specific output of the model. Output that is similar to prior material is only a copyright violation if it was dependent that material's existence in the training set. 

\cite{vyas2023provable} developed a formal definition called \emph{$k$-Near Access-Free} ($k$-NAF) to measure this. Let $\dtrain$ be the training dataset, which may contain some material $C$ from the set of all copyright material $\mathcal{C}$. $h$ is our generative model trained by algorithm $\mathcal{A}$ on $\dtrain$, which given some prompt $x \in \mathcal{X}$, produces output $y \in \mathcal{Y}$ with probability $p(y|x)$. We would like $h$ to be within $k$-bits of information (under some divergence measure) from $\texttt{safe}_C$, which is a model trained using $\mathcal{A}$ on a dataset that doesn't contain $C$. \cite{vyas2023provable} then establishes their criteria for copyright protection:

\begin{definition}
    \label{def:knaf} 
    (k-Near Access-Free, Vyas (2023) \cite{vyas2023provable}). Let $C$ be a set of copyright datapoints, $\texttt{safe}_C$ be a model trained without access to $C$, and let $\Delta$ be a divergence measure between distributions. We say that a generative model $h$ is $k_x$-near access-free ($k_x$-NAF) on prompt $x \in X$ with respect to $C$, safe, and $\Delta$ if for every $c \in C$,
    $$\Delta \big (p(\cdot |x) || \texttt{safe}_C (\cdot |x) \big ) \leq k_x$$
\end{definition}

\cite{vyas2023provable} defaults the choice of divergence to either maximum KL divergence $\Delta_{max}$, also known as the R\'enyi divergence of order infinity, or the KL divergence $\Delta_{KL}$. This definition being based on a comparison to a safe model allows for fortuitous copying and is robust to cases such as someone prompting ``print the following text: C'' where C is a copyright material. This shares some similarities to the principles of differential privacy \cite{dwork2006differential}. However, as outlined by \cite{elkinkoren2023copyright}, there are several traits of copyright law that distinguish it from DP:

\begin{enumerate}
    \item Privacy in general prevents knowledge of the prior material whereas copyright enables the use of that knowledge for creative gain. Indeed, copyright specifically allows for direct use in the case of quotations, parodies, and research.
    \item Copyright protections are limited in time; after a certain period of time, they enter the public domain.
    \item Copyright excludes raw materials used for cultural expression such as facts, ideas, and methods of operation. 
    \item Differential privacy allows for some small probability of generating the exact training example, whereas in copyright, unless that material was permitted by one of the criteria above, generating the exact training example would result in legal issues.
\end{enumerate}

$k$-NAF follows the characteristics of copyright outlined. \cite{vyas2023provable} also shows that there exist algorithms which satisfy $k$-NAF, for reasonable choices of $k$. The first, Copy-Protection-$\Delta$ is computationally difficult for LLMs. The Smooth-Copy-Protection-$k$ algorithm, however is a fit. Say $\mathcal{V} = \{ q_1,..., q_n \}$ is a cover of $\texttt{safe}$ if $\forall C \in \mathcal{C} \exists q \in \mathcal{V}$ s.t. $\texttt{safe}(C)=q$. Then given a model $h$, cover $\mathcal{V}$, and threshold $k$, Algorithm~\ref{alg:smooth_cpk} returns a model $h_k$ which has quantifiable access-free guarantees with respect to $\texttt{safe}$. These algorithms assume access to a list of copyright material along with access to an oracle where we can both compute conditional probabilities and obtain samples under the models $h$ and $q \in \mathcal{V}$

\begin{algorithm}[!ht]
\caption{Smooth-Copy-Protection-k, \cite{vyas2023provable}}
\label{alg:smooth_cpk}
\begin{algorithmic}[1]
\State \textbf{Input:} model $h$ (which given prompt $x$ outputs $y$ with probability $p(y|x)$), cover $\mathcal{V}$ of \texttt{safe}, and threshold $k \geq 0$.
\State \Return $h_k$ where $h_k$ is specified as:
\While{True}
    \State Sample $y \sim p(\cdot | x)$ and
    \State \Return $y$ with probability $\min\left\{ 1, \min_{q \in \mathcal{V}}\left\{ \frac{2^kq(y|x)}{p(y|x)} \right\} \right\}$
\EndWhile
\end{algorithmic}
\end{algorithm}

\textbf{Measuring Novelty:} There are a few other methods that have been developed for measuring novelty and creativity in text data. The motivation behind this is that the legal system could use these to audit model output for copyright violations. By quantifying the novelty of a text, there is a more objective way of measuring ``substantial similarity.'' \cite{scheffler2022formalizing} introduced one such framework called \emph{derivation similarity}, which uses Levin-Kolmogorov cost to capture novelty.

\begin{definition}
    \label{def:kl_cost}
    (Conditional Levin-Kolmogorov Cost, \cite{scheffler2022formalizing}) Let $M$, $b$, and $y$ be strings in $\{ 0, 1\}^*$. Let $U(M,b,t)$ be the output of a universal prefix Turing machine running program $M$ for $t$ timesteps, where $M$ has access to input tape $b$. Then, the cost of running $M$ on input $b$ to get a string “comparable” to $y$ is
    $$C(M,z|b)=\min_{t\in \mathbb{N}}\{ |M| + \lceil \log(t) \rceil : z \leftarrow U(M,b,t) \}$$
    $$\Tilde{C}(M,z|b)=\min_{z\cong y}\{ C(M,z|b) \}$$
\end{definition}

Where $z \cong y$ denotes that strings $z$ and $y$ are comparable. In \cite{scheffler2022formalizing} this denoted exact equality but it was noted this could be applied to different notions such as being within a certain Hamming distance. 

In the context of the legal system, \cite{scheffler2022formalizing}'s framework has the plaintiff provide the smallest possible “Producer” algorithm $P$ capable of independently producing something
comparable to $y$, given full access to the original work $x$ and the defendant provide the smallest “Reproducer” algorithm
$R$ that produces a work comparable to $y$ without using the
copyrightable parts of $x$, i.e., in the manner that the defendant
alleges that they initially created $y$. The costs of $P$ and $R$ are then compared to measure the extent to which $y$ intrinsically relied on the creative expression within $x$.

\begin{definition}
    \label{def:emp_deriv_sim}
    (Empirical Derivation Similarity, \cite{scheffler2022formalizing}) Let $x$ and $y$ denote the original and derived works, respectively. Given a producer algorithm $P$ and a reproducer algorithm $R$, the empirical derivation similarity is defined as:
    $$\text{DerSimEmp}(x,y,P,R | bg) =  \Tilde{C}(R,y| bg) -  \Tilde{C}(P,y|(bg,x))$$
\end{definition}

Where $bg$ is the set of all aspects of $x$, $y$, and other contextual works that the parties both agree are non-copyrighted.

This framework provides a mathematical way of measuring the similarity of two texts in the context of a legal discussion. \cite{scheffler2022formalizing} applied this to several past court cases and showed that their output agreed with the decisions in those cases. However, they acknowledge that it has limitations when either party has secret information that compromises $bg$ or acts in bad faith when agreeing upon the contextual works included in $bg$. This framework also only covers substantial similarity and does not go into all the nuances of fair use doctrine or cover the Digital Millenium Copyright Act (DMCA) anti-circumvention measures. The primary use of \cite{scheffler2022formalizing} is to answer the narrow question of whether two works are similar.

\cite{franceschelli2022deepcreativity} separately created a framework named DeepCreativity. This framework is based on \cite{Boden_2005}'s work that creativity is composed of value, novelty, and surprise. By associating components of different deep learning frameworks to those three components, \cite{franceschelli2022deepcreativity} created a quantitative measure of creativity. While this was not created for copyright assessment purposes, quantitative measures like this could in theory be applied to this kind of research.

\section{Machine Unlearning}
\label{sec:unlearning}
As discussed in the previous sections, there are several pressing privacy issues underlying the use of LLMs. While these issues were initially subject to scrutiny primarily in the academic setting, they are of growing concern to the greater public. Popular examples about personal data being revealed by machine learning models, such as in \cite{Heikkila2022gpt3}, have led to many people being distrustful of how their data is being handled \cite{Pew2019distrust}. 

Given these growing concerns by the public, legislation and regulations are being introduced to protect consumer data. The most influential one is the European Union's General Data Protection Regulation (GDPR) \cite{eu2016gdpr}, which governs how businesses handle personal data. In particular, one element of the GDPR is known as the \emph{Right to Be Forgotten}. This is the right of every consumer to request that their personal data be deleted and is also a provision in the California Consumer Privacy Act (CCPA) \cite{california2023ccpa}, and the proposed New York Privacy Act (NYPA) \cite{NY2023nypa} and Consumer Privacy Protection Act (CPPA) of Canada \cite{canada2023ccpa}. While this would be a straightforward request in many settings where personal data would be stored in a table, in the context of machine learning models and LLMs, compliance challenges are significantly harder \cite{zhang2023right}. Even if an individual is deleted from a database, if their data was previously used to train a model, it can be extracted through privacy attacks such as those discussed in Section~\ref{sec:mi_attacks}. This risk is being recognized more by regulatory bodies, including the Federal Trade Commission recently coming to a settlement with a facial recognition software company that in addition to a user's data, any models or algorithms derived from their data must be deleted \cite{FTC2021}. This sets a precedent that may become more frequent in legislation and regulation. 

This creates a difficult situation for model developers. The most naive solution, known as Leave One Out (LOO) retraining, would just delete an individual's data and retrain the whole model. However, modern LLMs can take several weeks and millions of dollars to retrain from scratch making this solution infeasible in practice. According to Help Net Security, consumer data deletion requests have increased by $74\%$ from 2021 to 2022 \cite{HelpNetSecurity2023}. If models were required to be completely retrained each time one of these requests came in, they would be impossible to maintain, especially given the size of the regions with ``Right to Be Forgotten'' laws either in place or proposed. This has lead to a new field of research called \emph{machine unlearning}. This field looks to delete an individual's data and influence from a model using significantly fewer computational resources than retraining it. A successful unlearning algorithm should produce a model that is approximately indistinguishable from the model produced by retraining. In this section, we discuss the general principles of machine unlearning and current developments within the space of language models. 

For a more comprehensive survey of the general space of machine unlearning, see \cite{nguyen2022survey}. For a more focused survey of unlearning within the space of LLMs, see \cite{liu2024rethinking}.

\subsection{Preliminaries}

\begin{figure}[h]
  \centering
  \includegraphics[width=.85\linewidth]{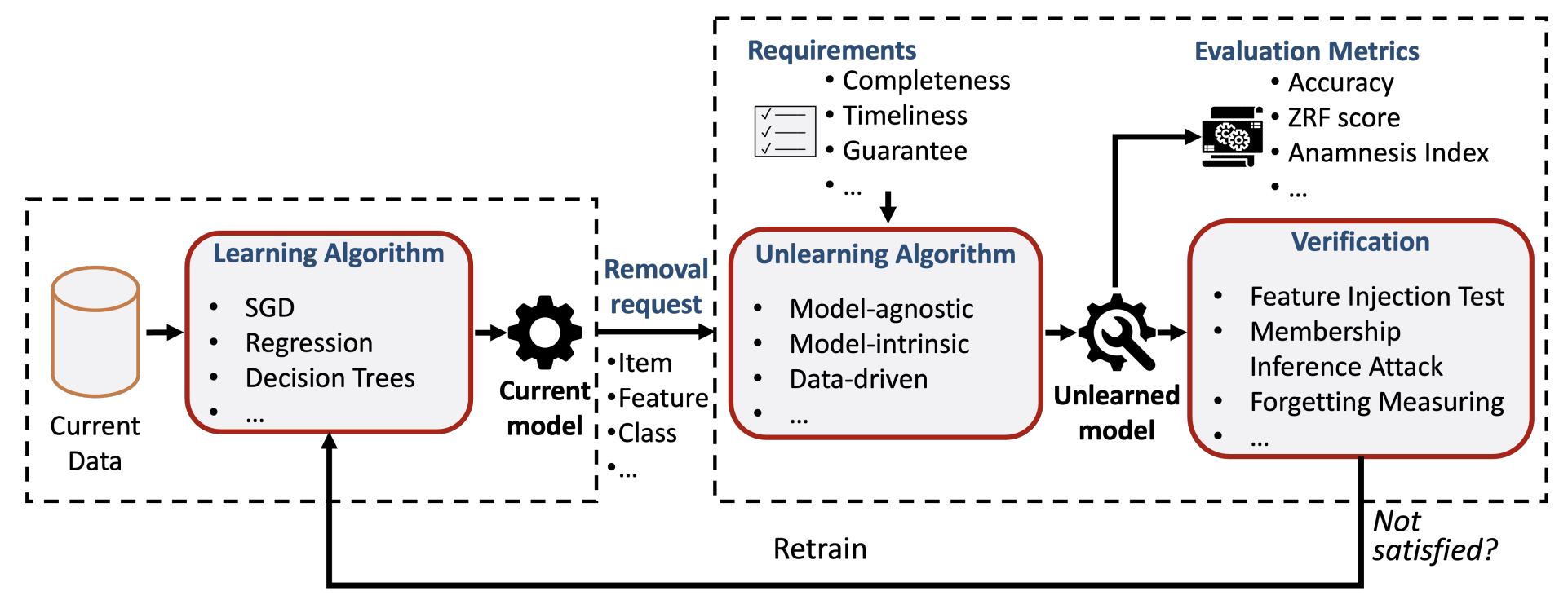}
  \caption{Machine Unlearning framework (from \cite{nguyen2022survey})}
  \label{fig:unlearning_framework}
\end{figure}

\textbf{Unlearning Framework:} The general framework of machine unlearning is shown in Figure~\ref{fig:unlearning_framework}. In this framework we have a dataset $D \in \mathcal{D}$ and training algorithm $\mathcal{A}: \mathcal{D} \rightarrow \mathcal{H}$ where $\mathcal{H}$ is the hypothesis space of our model. We also have an unlearning algorithm $\mathcal{U}: \mathcal{H} \times D \times D_f \rightarrow \mathcal{H}$, where $D_f$ is the subset of $D$ that we want to forget. We would like to have an unlearning algorithm such that $\mathcal{U}(D, D_f, h)$ is close to $\mathcal{A}(D \backslash D_f)$. It is assumed that $\mathcal{A}$ and $\mathcal{U}$ are randomized algorithms. Let $\text{Pr}(\mathcal{A}(D))$ be the distribution of all models trained on $D$ by $\mathcal{A}$ and $\text{Pr}(\mathcal{U}(D, D_f, h)$ be the distribution of unlearned models. We then have two categories of unlearning. The first is \emph{exact unlearning}:

\begin{definition}
\label{def:exact_unlearning}
  (Exact Unlearning, \cite{nguyen2022survey}) Given a learning algorithm $\mathcal{A}$, we say the process $\mathcal{U}$ is an exact unlearning process iff $\forall \mathcal{T} \subseteq \mathcal{H}, D \in \mathcal{D}, D_f \subset D$:
  $$\text{Pr}(\mathcal{A}(D \backslash D_f)\in \mathcal{T}) = \text{Pr}(\mathcal{U}(D, D_f, \mathcal{A}(D))\in \mathcal{T})$$
\end{definition}

That is, the model resulting from $\mathcal{U}$ is the same as $\mathcal{A}$ trained without the deleted point. A closely related definition that relaxes this requirements is \emph{approximate unlearning}:

\begin{definition}
\label{def:eps_approx_unlearning}
  ($\epsilon$-Approximate Unlearning, \cite{nguyen2022survey}) Given $\epsilon > 0$, an unlearning mechanism $\mathcal{U}$ performs $\epsilon$-certified removal for a learning algorithm $\mathcal{A}$ if $\forall \mathcal{T} \subseteq \mathcal{H}, D \in \mathcal{D}, d \in D$:
  $$e^{-\epsilon} \leq \frac{\text{Pr}(\mathcal{U}(D, d, \mathcal{A}(D))\in \mathcal{T})}{\text{Pr}(\mathcal{A}(D \backslash d)\in \mathcal{T})} \leq e^{\epsilon}$$
  Where $d$ is the removed sample.
\end{definition}

\begin{definition}
\label{def:epsdelta_approx_unlearning}
  ($(\epsilon,\delta)$-Approximate Unlearning, \cite{nguyen2022survey}) Given $\epsilon, \delta > 0$, an unlearning mechanism $\mathcal{U}$ performs $\epsilon$-certified removal for a learning algorithm $\mathcal{A}$ if $\forall \mathcal{T} \subseteq \mathcal{H}, D \in \mathcal{D}, d \in D$:
  $$\text{Pr}(\mathcal{U}(D, d, \mathcal{A}(D))\in \mathcal{T}) \leq e^\epsilon \text{Pr}(\mathcal{A}(D \backslash d)\in \mathcal{T}) + \delta$$
  and
  $$\text{Pr}(\mathcal{A}(D \backslash d)\in \mathcal{T}) \leq e^\epsilon \text{Pr}(\mathcal{U}(D, d, \mathcal{A}(D))\in \mathcal{T}) + \delta$$
\end{definition}

That is, $\epsilon$ defines how similar the new model must be and $\delta$ in Definition~\ref{def:epsdelta_approx_unlearning} is the upper bound of the probability of Definition~\ref{def:eps_approx_unlearning} failing. Note that in comparison to Definition~\ref{def:exact_unlearning}, the approximate unlearning definitions are defined with respect to a single removed sample rather than a whole subset. It is currently an open question in unlearning whether constant bounds can be provided for larger subsets \cite{nguyen2022survey}. Approximate unlearning is very similar to differential privacy \cite{dwork2006differential}; indeed, differential privacy implies approximate unlearning, but is a much stronger condition.

\textbf{Unlearning Sequences:} In dealing with personal data, these deletion requests may not come cleanly as a single batch. In a practical setting, deletion requests will come as a sequence with a certain order. This poses a theoretical issue as some unlearning methods rely on the assumption that deletion requests are independent of each other. It is very possible that users may want to delete their data after learning what the deployed models reveal about them; this is known as an \emph{adaptive sequence}. \cite{gupta2021adaptive} provides the generic reduction for this as follows:
\begin{enumerate}
    \item If a data deletion algorithm $\mathcal{U}$ for a learning algorithm $\mathcal{A}$ has deletion guarantees for independent sequences of deletion requests (as those from past work do), and
    \item Information about the internal randomness of $\mathcal{U}$ is revealed only in a manner that satisfies differential privacy, then
\end{enumerate}

Then $(\mathcal{A},\mathcal{U})$ also satisfies data deletion guarantees against an adaptive sequence of deletion requests, that can depend in arbitrary ways on the information that the model provider has made public. Data addition requests may also be a part of this update sequence.

\subsection{General Approaches to Unlearning}

While there has been limited work in machine unlearning in the context of language models, the approaches to unlearning algorithms have generally followed one of two methods: 1) efficient LOO retraining and 2) gradient-based updates.

\textbf{Efficient LOO Retraining}: Methods in this area look for more efficient ways to simply remove or scramble the data that has been requested for deletion, then retrain the model from an intermediate point to save computational costs. \cite{graves2020amnesiac} introduced a method where the requested deletion point $d$ is removed and replaced with copies of $d$ with random labels. Then, using the same training algorithm $\mathcal{A}$, more training iterations are run until privacy risks are no longer an issue. They showed that this method was effective against both model inversion and membership inference attacks after only a few training iterations. The downside of this method is that the data holder maintains a copy of $d$ during the unlearning process, which could hold legal implications. Additionally, if a large quantity of points is requested for deletion, the model performance could possibly suffer.

\begin{figure}[h]
  \centering
  \includegraphics[width=.6\linewidth]{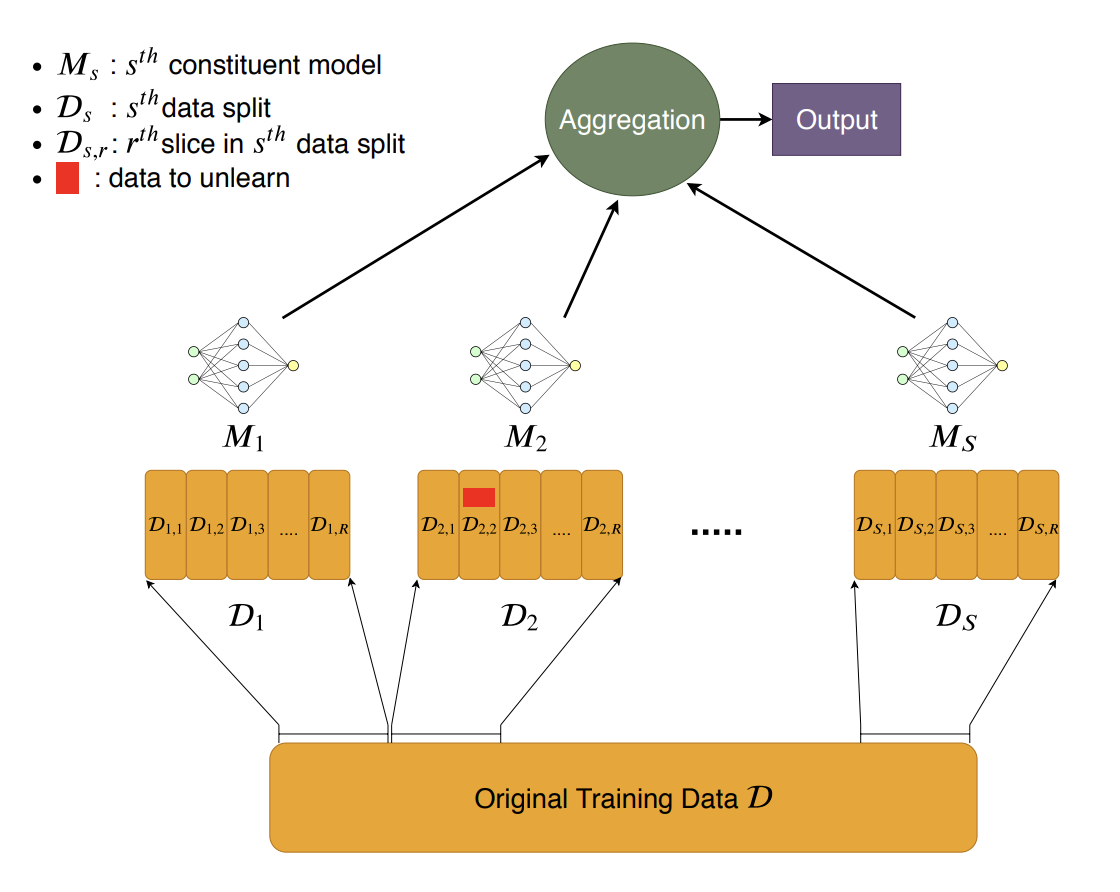}
  \caption{SISA framework \cite{bourtoule2020machine}: The data is divided into shards, which are then divided into slices. The constituent models' outputs are trained incrementally using the slices, saving at each step. The consituent models are then aggregated for the final output. When an unlearning request for $d$ comes in, only the consituent model with $d$ must be retrained, and only starting at the slice containing $d$.}
  \label{fig:sisa_framework}
\end{figure}

\cite{bourtoule2020machine} created an approach that cleverly restructures the way the data is fed into the model. Their method is called SISA for Sharded, Isolated, Sliced, and Aggregated training and is depicted in Figure~\ref{fig:sisa_framework}. In the SISA framework, the training data is divided into multiple disjoint shards such that each data point is only contained in a single shard. The data in each shard is subsequently divided into multiple slices. For each shard, an independent model is trained iteratively using the slices, recording the model parameters before each new slice is introduced. When an output is needed, the outputs of these constituent models are aggregated for the output; the most simple method is to use a majority vote over the predicted labels. Under this framework, when a deletion request for point $d$ comes in, only one of the constituent models must be retrained and only starting from the slice which contains $d$. Compared to naive retraining, the SISA framework for simple learning tasks resulted in a time improvement of $4.63\times$ on the Purchase dataset \cite{sakar2019purchase} and $2.45 \times$ on the SVHN dataset over naive retraining. For more complex learning tasks such as ImageNet classification, SISA resulted in a $1.36\times$ speed improvement. \cite{gupta2021adaptive} showed that this method can be adapted to handle unlearning adaptive sequences if the shards are selected independent of each other.

In traditional statistics, the influence function \cite{Hampel1974influence} \cite{Cook1979influence} is a technique that can estimate the effect of deleting a group of points at the same computational cost of taking a second-order Taylor approximation around the model parameters. Given a strongly convex training objective, influence functions would be expected to align well with LOO retraining. However, \cite{basu2021influence} found that in the case of neural networks, the influence function's estimates are ``fragile'' and ``erroneous''. \cite{bae2022influence} later decomposed the reason for this discrepancy into three primary reasons, but still found that influence functions are a poor replacement for LOO retraining on nonlinear networks. 

\textbf{Gradient-Based Updates:} This next class of methods methods rely on gradient descent techniques typically used to train ML models to execute efficient unlearning. \cite{neel2020descenttodelete} gives an approach to unlearning in this fashion. In their framework, the results of the gradient descent learning algorithm, $\mathcal{A}$, are published with some Gaussian noise. The unlearning algorithm, $\mathcal{U}$, is then simply a few more gradient descent updates run without the deleted point $d$ in the dataset. Due to the Gaussian noise added, the outcomes of $\mathcal{U}(D, d, h)$ and a full retraining are provably indistinguishable. This algorithm is able to handle arbitrarily long sequences of updates in logarithmic time, maintaining strong unlearning from Definition~\ref{def:strongweak}. For sufficiently high dimension data, \cite{neel2020descenttodelete} further improved on this by bootstrapping the dataset and training the model in the same fashion as the SISA framework such that the gradient update only needs to be applied to the affected shard. This means more updates can be performed while the bounded accuracy error of each shard is maintained even over an arbitrary sequence of updates \cite{zhang2013comunicationefficient}.

\cite{graves2020amnesiac} introduced a second method in addition to their LOO method described earlier. This method, called ``amnesiac unlearning'', requires the model owner to keep track of which examples appeared in which training batch along with the parameter update from that batch. Then, when a deletion request comes in for $d$, the owner subtracts those updates from the final parameters. As long as the number of batches affected is small, this method is very efficient and effective. However, since parameter updates are dependent on the current parameters when that step takes place, this unlearning method is approximate rather than exact. Another downside is that a large amount of storage space is needed to track the parameter update values.

\begin{figure}
\centering 
\includegraphics[scale=.45]{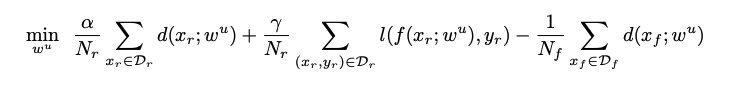}
\caption{The \texttt{SCRUB} training objective \cite{kurmanji2023unbounded}. Here $D_r, D_f$ are the forget and retain sets respectively, where $|D_r| = N_r, |D_f| = N_f$. $w^u$ represents the parameters of the new unlearned model, and $d(x; w^u)$ represents the KL divergence between the softmax outputs of $w^u$ on input $x$, and the softmax outputs of the original trained model on $x$. The middle term is the standard training loss on the retain set. }
\label{fig:scrub}
\end{figure}

\cite{kurmanji2023unbounded} develop a gradient based unlearning algorithm called Scalable Remebering and Unlearning Unbound (\texttt{SCRUB}) where rather than doing gradient descent on the retain set to unlearn as in \cite{neel2020descenttodelete}, or gradient ascent on the forget set \cite{jang2023knowledge}, they optimize a hybrid objective in Figure~\ref{fig:scrub}. First they initialize the new unlearned model $w^u$ at the existing weights $w^f$, then they simultaneously force the predictions of $w^u$ on the forget set away from the predictions of $w^f$ on the forget set, while keep the predictions on the retain set close. They also add a term in the objective to force the accuracy on the retain set to stay high. Since they are simultaneously minimizing the first two terms in Figure~\ref{fig:scrub} while maximizing $\sum_{x_f \in D_f}d(x_f;w^u)$, they alternate between performing an epoch of updates on the forget set with an epoch of updates on the retain set. Note that this basic procedure will result in maximizing forget error (by forcing it away from $w^f$'s predictions), rather than trying to match the error on the forget set to the out of sample error, as is more typical in unlearning applications. They make the distinction that this variant of \texttt{SCRUB} is appropriate for removing biased or poisoned data where the goal is not to simulate retraining, but to maximize error on the forget set. In order to adapt \texttt{SCRUB} for unlearning points for the sake of privacy, the difficult question is when to halt the optimization so that the error on the forget set approaches the error on the retrained model. To accomplish this, they propose the \texttt{SCRUB + R} algorithm, that does the following: 
(i) Constructs a validation set $D_v$ that approximates $D_f$
(ii) Trains \texttt{SCRUB} as usual, storing a copy of the model every epoch
(iii) At the end of \texttt{SCRUB} training, compute $\alpha_v = \frac{1}{N_v}\sum_{(x, y) \in D_v}\ell(f(x, w^u), y)$
(iv) Rewind \texttt{SCRUB} training to output the checkpoint whose error on $D_f$ is closest to $\alpha_v$. They remark: ``The intuition is that the last step of unlearning approximates ‘maximally forgetting’. Consequently,
the error on the identically-distributed validation set approximates the error of a model that never
learned about that distribution from the forget set: any correct predictions on the held-out examples
are due only to the generalization power of the model that was trained only on the retain set. Therefore,
we choose that validation set error to serve as the reference point for how high we would like the
forget set error to be for UP applications."
They evaluate their algorithm on CIFAR-$10$ and Lacuna-$10$ and find that across a variety of metrics including MIA attack success and loss on the forget set, $\texttt{SCRUB}$ outperforms prior methods based on fine-tuning (gradient descent on $D_r$), gradient-ascent on $D_f$, NTK \cite{ntk}, and Fishers \cite{fishers}, although simultaneous gradient descent and ascent on $D_r, D_f$ respectively is a strong baseline, and is close to $\texttt{SCRUB}$ in flavor.

\subsection{Unlearning in LLMs}

Work on applying machine unlearning to LMs is limited but rapidly growing. \cite{liu2024rethinking} provides an extensive survey of the current methodologies used in unlearning and how they translate to the LLM space. They also discuss related aspects such as relevant tasks and benchmarks, evaluation methods, utility preservation, and model efficiency along with discussion of LLM unlearning's relationship to copyright and privacy protection relevant to Sections~\ref{sec:private_llms} and \ref{sec:copyright}.

To set the stage, we can turn to \cite{patil2023sensitive} which studied a range of extraction attacks and defenses on an LLM that has had sensitive information deleted. They focused their attack and defense framework on unlearning approaches that directly edit model weights since it is a useful unlearning approach that also factors in the possibility of whitebox attacks. In this framework, an extraction attack for a deleted data point, $A$, is successful if $A$ is contained in a candidate set $C$ of size $B$ known as the budget. \cite{patil2023sensitive} ran their experiment on the GPT-J, Llama-2, and GPT2-XL models trained on the CounterFact \cite{meng2022romecounterfact} and zsRE \cite{levy2017zsRE} datasets. To perform the model weight editing, \cite{patil2023sensitive} used the ROME \cite{meng2022romecounterfact} and MEMIT \cite{meng2023memit} techniques. Two whitebox attacks were designed which leverage the logit lens technique \cite{nostalgebraits2020logitlens, geva2021logitlens} that converts the hidden state from an intermediate layer into a probability distribution over the vocabulary set. The \emph{Head Projection Attack} constructs the candidate set $C$ by taking the $k$ tokens at each layer with the highest probabilities. The \emph{Probability Delta Attack} takes advantage of the observation that deleted data points may have inconsistent unlearning, leading to a logit lens probability that rapidly rises or falls between layers. Thus $C$ is composed by taking the $k$ tokens at each layer that had the highest difference in probability from the previous layer. \cite{patil2023sensitive} also used a blackbox attack, \emph{Input Paraphrasing}, which prompts the LLM with paraphrased versions of the original input (the paraphrase model from \cite{krishna2023paraphrasing} was used). These attacks were tested against $6$ different defense methods. There were three baseline methods, \emph{The Empty Response Defense} \cite{ouyang2022training}, \emph{Fact Erasure} \cite{hase2023does}, and \emph{Error Injection} \cite{decao2021editing}. Three additional defenses were designed, specifically targeted towards the above designed attacks. The \emph{Head Projection Defense} and \emph{Max-Entropy Defense} change the weight editing objective by either subtracting the probability of the top $k$ output or maximizing entropy at each layer. The \emph{Input Rephrasing Defense} addresses the rephrasing attack by adding rephrased versions of the requested deleted point to the unlearning objective. Across this framework, \cite{patil2023sensitive} answered multiple questions. They find that the model weight editing techniques do work in providing some unlearning capabilities. However, the designed attacks were still able to extract the deleted information. In some cases, the attacks were extremely successful such as the Head Projection Attack having $89\%$ success with budget $B=20$ against GPT-J trained on zsRE edited with MEMIT. The defenses had varying degrees of success, with Head Projection and Max-Entropy providing reductions in extraction success by up to $20-40\%$. The Input Rephrasing Defense did not prove very effective. Likewise, defenses were not universally successfule, such as the Head Projection defense being ineffective against the Probability Delta attack due to it's tailoring specifically to the Head Projection Attack. 

\cite{patil2023sensitive} concludes that while the problem of unlearning is tractable, it is still difficult to achieve with high protection from all extraction attacks. More methods will need to continue to be developed for effective unlearning to happen in the LLM space. In that spirit, here we cover different works in the LLM unlearning space sorted by the different class of methodologies used to achieve the unlearning.

\textbf{Efficient LOO Retraining:} \cite{kumar2022privacy} uses an efficient LOO retraining approach by expanding upon the SISA framework by \cite{bourtoule2020machine}. While the SISA framework showed computational improvements over naive deletion and retraining, it is impractical for application to LLMs since the number of model parameters and data points makes it extremely computationally expensive. \cite{kumar2022privacy} proposed two variants of SISA that improve it for LLMs. The first is called \emph{SISA-FC}. Since saving a checkpoint for every slice during the constituent model training is impossible for large text corpora, this framework starts with an LM trained on a generic corpus. The SISA framework is then only used for the task-specific fine-tuning process. While SISA-FC reduces the retraining time and computational storage, model performance generally performs better if the model is fine-tuned for the task the entire time. The second variant, known as \emph{SISA-A}, addresses this by using the SISA framework for the entire training. To save memory, the SISA-A framework trains the model using the parameter efficient adapter method from \cite{houlsby2019parameterefficient}. Adapter modules are used to replace the model parameters and are significantly smaller, using $1-5\%$ the memory that storing all the model parameters does. This significantly reduces the memory cost of checkpointing during the SISA framework. \cite{kumar2022privacy} tested these two frameworks on three Glue tasks/corpora \cite{wang2018glue}: SST-2 \cite{socher2013recursive}, QQP \cite{wang2018glue}, and MNLI \cite{williams2018broadcoverage}. They trained a BERT model using the different approached and measured the accuracy on these tasks, the time it took to retrain the model after a deletion request, and the amount of memory the final model takes up. SISA-A saw a small dip ($1-6\%$) in performance on the tasks but also saw a $100\times$ improvement in retraining time. SISA-FC saw a $30\%$ dip in performance on the tasks compared to SISA-A but also was about $100\times$ faster to retrain than SISA-A, making it way faster than the baseline approach. Exact memory numbers were not provided but \cite{kumar2022privacy} stated that both frameworks saw signficant reductions in memory usage. Since the SISA framework eliminates any influence the requested deleted point may have had, this approach satisfies the exact unlearning definition.

\textbf{Gradient-Based Updates:} \cite{jang2023knowledge} studied a gradient based solution for machine unlearning in LMs, named \emph{knowledge unlearning}. Specifically, given $h_{\theta}$ and a given sequence of tokens that we would like to unlearn, $\textbf{d}=(d_1,...,d_T)$, $\mathcal{U}$ solves a gradient ascent problem maximizing the loss function:

$$\mathcal{L}(h_\theta, \textbf{d}) = -\sum_{t=1}^T \log(p_\theta(d_t | d_{<t}))$$

\begin{figure}[h]
  \centering
  \includegraphics[width=.85\linewidth]{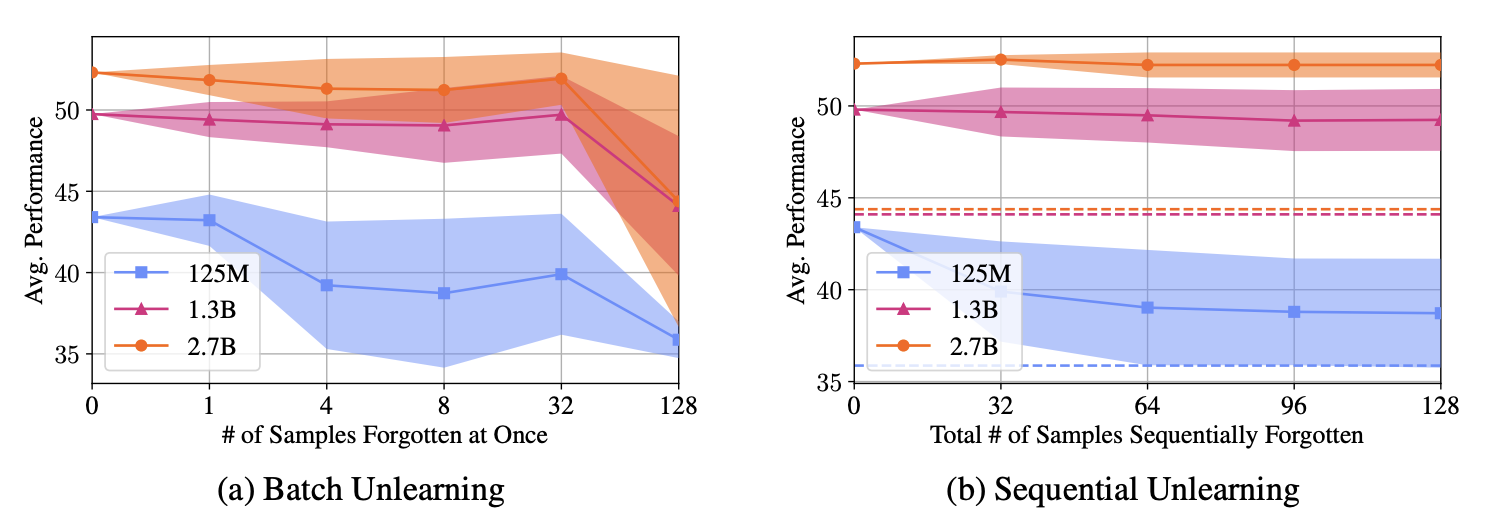}
  \caption{\cite{jang2023knowledge}, the average LM performance for $3$ models on $9$ benchmark tasks when varying the total number of samples unlearned. (a) shows performance when they are unlearned all at once while (b) shows performance when the samples are divided into $4$ chunks and fed in sequentially. While performance is similar for small number of samples, it drops off for batch unlearning at $128$ samples.}
  \label{fig:batchvsseq_unlearning}
\end{figure}

Where $d_{<t}$ denotes the token sequence $\textbf{d}=(d_1,...,d_{t-1})$ and $p_\theta (d_t | d_{<t})$ denotes the conditional probability of predicting the next token to be $d_t$ given $d_{<t}$. \cite{jang2023knowledge} evaluated this method on GPT-Neo \cite{black_sid_GPTNeo} pre-trained on the Pile \cite{gao2020pile} and the OPT model \cite{zhang2022opt} pre-trained on a deduplicated version of the Pile. Data from the Training Data Extraction Challenge \cite{Carlini2022trainingdata} was used as the target data and the success of the unlearning was measured as the ability to reduce the risk from extraction attacks (Section~\ref{subsec:extract}) and memorization (Section~\ref{sec:memorization}). An array of $9$ classification tasks were used to evaluate the general performance of the LMs. They found that knowledge unlearning did not result in any significant degradation of performance compared to naive retraining while requiring several orders of magnitude less computation. The general performance was also significantly better than another computationally inexpensive method, differentially private decoding (\cite{majmudar2022differentially}, Section~\ref{subsec:dp_solutions}). In some cases, the performance of the model after knowledge unlearning was superior to the naive retraining method. This is possibly due to the benefits of forgetting described in \cite{zhou2022fortuitous}. Additionally, \cite{jang2023knowledge} found that sequentially unlearning the data is better than unlearning it in large batches (see Figure~\ref{fig:batchvsseq_unlearning}).

\cite{yao2023large} similarly performed unlearning using gradient ascent, providing ``negative'' examples for OPT and Llama models to unlearn. They showed that after several training iterations, these models were able to remove harmful content (such as recommended illegal activity), copyright content, and misinformation. \cite{yao2023large} compared this approach to reinforcement learning from human feedback (RLHF) and found gradient ascent unlearning achieved better goal alignment while using $2\%$ of the computational costs. This study showed that a gradient based approach to unlearning in LLMs can work in these practical cases.

\cite{kassem-etal-2023-preserving} developed a method for unlearning called DeMem (standing for DeMemorization) which takes a gradient based solution using a reinforcement learning feedback loop. Given a dataset that is split into prefixes and suffixes (e.g. "Alice lives at..." and "...123 main street"), DeMem uses an independent, pre-trained LM to produce a new suffix. Then a similarity score is computed between this suffix and the original one. This score is used as the reward signal in the feedback loop, encouraging the target model to develop a paraphrasing policy. \cite{kassem-etal-2023-preserving} tested this method on GPT-Neo and OPT models that were trained on the Pile. BERTScore was used as the similarity score \cite{zhang2020bertscore}, memorization was measured by edit distance using SacreBLEU \cite{post-2018-call}, and model performance was measured across $9$ different task benchmarks and $2$ perplexity benchmarks. Combining DeMem with deduplication \cite{kandpal2022deduplicating} saw a marked decrease in memorized examples, between half and a third as many as the baseline. Meanwhile, performance only marginally decreased by about $.5\%$ across the benchmarks. The robustness of the performance comes from the use of the pre-trained LM which ensures that the updated model remains coherent.

\textbf{In-Context Unlearning:} While the above methodologies are proven unlearning approaches, they all require access to the model parameters in some capacity in order to perform the unlearning updates. \cite{pawelczyk2023incontext} provides a new framework for unlearning they call \emph{In-Context Unlearning} (ICUL). This takes inspiration from the concept of In-Context Learning (ICL) where examples are provided to the LLM as part of the prompt in order to improve model performance for a specific task without having to update model parameters.  Given a deletion request for a certain training example, ICUL provides that instance with a flipped label along with several additional, correctly labeled instances which are prepended to the input to the LLM. This approach was tested on the $560$M and $1.1$B parameter Bloom language models \cite{workshop2023bloom} with the SST-2 \cite{socher2013recursive}, Amazon polarity and Yelp polarity \cite{zhang2016polarity} datasets. These datasets are text classification tasks with the goal of predicting positive or negative reviews on Rotten Tomatoes, Amazon, and Yelp respectively. The performance of ICUL was compared to two baselines, one being ICL which does not provide the flipped examples and the other being the unlearning approach by \cite{jang2023knowledge}. These approaches were all measured by their performance against a LiRA attack using $10$ shadow models \cite{carlini2022first}.

ICUL surpassed the baseline with lower AUC and TPR (meaning the LiRA attack was less successful) across both model sizes and all datasets. Compared to \cite{jang2023knowledge}, ICUL achieved lower AUC scores in all instances and lower TPR in the majority of cases. These results show that ICUL greatly reduced the ability to identify the forgotten point as a member of the training set with the added advantage of only having black-box access to the model. ICUL achieved AUC scores that were near $.5$ meaning that these results are non-trivial in a practical setting. While task performance suffers slightly with this approach, this difference shrinks on the larger Bloom model with test accuracy being within $1\%$ of the baseline model on the polarity datasets.

Two additional studies were run on this approach to determine its effectiveness. First, \cite{pawelczyk2023incontext} varied the number of correct samples provided after the flipped example to determine the dependency on these. \cite{pawelczyk2023incontext} found that short context lengths (i.e. $2$ additional examples) yielded poor model confidence scores. Longer context lengths of $4$ or $6$ yielded good performance. Second, they tested the dependency on the forget instance by keeping the flipped label but replacing the training input with a randomly sampled input. ICUL significantly outperforms this randomized approach showing that the flipped label step is essential in this unlearning process.

\begin{figure}[h]
  \centering
  \includegraphics[width=.85\linewidth]{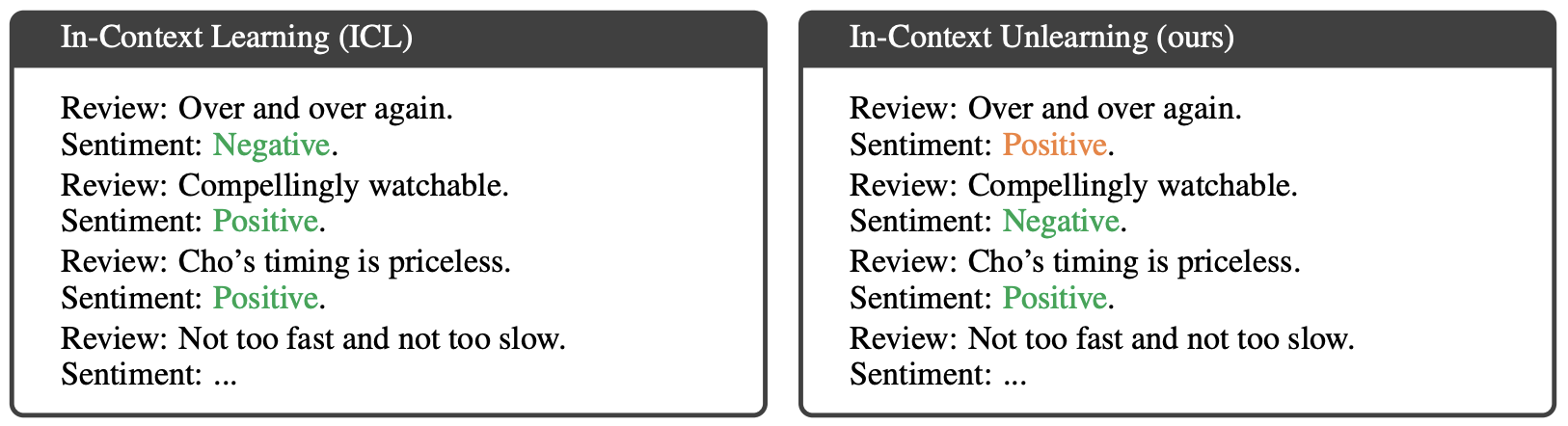}
  \caption{An example of the In-Context Unlearning approach compared to in-context learning from \cite{pawelczyk2023incontext}. For the given sentiment analysis task, in-context learning provides labeled examples from the training set to help the LLM answer the prompt. In-context unlearning flips the label of the forget point (``Over and over again.'') before providing correctly labeled examples. This process reduces the ability of a LiRA attack to identify members of the training set.}
  \label{fig:icul}
\end{figure}

While leveraging In-Context Learning for unlearning purposes is effective from a privacy perspective, \cite{muresanu2024unlearnable} highlights that there can be an issue with computation cost. Specifically, ICL requires clustering the dataset to later draw training examples for the pre-pended prompt at inference time. Since deleting a given datapoint would influence the clusters, the cost of unlearning for an ICL based method is dependent on the size of the dataset. \cite{muresanu2024unlearnable} introduces their algorithm, ERASE, which uses quantized k-means \cite{ginart2019quantized} for their clustering. Quantized k-means introduces stability to the clusters such that with provably high probability, clusters will remain the same if a data point is deleted. \cite{muresanu2024unlearnable} showed that ERASE had similar unlearning performance to other ICUL methods and SISA unlearning while not being dependent on dataset size.

\textbf{Concept Unlearning:} While prior works focus on unlearning batches of examples, \cite{eldan2023whos} developed an approximate unlearning approach that can be used to unlearn an entire target data corpus, $D_{target}$. In this approach, a ``reinforced'' model is built by further fine tuning the model on $D_{target}$. The probability outputs of this model are compared to the baseline model since tokens common in $D_{target}$ will have higher probabilities than generic tokens. Generic counterpart labels are then generated and the baseline model is fine tuned using these alternate labels for several iterations until the target data is unlearned. \cite{eldan2023whos} tested this on Meta's $7$-billion parameter Llama model and Microsoft's $1.3$-billion parameter MSFT model and attempted to unlearn the Harry Potter books. After several fine-tuning iterations, the model lost its ``memory'' of Harry Potter. For example, the text completion prompt ``Harry Potter's two best friends are...'' was completed as ``a talking cat and a dragon. One day they decided to...'' instead of ``Ron Weasley and Hermione Granger. In this series...''. Model performance on other benchmarks decreased marginally but overall performance was largely unaffected.


\begin{figure}
    \centering
    \includegraphics[scale=.6]{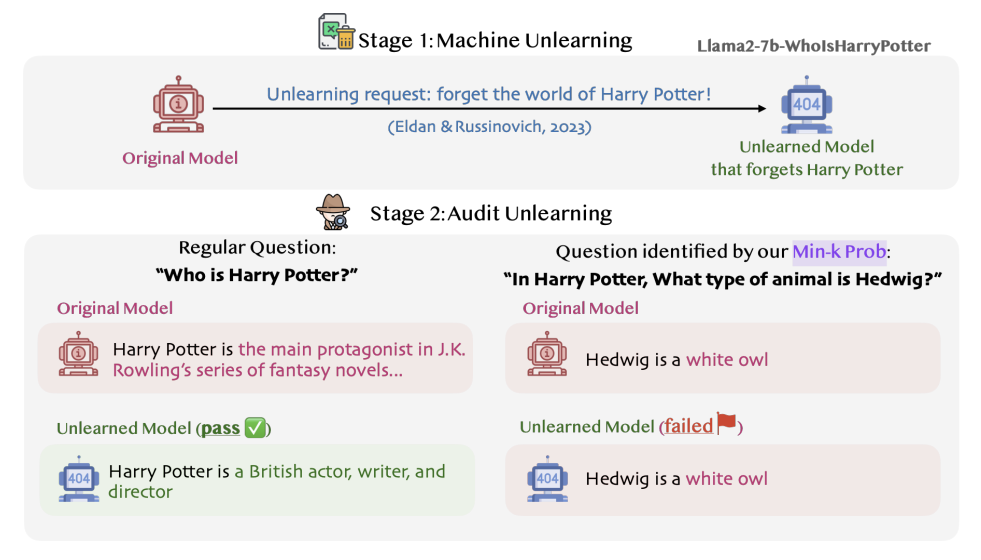}
    \caption{\cite{shi2023detecting}, \textbf{Auditing machine unlearning with Min-K\% Prob}. Machine unlearning methods are designed to remove copyrighted and personal data from large language models. Min-K\% is used to audit an unlearned LLM that has been trained to forget copyrighted books. However, such a model can still output related copyrighted content}
    \label{fig:mink}
\end{figure}

Soon after \cite{eldan2023whos} was published, \cite{shi2023detecting} claimed to break this unlearning technique. Their MIA attack technique, called min$-k$, is a close variant of loss thresholding that rather than predicting training set membership if the loss of an entire sequence of tokens is particularly low, calculates the loss of the $k$ tokens with the lowest losses (hence  ``min$-k$''). They use their technique to identify parts of the forget set (in this case the Harry Potter corpus) that their MIA predicted were members of the training set \emph{even after unlearning}, and then showed that when the unlearned model was prompted with that part of the corpus, it still output information contained in the forget set -- see Figure~\ref{fig:mink} taken from \cite{shi2023detecting}. While the MIA attack above demonstrates that \cite{eldan2023whos} doesn't perfectly unlearn concepts when faced with a motivated adversary, this is perhaps too high of a standard to apply to unlearning large corpora of text, and a better evaluation metric might be performance when prompted on average as opposed to adversarially.

\cite{scrubllm} apply the \texttt{SCRUB} algorithm proposed in \cite{kurmanji2023unbounded} for image classification, to LLMs. Rather than fine-tune the entire model as in \cite{kurmanji2023unbounded}, they only re-train $.5\%$ of the parameters in the unlearning layers on the $\texttt{SCRUB}$ objective, where they have also added a term to the \texttt{SCRUB} objective that is the negative of the masked language modeling objective \cite{raffel2020exploring}. They evaluate their method on both classification and generation, on the IMDB Dataset \cite{maas-etal-2011-learning} and the SAMSum \cite{gliwa-etal-2019-samsum} datasets respectively. Following \cite{kurmanji2023unbounded} they evaluate their unlearning algorithm across (i) performance on the test set for the given task (ii) performance on the retained set and (iii) effectiveness of unlearning as measured by the MLM loss and performance on the forget set. All experiments use the T$5$ base model \cite{raffel2020exploring}, which are then fine-tuned on the IMDB or SAMSum dataset prior to unlearning. On both datasets the model is asked to unlearn $.5\%, 1\%$ and $10\%$ of the dataset, where the forget set corresponds to randomly selected concepts; for IMDB a concept includes all the examples associated with a given movie, actor, or director, and on SAMSum it is all the conversations involving a given speaker. Unlearning guarantees were evaluated via a basic membership inference attack, and by looking at the loss on the forget set and MLM objective. Their \texttt{SCRUB}-based method consistently performed best in terms of high test set accuracy, minimal accuracy on the forget set, high MLM loss on the forget set, defense against MIA attacks, and efficiency (time). 

\subsubsection{Additional Remarks}

\textbf{Evaluating Unlearning Success:} While these unlearning algorithms continue to be developed for LLMs, the continued question remains of how strong the privacy protection is. \cite{hayes2024inexact} considers the success of an approximate unlearning algorithm as whether an inference model (called a U-MIA) can distinguish between the unlearned model and a completely retrained one. More notably, \cite{hayes2024inexact} distinguishes between a ``population U-MIA'', where one attack model is trained and used for all data points, and a ``per-example U-MIA'' where a dedicated attacker is instantiated for each data point. The baseline methods used to validate most unlearning methods are population U-MIAs. However, \cite{hayes2024inexact} adapted the LiRA attack by \cite{carlini2022first} towards this task (called U-LiRA) and found that it was highly successful in distinguishing the unlearned and retrained models. They tested this on the PaLM 2 model \cite{anil2023palm2}, fine tuned on LIMA \cite{zhou2023lima}, and unlearned using NegGrad and GradDesc \cite{jang2023knowledge}. Using this per-example U-MIA, \cite{hayes2024inexact} was able to identify the unlearned model over $80\%$ of the time. Given this, they argue that more stringent methods should be used to evaluate approximate unlearning methods to avoid a false sense of security.

\textbf{Unlearning as a Privacy Risk:} \cite{Zanella_B_guelin_2020} found that model updates (such as unlearning) can possibly pose a privacy risk if an adversary has access to the versions of the model pre and post update. Through a differential analysis of the model output, a powerful membership inference attack can be performed on the model. For more details on this, see Section~\ref{sec:mi_attacks}. Their attack highlights how machine unlearning is not an orthogonal concept to differential privacy, and can in some cases increase risk from a data privacy perspective.

\section{Conclusion}
In this paper we covered a broad range of privacy risks related to the use of Large Language Models. As these models continue to proliferate into both the public consciousness and into different application domains, the concern over the apparent privacy risks associated with their use is set to grow. This is reflected in the fact that forthcoming AI regulation in the US and EU devote significant attention to privacy risks. At the same time, basic questions related to LLM privacy remain largely unresolved. We know that LLMs can memorize swathes of training data (Section~\ref{sec:memorization}), but when compared to the reported attack success rates against discriminative deep learning models the existing attacks discussed in Section~\ref{sec:mi_attacks} are relatively weak. Getting a real sense of what privacy risks practical attacks can pose to LLMs will be key to ensuring that personal data is not unintentionally exposed, and on the other side, policymakers and AI developers don't overreact to theoretical risks that may not be likely in practice. Privacy-preserving training techniques like differential privacy seem like an effective way to preserve privacy and still maintain performance during fine-tuning, but much more research is needed to verify if these early findings hold across different domains, and to ease the path to adoption. Preserving privacy during LLM pre-training is largely unsolved. As the exact ways that copyright laws will apply to generative models crystallizes over the next few years, there will be a host of technical questions that arise. Among many others: Given a specific model output, how can we verify if it constitutes fair use with respect to a subset of the training data? If we want to fairly compensate the creators of the original training data with royalties related to a specific output, how can we perform this attribution? Existing methods to unlearn data from LLMs are heuristic -- how can we verify they work as intended? Approaches based on gradient descent require white-box updates to model parameters which is increasingly computationally expensive and restricted, and recent techniques based on in-context learning may not scale to large numbers of data deletions. More research is needed to both verify that these unlearning methods work as intended from a privacy standpoint, and can scale to practical models and deletion workloads.




\bibliographystyle{unsrtnat}
\bibliography{references}  






\newpage
\appendix
\renewcommand\thefigure{\thesection.\arabic{figure}} 

\section{Membership Inference Attacks}
\label{sec:mia_app}
\setcounter{figure}{0}

\begin{algorithm}[!ht]
\caption{MIA on Word Embeddings, \cite{song2020information}}
\label{alg:word_mia}
\begin{algorithmic}[1]
\State \textbf{Input:} Target window of words $C=[w_b,...,w_0,...,w_e]$, word embedding matrix $\Phi$, similarity function $\delta$.

\State Map words in $C$ with $\Phi$ to get $[\Phi_{w_b},...,\Phi_{w_0},...,\Phi_{w_e}]$
\State $\Delta \leftarrow \{ \delta_i|\delta_i = \delta(\Phi_{w_0},\Phi_{w_i}), \forall w_i \in C/\{ w_o \} \}$
\State \Return ``member'' if $\frac{1}{|\Delta|}\sum_{\delta_i \in \Delta} \delta_i \geq \tau_m$ else ``non-member''

\end{algorithmic}
\end{algorithm}

\begin{algorithm}[!ht]
\caption{Aggregated-level MIA on Sentence Embeddings, \cite{song2020information}}
\label{alg:sentence_mia}
\begin{algorithmic}[1]
\State \textbf{Input:} Target sentences in context $S = [s_1, ... , s_n]$, sentence embedding model $\Phi$, similarity function $\delta$.

\State Map sentences in $S$ with $\Phi$ and get $[\Phi(s_1),...,\Phi(s_n)]$.
\State $\Delta \leftarrow \{ \delta_i|\delta_i = \delta(\Phi(s_i),\Phi(s_{i+1}))\}_{i=1}^{n-1}$
\State \Return ``member'' if $\frac{1}{|\Delta|}\sum_{\delta_i \in \Delta} \delta_i \geq \tau_m$ else ``non-member''

\end{algorithmic}
\end{algorithm}

\begin{algorithm}[!ht]
\caption{Online Likelihood Ratio Attack (LiRA), \cite{carlini2022first}}
\label{alg:lira}
\begin{algorithmic}[1]
\State \textbf{Input:} model $h$, example $(x,y)$, data distribution $\mathcal{D}$
\State $confs_{in}=\{ \}$
\State $confs_{out}=\{ \}$
\For{$N$ times}
    \State $D_{attack} \leftarrow \mathcal{D}$ \hspace{4cm} \% Sample shadow dataset
    \State $h_{in} \leftarrow \mathcal{T}(D_{attack} \cup \{ (x,y) \})$ \hspace{1.65cm} \% Train \texttt{IN} model
    \State $\text{confs}_{in} \leftarrow \text{confs}_{in} \cup \{ \phi(h_{in}(x)_y) \}$
    \State $h_{out} \leftarrow \mathcal{T}(D_{attack} \textbackslash \{ (x,y) \})$ \hspace{1.7cm} \% Train \texttt{OUT} model
    \State $\text{confs}_{out} \leftarrow \text{confs}_{out} \cup \{ \phi(h_{out}(x)_y) \}$
\EndFor
\State $\mu_{in} \leftarrow mean(\text{confs}_{in})$
\State $\mu_{out} \leftarrow mean(\text{confs}_{out})$
\State $\sigma^2_{in} \leftarrow var(\text{confs}_{in})$
\State $\sigma^2_{out} \leftarrow var(\text{confs}_{out})$
\State $\text{confs}_{obs} = \phi (h(x)_y)$ \hspace{3.45cm} \% query target model

\State \Return $\Lambda = \frac{p(\text{conf}_{obs} | \mathcal{N}(\mu_{in}, \sigma^2_{in})}{p(\text{conf}_{obs} | \mathcal{N}(\mu_{out}, \sigma^2_{out})}$

\end{algorithmic}
\end{algorithm}

\section{Private Learning}
\label{sec:private_learning_app}
\setcounter{figure}{0}

\begin{figure}[h]
  \centering
  \includegraphics[width=.6\linewidth]{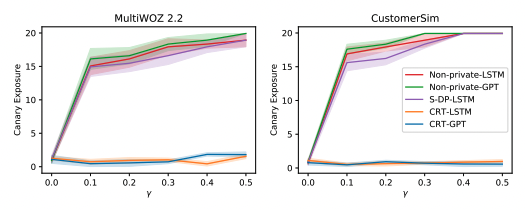}
  \caption{Results of a canary extraction attack on CRT model from \cite{zhao2022provably}. LSTM and GPT refer to model type tested. S-DP-LSTM model was trained using selective differential privacy from \cite{shi2022selective}. $\gamma$ is the error rate in the policy function screening for sensitive data. \cite{zhao2022provably}'s CRT method is effective at preventing this attack while S-DP-LSTM still is at risk.}
  \label{fig:zhao_canary}
\end{figure}

\begin{algorithm}[!ht]
\caption{\textit{SubMix} Training, \cite{ginart2022submix}}
\label{alg:submix_training}
\begin{algorithmic}[1]
\State \textbf{Input:} User-level private corpus $\mathcal{D}$, LM fine-tuning routine $\mathcal{L}$
\State \textbf{Outputs:} Fine-tuned LMs $h_{\pi_{i}}$,$h_{\pi_{i}'}$ for $i=1,...,k$
\State \textbf{Hyperparameters:} \# of parts $k$

\State $\prod \leftarrow$ Random $k$-fold partition of $\mathcal{D}$ with $|\prod_i|=|\mathcal{D}|/k$
\For{$i \in $ \{ $1,...,k$ \} }
    \State $(\pi_i, \pi_i') \leftarrow$ Randomly split part $\prod_i$ into two subparts.
    \State $h_{\pi_i} \leftarrow \mathcal{L}(\pi_i), h_{\pi_i'} \leftarrow \mathcal{L}(\pi_i)$
\EndFor
\end{algorithmic}
\end{algorithm}

\begin{algorithm}[!ht]
\caption{\textit{SubMix} Prediction, \cite{ginart2022submix}}
\label{alg:submix_prediction}
\begin{algorithmic}[1]
\State \textbf{Input:} Fine-tuned LMs $h_{\pi_{i}}$,$h_{\pi_{i}'}$ for $i=1,...,k$, privacy parameters $\epsilon$, time step $t$, query context $x_t \in \Sigma*$
\State \textbf{Outputs:} Next-token response $y_t \in \Sigma$ 
\State \textbf{Hyperparameters:} R\'enyi divergence order $\alpha$, target leakage $\beta$.

\If{$t=1$}
    \State $\varepsilon_i \leftarrow \epsilon$ for $i=1,...,k$
\ElsIf{STOP has been issued}

    \Return $y_t \sim h_{\emptyset}(x_t)$
\EndIf
\For{$i=1,...,k$}
    \State $\bar{h}_i(x_t) \leftarrow \frac{1}{2}(h_{\pi_i}(x_t)+ h_{\pi_i'}(x_t))$
    \State $\lambda_i \leftarrow \max_{lambda \in [0,1]} \{ \lambda: D_i(x_t, \lambda) \leq \beta \}$
\EndFor

\State $\lambda^* \leftarrow \frac{1}{k} \sum_{i=1}^k \lambda_i$
\State $\bar{h}(x_t) \leftarrow \frac{1}{k}\sum_{i=1}^k \bar{h}_i(x_t)$
\State $h(x_t) \leftarrow \lambda* \bar{h}(x_t) + (1-\lambda^*) h_{\emptyset}(x_t)$

\For{$i=1,...,k$}
    \State $\lambda_{-i}* \leftarrow \frac{1}{k-1} \sum_{j \neq i} \lambda_j$
    \State $\bar{h}_{-i} \leftarrow \frac{1}{k-1}\sum_{j \neq i} \bar{h}_j(x_t)$
    \State $\mathfrak{h} \leftarrow \lambda^* \bar{h}(x_t) + (1-\lambda^*) h_{\emptyset}(x_t)$
    \State $\mathfrak{h}' \leftarrow \lambda_{-i}^* h_{-i}(x_t) + (1-\lambda_{-i}^*) h_{\emptyset}(x_t)$
    \State $\varepsilon_i \leftarrow \varepsilon_i - \max \{ D_{\alpha}(\mathfrak{h} \parallel \mathfrak{h}'), D_{\alpha}(\mathfrak{h}' \parallel \mathfrak{h}) \}$
\EndFor
\If{$\forall i: \varepsilon_i > 0$}
    \State $y_t \sim h()x_t$
\Else
    \State Issue STOP signal
    \State $y_t \sim h_{\emptyset}(x_t)$
\EndIf

\State \Return $y_t$
\end{algorithmic}
\end{algorithm}

\begin{figure}[h]
  \centering
  \includegraphics[width=.45\linewidth]{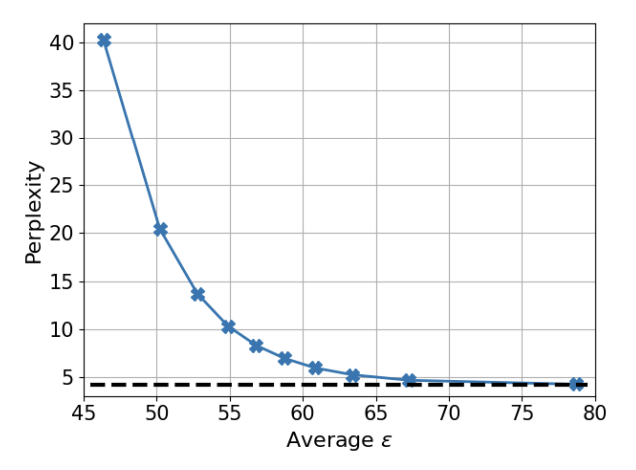}
  \caption{Privacy-Utility tradeoff from the perturbation methods of \cite{majmudar2022differentially}. This figure shows perplexity as a function of DP-$\epsilon$ values. The dotted line is the perplexity of the model without privacy guarantees. We see that the model seems to lose most utility at an $\epsilon \leq 50$.}
  \label{fig:majmudar}
\end{figure}

\begin{algorithm}[!ht]
\caption{\texttt{DP-FedAvg} and \texttt{DP-FedSGD}, \cite{mcmahan2018learning}}
\label{alg:dpfed}
\begin{algorithmic}[1]
\State \textbf{Main Algorithm}
\State \textbf{Input:} User selected probability $q \in (0,1]$, per-user example cap $\hat{w} \in \mathbb{R}^+$, noise scale $z \in \mathbb{R}^+$, estimator $\tilde{f}_f$ or $\tilde{f}_c$ with param $W_{min}$, UserUpdate (for FedAvg or FedSGD), ClipFn (FlatClip or PerLayerClip), moments accountant $\mathcal{M}$ \cite{Abadi_2016}

\State Initialize model $\theta^0$, moments accountant $\mathcal{M}$, $w_k = min(\frac{n_k}{\hat{w}},1)$ for all users $k$, $W = \sum_{k \in d} w_k$.

\For{each round $t=0,1,...$}
    $\mathcal{C}^t \leftarrow$ (sample users with probability $q$)
    \For{each user $k \in \mathcal{C}^t$}
        \State $\Delta^{t+1}_k \leftarrow \text{UserUpdate}(k,\theta^t, \text{ClipFn})$
    \EndFor
    \State $ \Delta^{t+1} = \begin{cases}
        \frac{\sum_{k \in \mathcal{C}^t}w_k \Delta_k}{qW} & for \tilde{f}_f \\
        \frac{\sum_{k \in \mathcal{C}^t}w_k \Delta_k}{\max (qW_{min}, \sum_{k \in \mathcal{C}^t}w_k)} & for \tilde{f}_c
    \end{cases} $
    \State $S \leftarrow$ (bound on $\lVert \Delta_k \rVert$ for \text{ClipFn})
    \State $\sigma \leftarrow \{ \frac{zS}{qW}$ for $\tilde{f}_f$ or $ \frac{2zS}{qW_{min}}$ for $\tilde{f}_c \}$
    \State $\theta^{t+1} \leftarrow \theta^t + \Delta^{t+1}+\mathcal{N}(0, I\sigma^2)$
    \State $\mathcal{M}$.\texttt{accum\_priv\_spending}(z)
\EndFor
\State print $\mathcal{M}$.\texttt{get\_privacy\_spent}()
\\

\State \textbf{FlatClip}($\Delta$):
\State \textbf{Parameters}: $S$
\State \Return $\pi(\Delta, S)$
\\

\State \textbf{PerLayerClip}($\Delta$):
\State \textbf{Parameters}: $S_1,...,S_m$
\State $S = \sqrt{\sum_j S^2_j}$
\For{each layer $j \in \{ 1,...,m \}$}
    \State $\Delta'(j) = \pi(\Delta(j),S_j)$
\EndFor
\State \Return $\Delta'$
\\

\State \textbf{UserUpdateFedAvg}($k, \theta^0$, ClipFn):
\State \textbf{Parameters}: $B, E, \eta$
\State $\theta \leftarrow \theta^0$
\For{each local epoch $i$ from $1$ to $E$}
    \State $\mathcal{B}\leftarrow$($k$'s data split into size $B$ batches)
    \For{batch $b \in \mathcal{B}$}
        \State $\theta \leftarrow \theta-\eta \nabla l (\theta;b)$
        \State $\theta \leftarrow \theta^0 +$ ClipFn($\theta-\theta^0$)
    \EndFor
\EndFor
\State \Return update $\Delta_k = \theta-\theta^0$ // \textit{Already clipping}
\\

\State \textbf{UserUpdateFedSGD}($k, \theta^0$, ClipFn):
\State \textbf{Parameters}: $B, \eta$
\State select a batch $b$ of size $B$ from $k$'s examples
\State \Return update $\Delta_k =$ ClipFn($-\eta \nabla l(\theta;b)$) 

\end{algorithmic}
\end{algorithm}

\section{Unlearning}
\label{sec:unlearning_app}
\setcounter{figure}{0}

\textbf{Strong vs Weak Unlearning:} When unlearning sequences, which in practical circumstances could grow arbitrarily long, it is important to consider the computational cost of $\mathcal{U}$, assuming a bound on the accuracy of the unlearning algorithm. This leads us to the definitions of \emph{strong} and \emph{weak} unlearning. First, we must define \emph{$(\alpha, \beta)$-accuracy}:

\begin{definition}
\label{def:ab_accuracy}
  ($(\alpha, \beta)$-Accuracy, \cite{neel2020descenttodelete}) We say a pair $(\mathcal{A},\mathcal{U})$ of learning and unlearning algorithms is $(\alpha,\beta)$-accurate if for every dataset $D \in \mathcal{D}$ and every update sequence $S$, the following condition holds: for every $i \geq 0$:
  $$\text{Pr} \left[ f_{D_i}(\Tilde{h}_i) - \min_{h \in \mathcal{H}}f_{D_i}(h) \right] < \beta$$
\end{definition}

Where $f_{D_i}(h)$ is the loss of model $h$ on the dataset after update $i$ and $\Tilde{h}_i$ is the most recent published model.

\begin{definition}
\label{def:strongweak}
  (Strong vs Weak Unlearning, \cite{neel2020descenttodelete}) Fix any pair $(\mathcal{A},\mathcal{U})$ of learning and unlearning algorithms that satisfy $(\alpha,\beta)$-accuracy with respect to some publishing function $f_{publish}$. Let $C_i$ represent the overall computational cost of the unlearning algorithm at step $i$ of the update. We say $\mathcal{U}$ is a ``strong'' unlearning algorithm for $\mathcal{A}$ if:
  \begin{enumerate}
      \item $\alpha$ and $\beta$ are independent of the length of the update sequence, and
      \item For every $i \geq 1, C_i/C_1 = \mathcal{O}(\log(i))$, i.e. the computation cost of the unlearning algorithm must grow at most logarithmically with $i$.
  \end{enumerate}
  If (1) holds and $\forall i \geq 1, C_i/C_1 = \Omega(poly(i))$, we say $\mathcal{U}$ is a ``weak'' unlearning algorithm for $\mathcal{A}$.
\end{definition}

Most work in the LLM unlearning space has been done under the weak unlearning framework.

\end{document}